\pdfoutput=1

\documentclass[nohyperref,final]{article}

\PassOptionsToPackage{final,pagebackref=true}{hyperref}
\PassOptionsToPackage{dvipsnames}{xcolor}
\usepackage{hyperref}

\usepackage[accepted]{icml2022}

\usepackage{preamble}
\icmltitlerunning{Evaluating SSL via Risk Decomposition}

\begin{document}

%\doparttoc % Tell to minitoc to generate a toc for the parts
%\faketableofcontents % Run a fake tableofcontents command for the partocs
%% \part{} % Start the document part
%% \parttoc % Insert the document TOC

\twocolumn[
\icmltitle{Evaluating Self-Supervised Learning via Risk Decomposition}
\ydnote{Other title ``analyzing self-supervised learning via a risk decomposition''}

% It is OKAY to include author information, even for blind
% submissions: the style file will automatically remove it for you
% unless you've provided the [accepted] option to the icml2022
% package.

% List of affiliations: The first argument should be a (short)
% identifier you will use later to specify author affiliations
% Academic affiliations should list Department, University, City, Region, Country
% Industry affiliations should list Company, City, Region, Country

% You can specify symbols, otherwise they are numbered in order.
% Ideally, you should not use this facility. Affiliations will be numbered
% in order of appearance and this is the preferred way.
\icmlsetsymbol{equal}{*}

\begin{icmlauthorlist}
\icmlauthor{Yann Dubois}{stanford}
\icmlauthor{Tatsu Hashimoto}{stanford}
\icmlauthor{Percy Liang}{stanford}
\end{icmlauthorlist}

\icmlaffiliation{stanford}{Department of Computer Science, Stanford University}

\icmlcorrespondingauthor{Yann Dubois}{yanndubs@cs.stanford.edu}

% You may provide any keywords that you
% find helpful for describing your paper; these are used to populate
% the "keywords" metadata in the PDF but will not be shown in the document
\icmlkeywords{Machine Learning, Self-Supervised Learning, Evaluation, Large scale}

\vskip 0.3in
]

% this must go after the closing bracket ] following \twocolumn[ ...

% This command actually creates the footnote in the first column
% listing the affiliations and the copyright notice.
% The command takes one argument, which is text to display at the start of the footnote.
% The \icmlEqualContribution command is standard text for equal contribution.
% Remove it (just {}) if you do not need this facility.

\printAffiliationsAndNotice{}  % leave blank if no need to mention equal contribution
%\printAffiliationsAndNotice{\icmlEqualContribution} % otherwise use the standard text.

\newcommand\pl[1]{\textcolor{red}{[PL: #1]}}

\begin{abstract}
Self-supervised learning (SSL) pipelines differ in many design choices such as the architecture, augmentations, or pretraining data.
Yet SSL is typically evaluated using a single metric: linear probing on ImageNet.
This does not provide much insight into why or when a model is better, nor how to improve it.
To address this, we propose an SSL risk decomposition, which generalizes the classical supervised approximation-estimation decomposition by considering errors arising from the representation learning step.
Our decomposition consists of four error components:
approximation, representation usability, probe generalization, and encoder generalization.  
We provide efficient estimators for each component and use them to analyze the effect of 30 design choices on $\Npre$ SSL vision models evaluated on ImageNet.
Our analysis gives valuable insights for designing and using SSL models.
For example, it highlights the main sources of error and shows how to improve SSL in specific settings (full- vs few-shot) by trading off error components. All results and pretrained models are at
\codeurl{}%\looseness=-1
\end{abstract}

\section{Introduction}
\label{sec:introduction}

Self-supervised learning (SSL) is a popular approach for pretraining an encoder from minimal supervision, such that linear probes trained on the encoder's representation perform well on downstream tasks.
SSL pipelines differ in many design choices, such as the objective \cite{chen_simple_2020,he_masked_2022}, architecture \cite{caron_emerging_2021,bardes_vicregl_2022}, augmentations \cite{tian_contrastive_2020,dubois_improving_2022} or pretraining data.
Yet SSL models are typically evaluated using a single metric: linear probing on ImageNet.
This is convenient for leaderboards but does not provide much insight into why or when a model is better, nor how to improve it.
What are the major sources of errors in current SSL methods?
Are there tradeoffs between SSL models across different settings (\eg full- vs few-shot probing)?
How does each design choice affect the SSL model?
Those are difficult to answer using a single metric.\looseness=-1

In supervised learning, one can get more fine-grained insights using the estimation/approximation (or bias/variance)
risk decomposition, which is estimated using the training and validation errors.
For example, models with low training error and high generalization gap often perform better in large-data regimes and can be improved via regularization.
In this paper, we generalize this classical decomposition to SSL. 
 Our decomposition consists of four sources of errors:
 \begin{enumerate}[noitemsep,leftmargin=*]
\item \textbf{approximation} errors due to the encoder's architecture not having the capacity to perform the task;
\item \textbf{representation usability} errors due to using SSL followed by linear probing. 
Usability error is large if a given SSL algorithm fails to produce linearly separable representations that can be used to predict desired tasks;
\item \textbf{probe generalization} errors due to finite training data; 
\item \textbf{encoder generalization} errors due to pretraining the encoder on finite data.
 \end{enumerate}

 We further provide computationally efficient estimators for each risk component, akin to the training and validation errors in supervised learning. 
Using those estimators, we analyze $\Npre$ pretrained SSL models and the effect of $\Nhparam$ design choices.
These results provide insights into the state of the field, help understand design choices, and suggest which SSL encoder to choose in various settings.

\begin{figure}[h]
\centering
\includegraphics[width=0.89\linewidth]{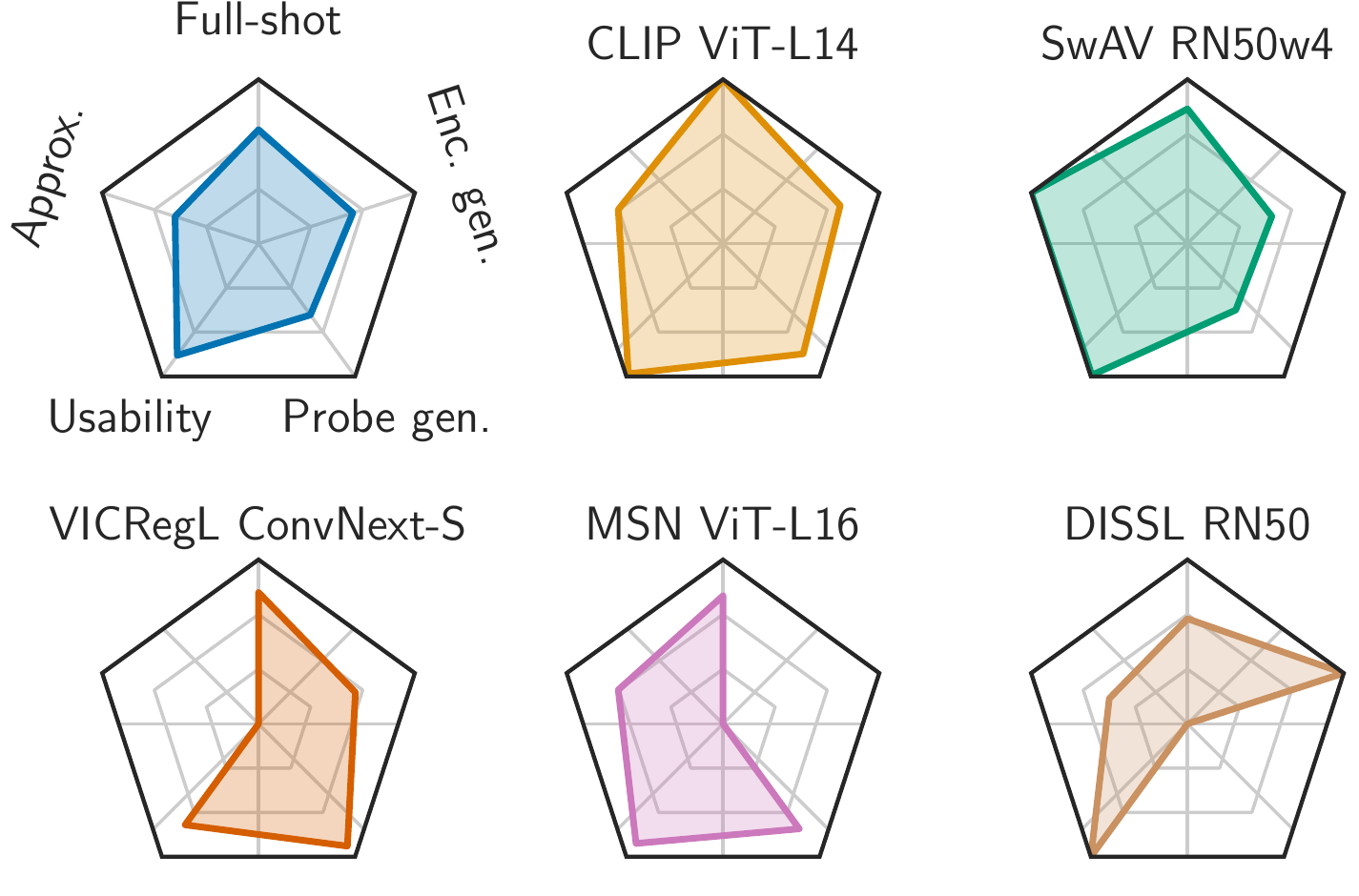} 
\caption{
No model is uniformly better over risk components.
``full-shot'' axis shows linear probing on ImageNet. 
Other axes show normalized risk components.
Higher is better.
Top left (blue) shows average over all \Npre{} models.
}
\label{fig:radar}
%\vspace*{-1.25em}
\end{figure}

Our analysis highlights that the most important source of error used to be representation usability but, since SimCLR, it is now the probe generalization.
Furthermore, we show that some design choices (\eg large projection heads, ViT encoders) improve all error components simultaneously.
But others (\eg representations' dimensionality or SSL objective) tradeoff components and thus only help in specific settings. 
For example, \cref{fig:radar} shows that SwAV RN50w4 gives more usable representations (bottom left) than MSN ViT-L16 \cite{assran_masked_2022} but induces worst probe generalization (bottom right).
This results in the former being better in full-shot probing (76\% vs 74\% accuracy) but worse in 3-shot (37\% vs 63\% ). In summary, we:
\begin{itemize}[noitemsep,leftmargin=*]
\item  provide an SSL risk decomposition with an efficient estimator for each error component;
\item show that the main source of error for modern SSL is the generalization error of linear probes;
\item highlight a tradeoff between usability and probe generalization, which leads to a few- vs full-shot tradeoff; 
\item analyze how $\Nhparam$ design choices affect the risk components and full-/few-shot performance of $\Npre$ SSL models.
\end{itemize}

\section{Supervised risk decomposition}
\label{sec:excess_risk_supervised}

In supervised learning, one learns a predictor $\predS{}$ from a hypothesis class $\Qx{}$ using a finite set of supervised samples $S$.
The goal is for the predictor to achieve low population risk $\rS$, which can be evaluated using a test set.
When designing models, it is nevertheless typical to consider both the training performance and the generalization gap (the difference between validation and training performance).
This is useful to understand which component of the pipeline to improve (regularization, architecture, etc) and which model should be favored depending on the training size $|S|$.

\begin{figure}[h]
    \centering    
    \vspace*{-2em}
    \includegraphics[width=0.7\linewidth]{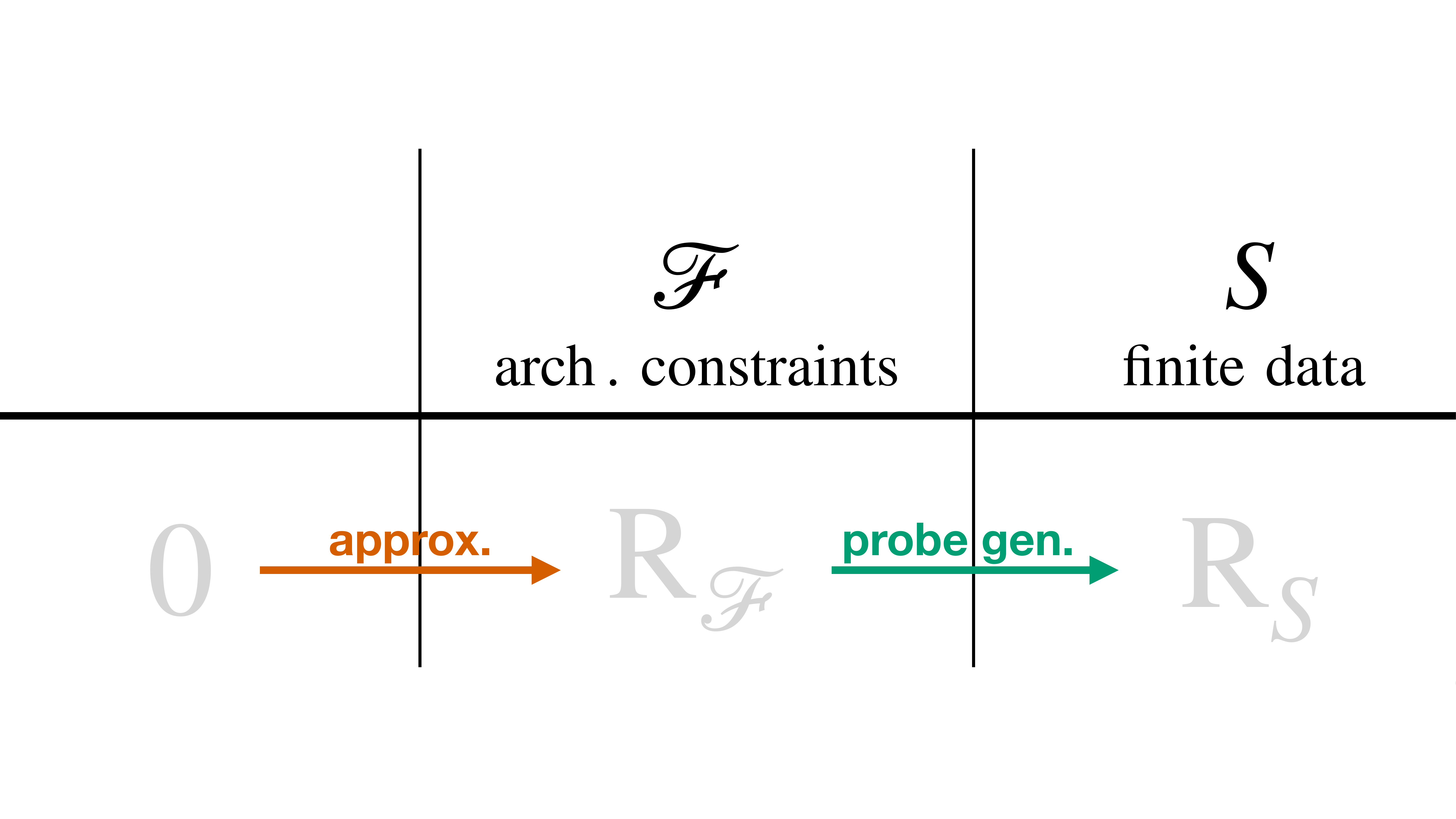}
    \vspace*{-1.5em}
    \caption{
    The risk decomposition is a path between settings of increasing expected risk for training the probe: 0 $\to$ $\rF$ (constrained family $\Qx{}$) $\to$ $\rS$ (finite supervised data).
    }
    \label{fig:main_row} 
\end{figure}

The training performance and generalization gap are respectively estimators of the \textit{approximation error} and the \textit{estimation error} from the supervised risk decomposition  \cite{barron_approximation_1994,shalevshwartz_understanding_2014}.
\footnote{For conciseness, we assume in the main paper that the irreducible error is 0, as it is independent of any design choice.
In appendices, we instead decompose the excess risk.
}
The approximation error $\rF$, is the error that a predictor $\predF$ trained on infinite data incurs, \ie, the error due to the choice of a constrained family $\Qx{}$.
The estimation error is the error due to training on finite samples, \ie, $\rF - \rS$.
As seen in \cref{fig:main_row}, the decomposition arises by considering the difference of risk incurred in settings of increasing expected risk.

Formally, we learn a predictor $\predS \defeq \algsup{}(\hp{S})$ from a family $\Qx{} \subseteq \{f: \Xc\to\actspace{} \}$ using an algorithm $\algsup{}$ (\eg ERM) on an empirical distribution $\hp{S}$ induced by a training set $\Stask \iidsim \ptask{}(X,Y)$.
Denote by  $\Lp{f} \defeq \Ep{\ptask{}}{\ell(Y, f(X))}$ the risk w.r.t. a desired loss $\ell$.
To derive the decomposition we order the two risks $\rS{} \defeq 
\Lp{\predS}$, $\rF \defeq \inf_{f \in \Qx{}} \Lp{f}$ and use a telescoping sum. Details at \cref{appcs:sec:risk_dec:supervised}.%\looseness=-1

\section{SSL risk decomposition}
\label{sec:excess_risk_representation}

Our goal is to derive a risk decomposition for representation learning that allows better development and understanding of SSL.
SSL pipelines consist of two models: an encoder $\phi$ and a probe $f$.
The probe is trained in a supervised fashion and, following \cref{sec:excess_risk_supervised}, it is useful to consider the errors that arise from using a constrained family $\Qz$ and finite data $\Stask$.

The difference with \cref{sec:excess_risk_supervised} is that the probe does not predict from inputs $X$ but from their representations $\phi(X)$.
As a result, errors also arise from the encoder $\phi \in \Phi$, which is pretrained from a family $\Phi$ using an SSL algorithm $\algrep$ and finite unsupervised data $\Spre \iidsim \ppre$.
Errors can thus come from each of the probe's limitations (constrained $\Qx{}$, finite $\Stask$) as well as each of the encoder's limitations (constrained $\Phi$, SSL algorithm $\algrep$, finite $\Spre$).
We now give an overview of each error component, which we formalize later.

\begin{figure}[h]
    \centering
    \includegraphics[width=0.9\linewidth]{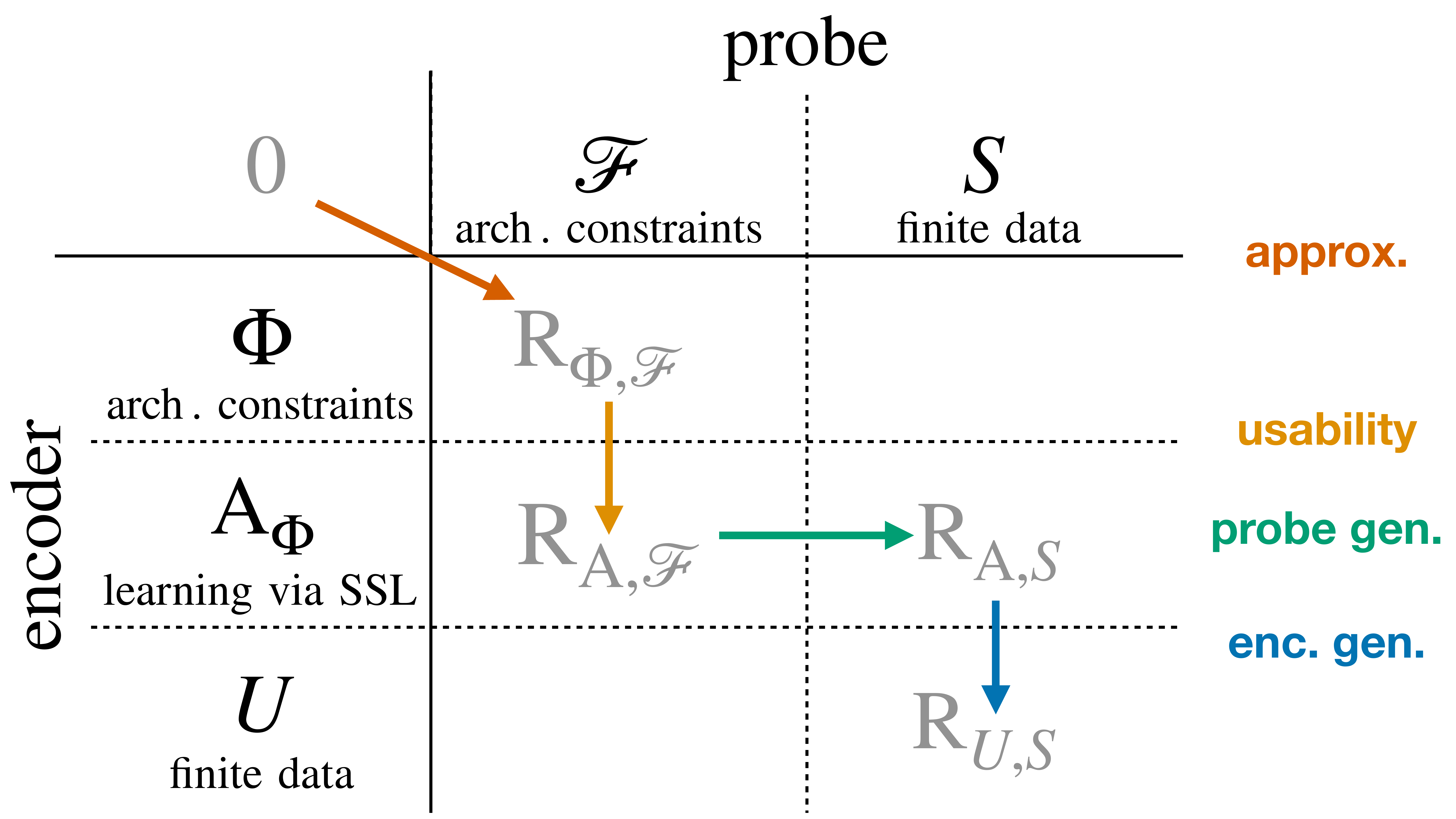}
    \caption{
    Our SSL decomposition is a path between settings of increasing expected risk.
    Columns show probe's limitations (constrained $\Qx{}$, finite supervised data $\Stask$) as in \cref{fig:main_row}.
    Rows show encoder's limitations (constrained $\Phi$, SSL algorithm $\algrep$, finite unlabeled data $\Spre$).
    %The $1$\textsuperscript{st} and $2$\textsuperscript{nd} subscript of each risk respectively indicate the row and the columns of that cell.
    Risk components (colored) are the differences between risks in two settings. 
    %The first subscript of each risk indicates the column (encoder), while the second 
    }
    \label{fig:main_matrix}
    %\vspace*{-1em}
\end{figure}

\begin{table*}[t]
\begin{equation}\label{eq:representation_risk_decomposition}
\underbrace{\vphantom{\int } \rUS }_{\vphantom{\int x^2} \text{Risk}}
 =  
 \underbrace{ \vphantom{\int } \rUS - \rAS}_{\vphantom{\int x^2} \color{myblue} \text{encoder generalization}} + 
  \underbrace{ \vphantom{\int } \rAS - \rAF}_{\vphantom{\int x^2} \color{mygreen} 
 \text{probe generalization}} + 
  \underbrace{ \vphantom{\int }   \rAF - \rFF}_{\vphantom{\int x^2}\color{myorange} \text{representation usability}} + 
 \underbrace{  \vphantom{\int } \rFF }_{\vphantom{\int x^2}\color{myred} \text{ approximation}} 
\end{equation}
\vspace*{-1em}
\end{table*}

The \textbf{approximation error} measures errors due to the architecture of the encoder $\Phi$ (\eg ResNet50) 
and probe $\Qz$ (\eg linear) being too constrained to perform even the supervised task. 
Intuitively, it decreases with the capacity of $\Phi,\Qz$.

The \textbf{representation usability error} measures errors due to learning representations via an SSL pipeline $\algrep{},\ppre{}$, rather than supervised learning.
Intuitively, it is small if the SSL algorithm ensures that representations retain information that is usable by probes $\Qz$, \eg, linearly separable classes.

The \textbf{probe generalization error} measures the drop in performance due to training the probe on finite samples $\Stask{}$ instead of $\ptask$.
Intuitively, it is small if: 
\begin{inlinelist}
\item the number of training samples $|\Stask|$ is large, or
\item representations ensure that downstream probes are sample efficient, \eg, by minimizing the margin between same-class examples.
\end{inlinelist}

The \textbf{encoder generalization error} measures the drop in performance due to pretraining the encoder on finite samples $\Spre$ compared to the population $\ppre$.
Intuitively, it is small if: 
\begin{inlinelist}
\item $\algrep{}$ makes pretraining sample efficient, or
\item there are many pretraining examples $|\Spre|$.
\end{inlinelist}

To derive those risk components we follow \cref{sec:excess_risk_supervised} and take the difference in risk between settings of increasing expected risk for the encoder $(\Phi, \algrep, \Spre)$ and probe $(\Qx{}, \Stask)$.
This gives our SSL risk decomposition \cref{eq:representation_risk_decomposition}, which we illustrate in \cref{fig:main_matrix} as a path through the matrix $(\Phi, \algrep, \Spre) \times (\Qx{}, \Stask)$. %, which extends \cref{fig:main_row} by adding rows for each encoder's setting.
Each cell corresponds to the risk incurred for a specific limitation for the encoder (row and $1$\textsuperscript{st} subscript) and the probe (column and $2$\textsuperscript{nd}  subscript). Formally:\looseness=-1

\begin{itemize}[noitemsep,leftmargin=*]
\item  $\bm{\rFF{}}  \defeq \inf_{f\in \Qz} \inf_{\phi \in \Phi} \Lp{f \circ \phi}$ is the best possible risk for encoders in $\Phi$ and probes in $\Qz$.
\item  $\bm{\rAF{}}  \defeq \inf_{f\in \Qz}  \Lp{f \circ \encA{}}$ is the risk of the best probe in $\Qz$ and an encoder $\encA{} \defeq \algrep(\ppre) \in \Phi$ pretrained using the desired SSL algorithm and the population distribution.
\item $\bm{\rAS} \defeq  \Lp{\probeS{\encS} \circ \encA{}}$ is the risk incurred by the same encoder but using a probe trained from finite samples $\probeS{\encA} \defeq \algsup(\hpenc{\encA})$, where $\encA(S) \defeq \set{(\encA{}(x),y) \cond (x,y) \in \Stask}$ is the represented training set.
\item $\bm{\rUS} \defeq  \Lp{\probeS{\encS} \circ \encS{}}$ is the risk when the probe and encoder are trained from finite samples $\encS{} \defeq \algrep(\hp{U})$.
%\item  $\bm{\rUF} \defeq \inf_{f\in \Qz}  \Lp{f \circ \encS{}}$ is the risk when only the probe is trained from finite samples.
\end{itemize}

Our decomposition (\cref{eq:representation_risk_decomposition}) corresponds to the specific path  $ 0 \to \rFF{} \to \rAF{} \to \rAS \to \rUS$ in \cref{fig:main_matrix}.
Considering different paths through the matrix would give different decompositions.
In \cref{appcs:sec:risk_dec:alternatives}, we provide all other decompositions and show that those would be harder to estimate.

\section{Estimating risk components for SSL}
\label{sec:practical_estimation}

Our goal is to compare pretrained SSL models using our decomposition.
We would thus like estimators $\hat{R}$ of each risk component $R$ that are simple, computationally efficient, and applicable in the standard SSL ImageNet setting.
In this section, we provide such estimators.\looseness=-1

Compared to supervised learning, the main new challenge for estimating our risk components compared to supervised learning is that pretraining additional SSL encoders is computationally prohibitive, so we want each of our estimators to use the same SSL encoder.
This is a challenge because our risk components are defined using three different encoders ($\phi,\encA,\encS$).
Our key insight is that we can estimate risk components by changing the training and evaluation set of the probe using the same pretrained SSL encoder.

In the following, we illustrate this for the standard ImageNet SSL setting where the metric comes from pretraining encoders and training probes on the \textit{same} inputs $\Str$, and evaluating them on $\iid$ examples $\Ste$.
As a result, we can estimate risk components by training and evaluating probes on specific partitions of $\Str \cup \Ste$
 as summarized in \cref{tab:summary_estimation}.
We now provide the intuition behind each estimator.
For formal derivations, properties, and pseudocode 
see \cref{appx:sec:estimators}.
As a reminder, the encoder is always pretrained on $\Str$. 

\begin{itemize}%[leftmargin=*]
\item $\bm{\hrUS{}}$: \quad  We need to estimate the risk when both the encoder and the probe are trained on finite data.
They should thus both be evaluated on unseen data.
We do so by training the probe on $\Str$ and evaluating it on $\Ste$, \ie, we use the standard SSL metric.
As $\Ste$ is disjoint from both the encoder's and probe's (pre)training set $\Str$, this ensures that both models are evaluated on unseen data.
\item $\bm{\hrAS{}}$: \quad  We need to estimate the risk when the probe is trained on finite samples but the encoder is pretrained on the population.
To do so we use $\Str$ as a plug-in estimate for the population data, which we split into a training $\Ssub \subset \Str$ and testing set $\Str \setminus \Ssub$ for the probe.
This ensures that the probe is evaluated on unseen data but not the encoder.
\item $\bm{\hrAF{}}$: \quad We need to estimate the SSL risk when both the encoder and the probe are (pre)trained on the population distribution.
We do so by using the \textit{same} pretraining, training, and evaluating set $\Str$, which ensures that the encoder and probe are evaluated on data they were trained on.
$\hrAF{}$ is thus the training error of the probe used for standard evaluation.
\item $\bm{\hrFF{}}$: \quad 
We need to estimate the risk of the best possible predictor in the composed family $\Qz \circ \Phi$, without considering SSL or finite samples.
We do so using the \emph{training} error of a supervised model with architecture $\Qz \circ \Phi$, \eg, a ResNet50 on ImageNet.\footnote{$\hrFF{}$ requires training a supervised encoder $\phi \in \Qz \circ \Phi$, which can be inefficient. 
Thankfully, this can be reused for SSL models with the same architecture and can often be found online.}
\end{itemize}

Our estimators are simple and computationally efficient as they do not require retraining any other SSL encoder. 
Furthermore, each estimator improves as the dataset size increases.
This is similar to how supervised training and testing errors estimate $\rF{}$ and $\rS{}$.\looseness=-1

\begin{table}[h]
    \centering
    \caption{We estimate risk components of an encoder $\encS \in \Phi$ pretrained on ImageNet's train set $\Str$,
by training and evaluating probes on different partitions of ImageNet's train $\Str$ and test set $\Ste$.
  $\Ssub \subset \Str$ is a small training subset.
$\encSUP \in \Phi$ is a supervised encoder of the same family.\looseness=-1
    }
    \label{tab:summary_estimation}
    \begin{tabular}{lrrrr}
    \toprule
   &  &  \multicolumn{3}{c}{Dataset}  \\
\cmidrule(lr){3-5}\
   Estimator & Encoder  & Pretrain & Train  & Eval    \\
    \midrule
    $\hrUS$  &   $\encS$ & $\Str$ & $\Str$ & $\Ste$  \\
      $\hrAS$  &  $\encS$ & $\Str$ & $\Str \setminus \Ssub$ & $\Ssub$  \\
         $\hrAF$  &    $\encS$  & $\Str$& $\Str$ & $\Str$  \\
       $\hrFF$  & $\encSUP$  & $\Str$ & $\Str$ & $\Str$  \\
         \bottomrule
    \end{tabular}
\end{table}

\begin{table}[H]
\caption{Best performing models for ImageNet linear probing.
The first 4 categories of rows show models pretrained on ImageNet-1K of various architectures (RN50, any CNN, ViT-S/16, any ViT).
The last category allows any data and architecture.
Underlined results are best in their category, 
bolded ones are best overall.
Duplicate rows are removed.}
\label{tab:best_results}
\begin{small}
\begin{tabular}{lllrrr}
\toprule
&& & \multicolumn{3}{c}{ImageNet probe acc.} 
 \\
  \cmidrule(lr){4-6}
Obj. & Arch. & Param. & 100\% & 1\% & 3-shot \\
\midrule
MoCo-v3 & RN50 & 24M & 73.7 & \underline{55.5} & \underline{40.4} \\
DINO & RN50 & 24M & \underline{74.2} & 52.9 & 35.9 \\
\midrule
SwAV & RN50w4 & 375M & \underline{76.2} & 56.2 & 36.9 \\
VICRegL & CnvNxt-B & 85M & 74.8 & \underline{64.3} & \underline{56.3} \\
\midrule
MUGS & ViT-S16 & 22M & \underline{77.3} & 62.9 & 49.6 \\
MSN & ViT-S16 & 22M & 76.1 & \underline{67.5} & \underline{60.4} \\
\midrule
MSN & ViT-B4 & 86M & 80.1 & \underline{75.1} & 69.3 \\
MUGS & ViT-L16 & 303M & \underline{80.9} & 74.0 & 68.5 \\
MSN & ViT-L7 & 303M & 79.9 & 74.9 & \textbf{69.8} \\
\midrule
CLIP & ViT-L14 & 304M & \textbf{85.0} & 75.2 & 62.9 \\
OpenCLIP & ViT-H14 & 632M & 84.4 & \textbf{75.8} & 63.7 \\
\bottomrule
\end{tabular}
\end{small}
\end{table}

\section{Experimental results}
\label{sec:results}

In the following, we use our risk decomposition to answer the three motivating questions from \cref{sec:introduction}:
What are the major sources of errors in current SSL?
Are there tradeoffs affecting which models to prefer in certain settings?
How does each design choice affect the SSL model?

To do so we analyze $\Npre$ SSL pretrained encoders, 
across $\Nobj$ objectives, $\Narch$ architectures, and 7 years.
For each model, we collected \Nhparam{} design choices or hyperparameters, estimated our error components, and evaluated the ImageNet test performance of well-tuned linear probes trained on different subsets of ImageNet ($100\%$, 30-shot, $1\%$, 5-shot, 3-shot).
Note that only 14 of the encoders were pretrained by us, so there might be undesirable selection bias.
\looseness=-1

In our pursuit of addressing our motivating questions, we thus provide the most comprehensive benchmarking of self-supervised learning models to date.
We highlight the best-performing models in various settings in 
\Cref{tab:best_results}, which we will refer to throughout the section.

We also provide a simple \texttt{torch.hub} API at \codeurl{} to load all pretrained encoders, metadata, and results. 
For experimental details see \cref{appx:sec:reproducibility}, for raw results see \cref{appx:sec:raw_results}, and for extended analysis see \cref{appx:sec:res:hparams,appx:sec:secondary_results}.%\looseness=-1

\subsection{Major sources of errors}
\label{sec:results:trends}

In this section, we aim to understand the main sources of errors in current SSL, and how this might change over time.
Identifying important sources of errors is potentially useful for understanding what research to prioritize.

\begin{figure}[h]
    \centering
    \includegraphics[width=0.8\linewidth]{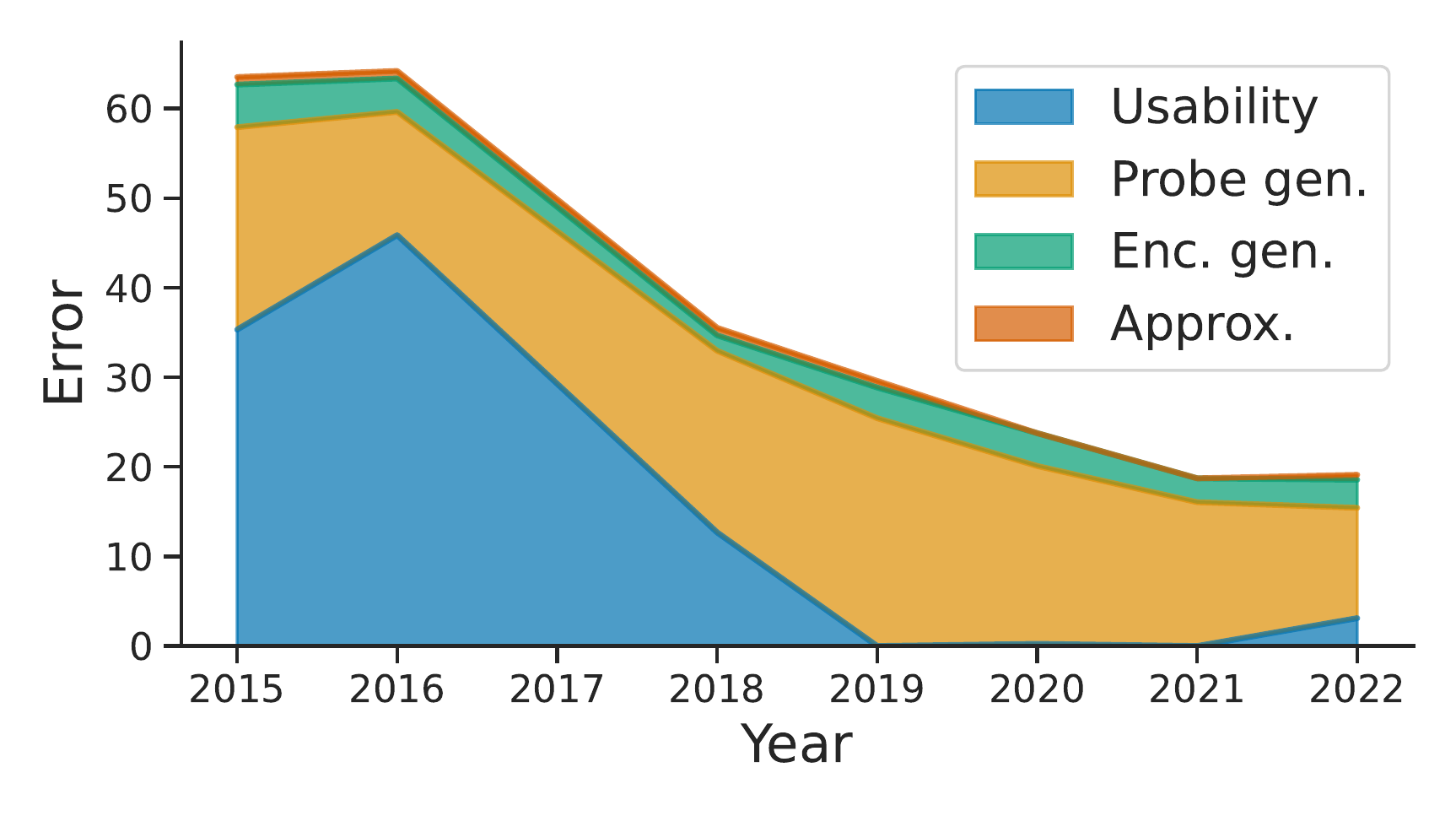}
    \vspace*{-1em}
    \caption{The major SSL improvements came from usability, but probe generalization is now the largest source of error.
    The plot shows risk components of the best ImageNet-pretrained model published in a given year.
    Lower is better.
    In \cref{appx:sec:results:trends} we show similar trends for the average models.
    }
    \label{fig:trends_main}
\end{figure}

\Cref{fig:trends_main} shows how error components have changed over time.
We now discuss each of them in detail.\looseness=-1

\begin{paragraph}{Representation usability drove improvements}
We see that representation usability, \ie, the inability of linear probes to extract information from representations, used to be the largest source of error but it has improved steadily between 2016-2019.
In \cref{appx:sec:results:trends} we show that advances in contrastive learning mostly drove those improvements.
\end{paragraph}

\begin{paragraph}{Probe generalization is now the bottleneck}
We see that probe generalization is now the largest source of error, which suggests that it should be prioritized.
For example, since 2019, the field has been able to improve overall performance by improving significantly this source of error.
\end{paragraph}

\begin{paragraph}{Encoder generalization is small and constant}
We see that the encoder generalization has been relatively small over time but might become important in the future.
\end{paragraph}

The fact that the generalization error is smaller for the encoder than the probe is surprising.
Indeed, they are both (pre)trained on the same data (ImageNet's training set) but the encoder is more ``complex'' than a regularized linear probe.
This requires further analysis but could be due to overparametrization \cite{belkin_reconciling_2019,yang_rethinking_2020}.

\begin{paragraph}{Approximation error is negligible}
Unsurprisingly, current encoders have the capacity to perform the desired task.
\end{paragraph}

For the rest of the paper, we focus on the most common
sources of errors: usability and probe generalization.

\subsection{Tradeoffs and full- vs few-shot performance}
\label{sec:results:tradeoffs_performance}

In this section, we first show that our estimators of usability and probe generalization are useful in choosing which models to prefer in full- or few-shot settings.
We then highlight a tradeoff between those two components that directly translates to a tradeoff between full- and few-shot performance.

\subsubsection{Predicting performance across settings}
\label{sec:res:settings}

Our risk decomposition isolates generalization errors, and should by construction give insights into which models to favor in full- vs few-shot settings.
Let us test whether this is also true when using our simple estimators.
As a reminder, error components are estimated on all of ImageNet but we analyze the performance of probes trained on varying numbers of train samples ($100\%$, $1\%$ and 30-, 5-, 3-shot).

\begin{figure}[h]
\centering
\begin{subfigure}[t]{0.49\linewidth}
\centering
\includegraphics[width=\linewidth]{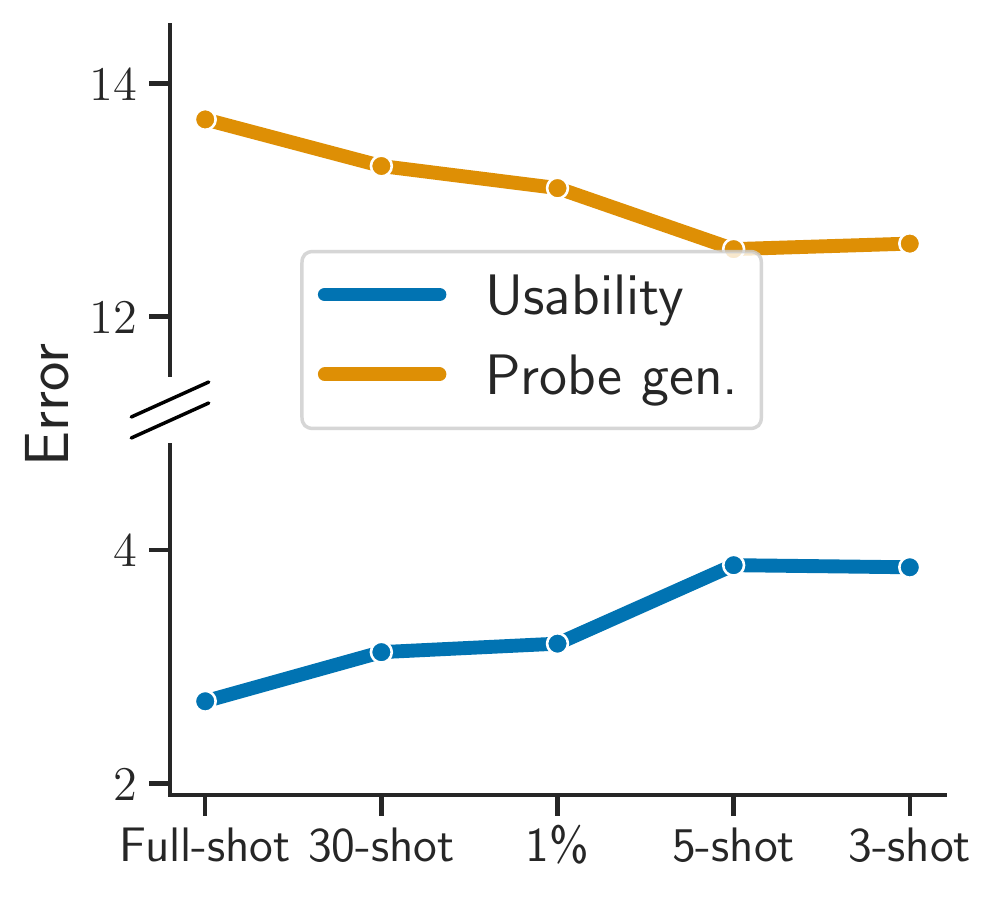}
\caption{Risk components}
\label{fig:evaluation_tradeoff}
\end{subfigure}
\hfill{}
\begin{subfigure}[t]{0.47\linewidth}
\includegraphics[width=\linewidth]{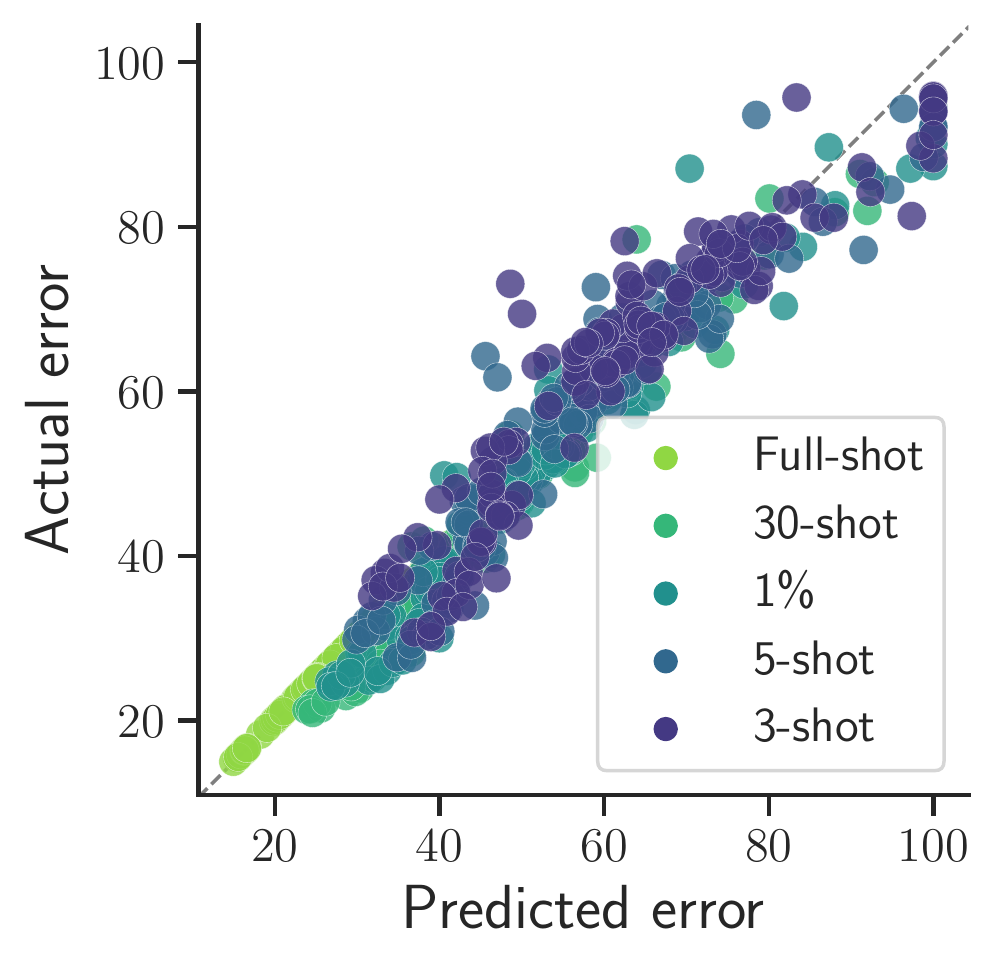}
\caption{Scaling law prediction}
\label{fig:evaluation_scalinglaw}
\end{subfigure}
\caption{
Our estimated risk components are tightly related with performance in different settings.
(a) Usability error of the best $20\%$ of models increases as the training samples decreases, while probe generalization error decreases.
(b) The performance predicted by our scaling law \label{eq:our_scaling_law} (x-axis) is close to the true performance (y-axis) for all data settings.}
\label{fig:evaluation_main}
\end{figure}

\paragraph{Probe generalization signals sample efficiency}
Intuitively, models with low probe generalization error perform better in few-shot settings (less variance) while those with low usability error perform better in full-shot settings (less bias).
\Cref{fig:evaluation_tradeoff} shows that, indeed, the best encoders in few-shot regimes have smaller probe generalization errors.
Can we use this relation to predict performance across settings?\looseness=-1

\paragraph{Error components predict performance across settings}
In \cref{appx:sec:results:scaling_law} we propose a simple 2-parameter scaling law that fits the performance of all \Npre{} models as a function of estimated error components and the number of training samples $|\Stask|$ (see \cref{fig:evaluation_scalinglaw}). 
We show that it performs significantly better than standard scaling laws \cite{kaplan_scaling_2020,rosenfeld_scaling_2021} both in held-out settings (test $R^2=0.94$) and held-out encoders (test  $R^2=0.96$ when holding out contrastive encoders).
While the scaling law will not save much computation (probes are efficient to train), it is a useful validation of our risk decomposition and estimators.

\subsubsection{Tradeoffs}
\label{sec:results:tradeoffs}

One advantage of the supervised risk decomposition is that it highlights a tradeoff between approximation/estimation.
Although this tradeoff does not always hold \cite{neal_modern_2018, yang_rethinking_2020,dar_farewell_2021}, it is a useful conceptual framework for developing models. 
For example, it suggests that high-capacity predictors perform better when there is plenty of training data and can benefit from regularization.\looseness=-1

In \cref{appcs:sec:risk_dec:tradeoffs} we derive three corresponding tradeoffs in SSL.
Two of those are not insightful as they depend on the negligible approximation error.
More interestingly, we derive a usability/probe generalization (U/P) tradeoff.
This corresponds to the standard approximation/estimation tradeoff but the gains in capacity come from changing the data (via encoding) rather than the predictor's family $\Qz$.
As an illustration, constant representations lead to probes that perform badly on training (high usability error) but have zero generalization error.
In contrast, if the representations are one-hot encodings of inputs, then linear probes can achieve perfect training performance (usability) but will not generalize.\looseness=-1

\begin{figure}[H]
%\vspace*{-0.4em}
    \centering
    \includegraphics[width=0.8\linewidth]{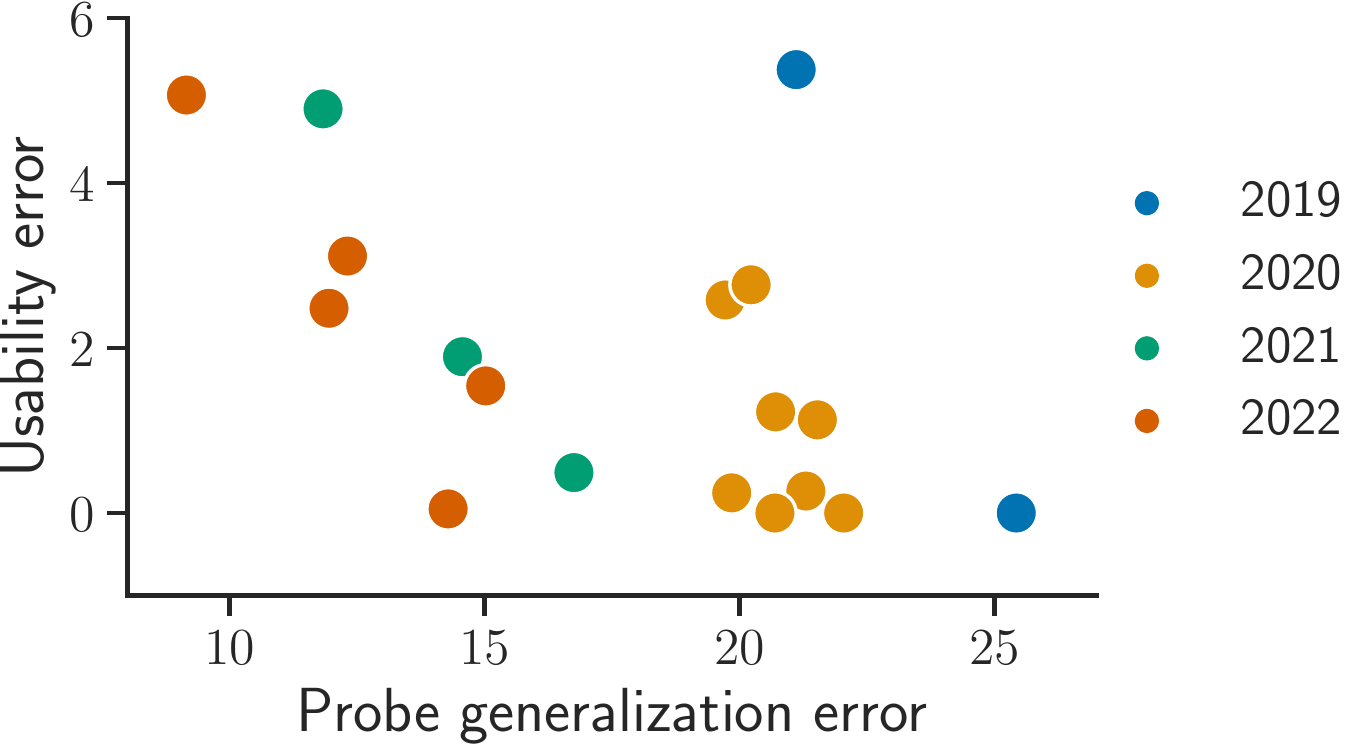}
    %\vspace*{-2em}
    \caption{Usability %(y-axis) 
    vs probe generalization %(x-axis) 
    tradeoff for the best 20\% of models for each year (color).
    Models differ in many design choices (\eg objective, architecture, epochs).
    }
    \label{fig:tradeoff_main}
\end{figure}

\paragraph{Usability/probe generalization tradeoff}
Similarly to approximation/estimation, U/P is not an exact tradeoff but suggests that decreasing one tends to increase the other.
This can be seen in
\cref{fig:trends_main}: between 2016 and 2019 usability decreased at the expense of probe generalization, and vice-versa since 2019.
This can also be seen in \cref{fig:tradeoff_main}: at every point in time, the best models seem to form a tradeoff curve.

\begin{table*}[t]
\caption{Effect of design choices on error components and full-/3-shot.  
\vgood{}: much better, \good{}: better,
\bad{}: worse,
\vbad{}: much worse.}
\label{tab:summary_hyperparam}
\centering
\begin{tabular}{lrrrrrrrrr}
\toprule
& \# dim. $\downarrow$ & \# views $\uparrow$ & ViT & \# param.$\uparrow$ & MLP proj. & generative SSL & \# epoch $\uparrow$ & Adam & \\
\midrule
 Usability error & \vbad{} & \vgood & & \vgood & \vgood & \vbad &  \good  &  &\\
 Probe gen. error & \vgood & \good & \vgood &&\good & \good & \good  & \vgood&\\
Full-shot error & \bad{} & \vgood & \good & \vgood & \vgood & \vbad & \vgood& \vgood &\\
 3-shot error & \vgood{} & \vgood & \vgood &\vgood &\good & \good & \vgood & \vgood &\\
 \bottomrule
\end{tabular}
\end{table*}

% \begin{figure}[h]
%     \centering
%     \begin{subfigure}[t]{0.48\linewidth}
%          \centering
%  \includegraphics[width=\linewidth]{figures/tradeoffs/tradeoff_probe-usability_all.pdf}
%     \caption{All models}
%     \label{fig:tradeoff_all}
%     \end{subfigure}
%     %
%     \hfill{}
%     %%
%     \begin{subfigure}[t]{0.48\linewidth}
% \includegraphics[width=\linewidth]{figures/tradeoffs/tradeoff_probe-usability_year.pdf}
% \caption{Best 15\% per year}
% \label{fig:tradeoff_year}
%     \end{subfigure}
%     \caption{
%     The usability vs probe gen tradeoff not 
%     }
%     \label{fig:tradeoff_main}
% \end{figure}

\paragraph{Full-/few-shot tradeoff} Given the relation between usability/probe generalization and performance in different settings (\cref{sec:res:settings}), we expect the U/P tradeoff to translate in a full-/few-shot tradeoff.
\Cref{tab:best_results} shows that, indeed, the best models in full-shot (100\%) settings are never the best ones in 3-shot.
This is true for the 5 considered categories.
\Cref{fig:evaluation_main} suggests that this is indeed driven by the U/P tradeoff.

\subsection{Analysing design choices}
\label{sec:res:hparams}

In this section, we analyze the impact of important SSL design choices on risk components and the performance in full- and 3-shot settings.
\Cref{tab:summary_hyperparam} summarizes our findings.
We use the following three methods to analyze our results:
\begin{itemize}[leftmargin=*]
\item \textbf{Controlled analysis (CA).} Whenever possible we analyze the effect of a design choice while fixing others.
To do so quantitatively, we fit a linear model from the current (possibly log-transformed) design choice to the metric: $\textit{metric} = \alpha \cdot \textit{hparameter} + \bm{\beta}^T  \mathds{1}[\textit{model}]$, where $\mathds{1}[\textit{model}]$ is a one-hot encoding of the value of all other design choices.
The downside is that we can only apply CA if we have encoders that only differ in the desired design choice.\looseness=-1
\item \textbf{XGBoost+SHAP.} For each risk component and metric, we train one XGBoost model \cite{chen_xgboost_2016} using all design choices and potential confounders (\eg year). 
We then perform feature selection to avoid feature redundancy.
Finally, we analyze the SHAP value \cite{lundberg_unified_2017} of the desired design choice.
The main disadvantage of XGBoost+SHAP is that there might be other confounders we did not consider.
\item \textbf{Global linear analysis (GLA)} For each metric and design choice, we train a linear model from all metadata that we think are either important to predict the metric or may be confounders.
The downsides of GLA are that it depends on our incomplete ``expert knowledge'' of how variables interact, and it makes a linearity assumption.
\end{itemize}

In the main paper, we focus on results from SHAP and qualitative CA, but write ``(GLA p-value)'' or ``(CA p-value)'' to show that the other analyses give consistent conclusions.
Although different analyses with consistent conclusions mitigate issues with the overall analysis, they do not imply any causal conclusions.
For more methodological details see \cref{appx:sec:details:hparam_impact}. For extended analysis of all results see \cref{appx:sec:res:hparams}.

\subsubsection{Dimensionality}
\label{sec:res:hparams:dimensionality}

\begin{figure}[h]
\centering
\vspace*{-0.5em}
\includegraphics[width=\linewidth]{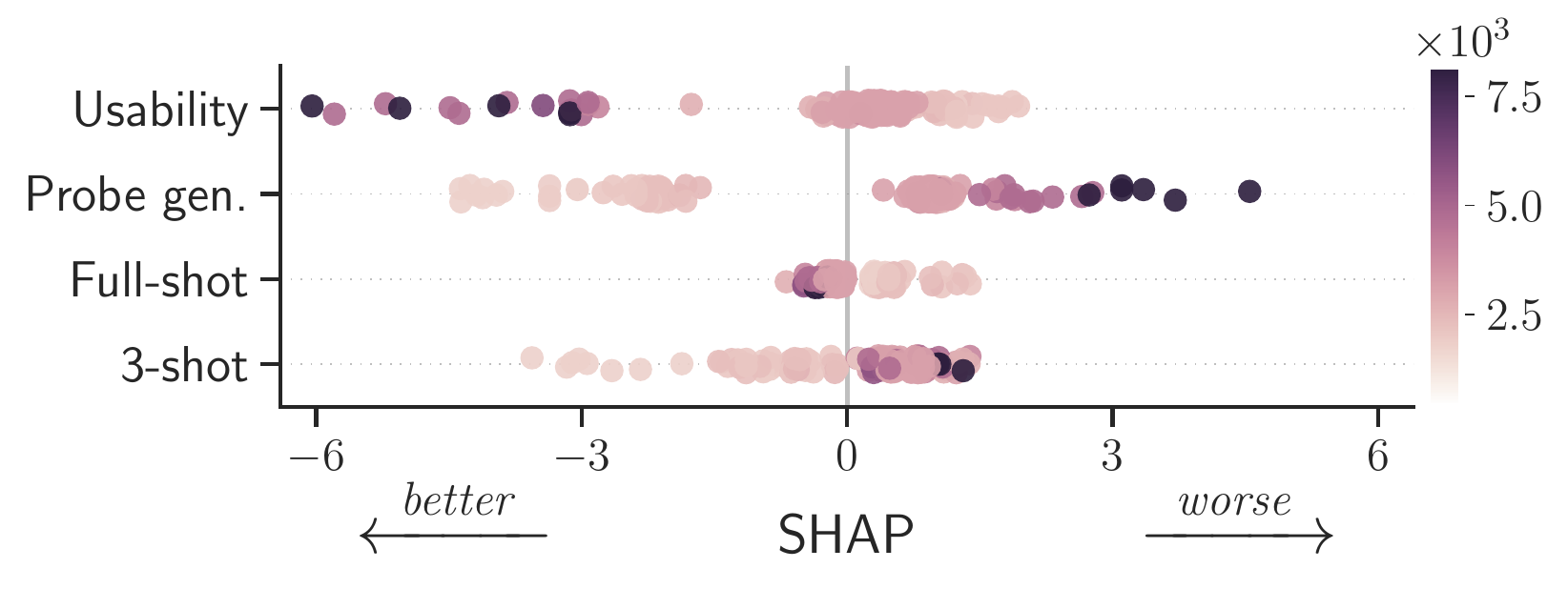}
%\vspace*{-2em}
\caption{Impact of the representation's dimensionality (color) on the usability error, probe generalization error, and full-/3-shot linear probing.
Impact is measured by SHAP values (x-axis).
Lower is better as it decreases the risk.
}    
\label{fig:dimensionality:main_shap}
\end{figure}

\paragraph{Increasing dimensionality improves usability at the expense of probe generalization}
\Cref{fig:dimensionality:main_shap} shows that increasing dimensionality improves usability but worsens probe generalization, which in turn worsens few-shot performance (\cref{sec:res:settings}).
This is further supported by our linear model in the global and controlled setting (GLA/CA p-values ${<}1\sci{9}$).
In \cref{appx:sec:res:dimensionality} we show that what matters is the effective dimensionality (rank) of the representation.

The effect of dimensionality can be intuitively understood by the fact that the capacity of linear classifiers depends on the input dimension $d$ \cite{vapnik_on_1971}, so increasing $d$ may improve performance but cause overfitting.
For a formal explanation see \citet{dubois_improving_2022}.

\begin{figure}[h]
\centering
%\vspace*{-1em}
\includegraphics[width=0.7\linewidth]{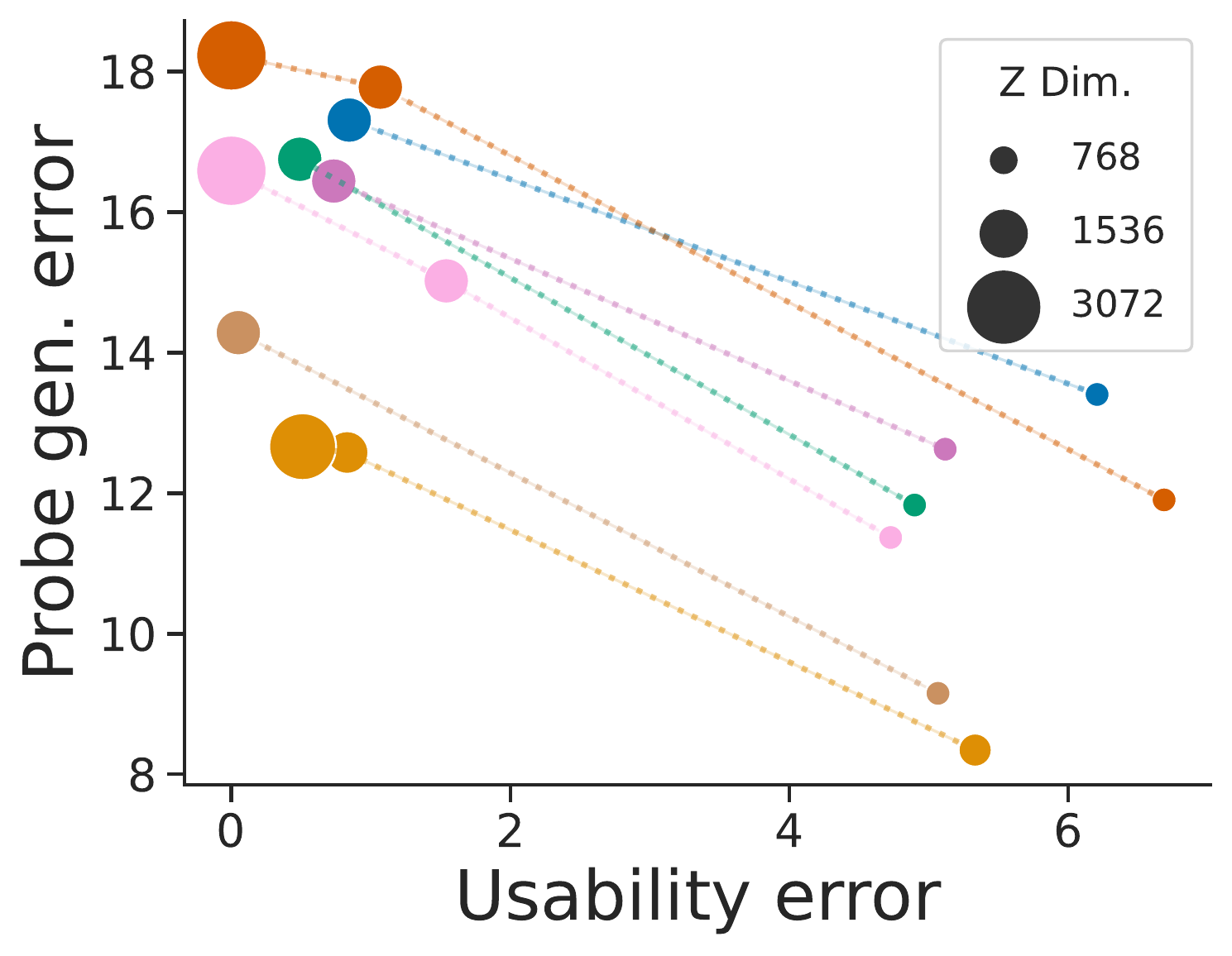}
\vspace*{-0.5em}
\caption{The representation's dimensionality trades off probe generalization and usability. 
Colors indicate representations from the same ViT. 
We concatenate CLS tokens from different blocks to vary the dimensionality (dot size).
}    \label{fig:dimensionality:main_tradeoff_components}
\end{figure}

\paragraph{Moving along the U/P tradeoff without retraining}
\Cref{appx:sec:res:dimensionality} suggests that dimensionality might be a simple way to move along the U/P tradeoff.
To test this, we vary the dimensionality ($d$, $2d$, $4d$) of representations from ViT encoders by either taking the [CLS] token from the last block, by concatenating the [CLS] token and the average of all other tokens, or by concatenating the [CLS] tokens from the last 4 
ViT blocks.
\Cref{fig:dimensionality:main_tradeoff_components} shows that this method allows trading off usability and probe generalization.

\begin{table}[h]
\caption{We improve few-shot performance by using representations from layers of smaller dimensionalities (``ours''). 
}
\label{tab:different_zlayer}
\begin{small}
\begin{tabular}{llllrrr}
\toprule
 Ours & Obj. & ViT & Dim. & 100\% & 1\% & 3-shot \\
\midrule
\xmark{} & MUGS & S16 & 1536 & \textbf{77.3} & 62.9 & 49.6 \\
$\checkmark{}$ & MUGS & S16 & 384 & 77.0   & \textbf{66.6} & \textbf{57.9} \\
\midrule
\xmark{} & OpenCLIP & H14 & 1280 & \textbf{84.4} & 75.8 & 63.7 \\
$\checkmark{}$  & OpenCLIP & H14 & 1024 & 84.3 & \textbf{76.5} & \textbf{65.5} \\
\bottomrule
\end{tabular}
\end{small}
\end{table}

\paragraph{Improving performance without retraining}   \Cref{fig:dimensionality:main_tradeoff_components} and \cref{sec:results:tradeoffs_performance} suggest that we can extract representations of different dimensionalities from the same encoder to improve performance in desired settings.
Indeed, \cref{tab:different_zlayer} shows that we can improve few-shot performance by decreasing dimensionality.
Extracting smaller dimensional representations from the OpenCLIP model even achieves the best overall performance for 1\% 
 as seen in \cref{tab:best_results,tab:different_zlayer}.
 This explains why previous works, \eg \cite{caron_emerging_2021}, showed full-shot improvement when concatenating outputs of ViT blocks, namely, they were increasing the dimensionality.\looseness=-1

\subsubsection{Data and augmentations}
\label{sec:res:hparams:augmentations}

We now analyze the effect of the number of augmentations. 
We focus on multi-crops given that we have many pretrained models that only differ in this augmentation.

\begin{figure}[h]
\centering
\begin{subfigure}[t]{\linewidth}
\centering
\includegraphics[width=\linewidth]{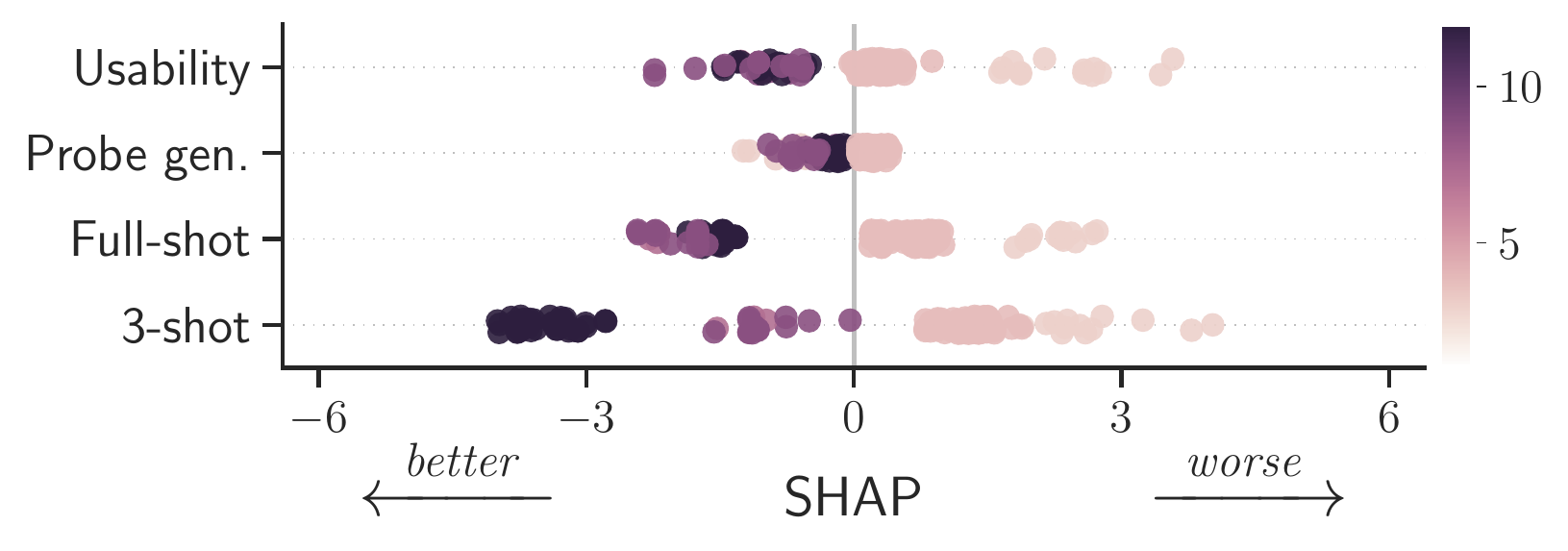}
\caption{SHAP}
\label{fig:nviews:main_shap}
\end{subfigure}
\begin{subfigure}[t]{\linewidth}
\centering
\includegraphics[width=\linewidth]{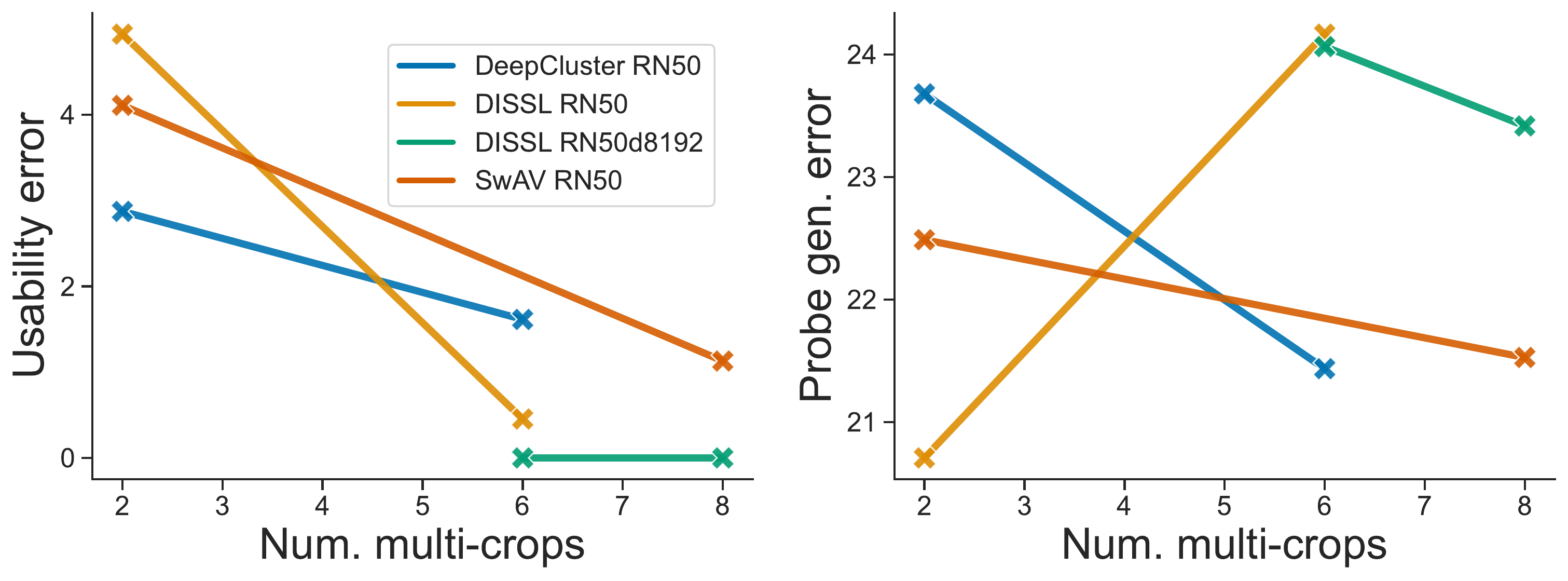}
\caption{controlled analysis}
\label{fig:nviews:main_controlled}
\end{subfigure}
\caption{
Effect of the number of multicrops on usability and probe generalization error, (a) when considering all models; and (b) when all other hyperparameters are constant.
}    
\vspace{-1em}
\end{figure}

\paragraph{Augmentations improve usability and probe gen}
A priori, one might think that using more augmentations improves generalization by acting as a regularizer.
\Cref{fig:nviews:main_shap} shows that increasing the number of multi-crops actually mostly improves usability --- although it can also help probe generalization. 
\Cref{fig:nviews:main_controlled} shows similar results when controlling for confounders.
Increasing the number of multi-crops thus overcomes the U-P tradeoff, which improves both full- and the few-shot performance (\cref{fig:nviews:main_shap}).
In \cref{appx:sec:res:augmentations} we show similar results for other augmentations.

Strengthening augmentations intuitively improves probe generalization by increasing the invariance of the SSL encoder, which will retain less information that probes can overfit to 
\citep{tsai_self-supervised_2021,tian_what_2020,federici_learning_2020,mitrovic_representation_2021,wu_on_2021,ruan_optimal_2022}.
The beneficial impact that augmentations have on usability is less obvious but has been suggested by \citet{dubois_improving_2022}.
Specifically, they prove that stronger augmentations decrease the number of potential tasks and thus the required capacity of probes.
Strengthening augmentations thus has a similar impact on usability as increasing the probe's capacity by increasing dimensionality (\cref{fig:dimensionality:main_shap}).

\paragraph{Additional pretraining data can worsen generalization} In \cref{appx:sec:res:augmentations} we show that pretraining on ImageNet-22K, instead of its subset
ImageNet-1K, worsens the encoder's and probe's generalization but can improve usability.\looseness=-1

\subsubsection{Architecture}
\label{sec:res:hparams:architecture}

We now analyze the impact of the encoder's architecture.

\begin{figure}[h]
\centering
\begin{subfigure}[t]{\linewidth}
\centering
\includegraphics[width=\linewidth]{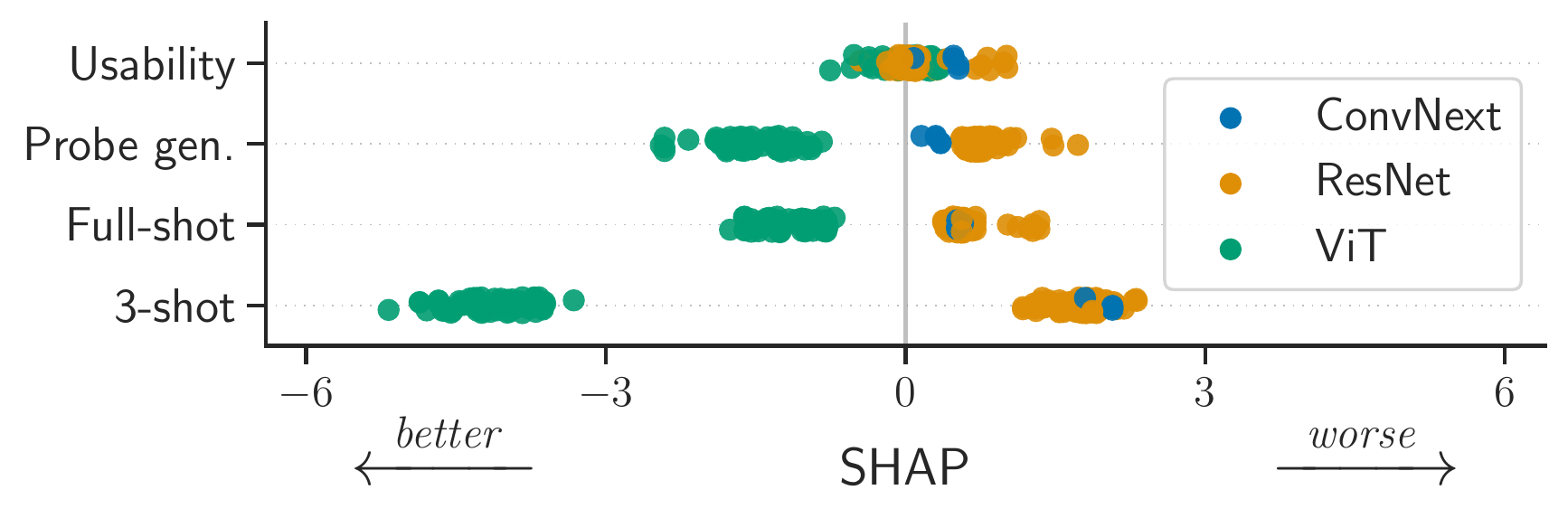}
\caption{Achitecture's family}
\label{fig:architecture:family_shap}
\end{subfigure}
\begin{subfigure}[t]{\linewidth}
\centering
\includegraphics[width=\linewidth]{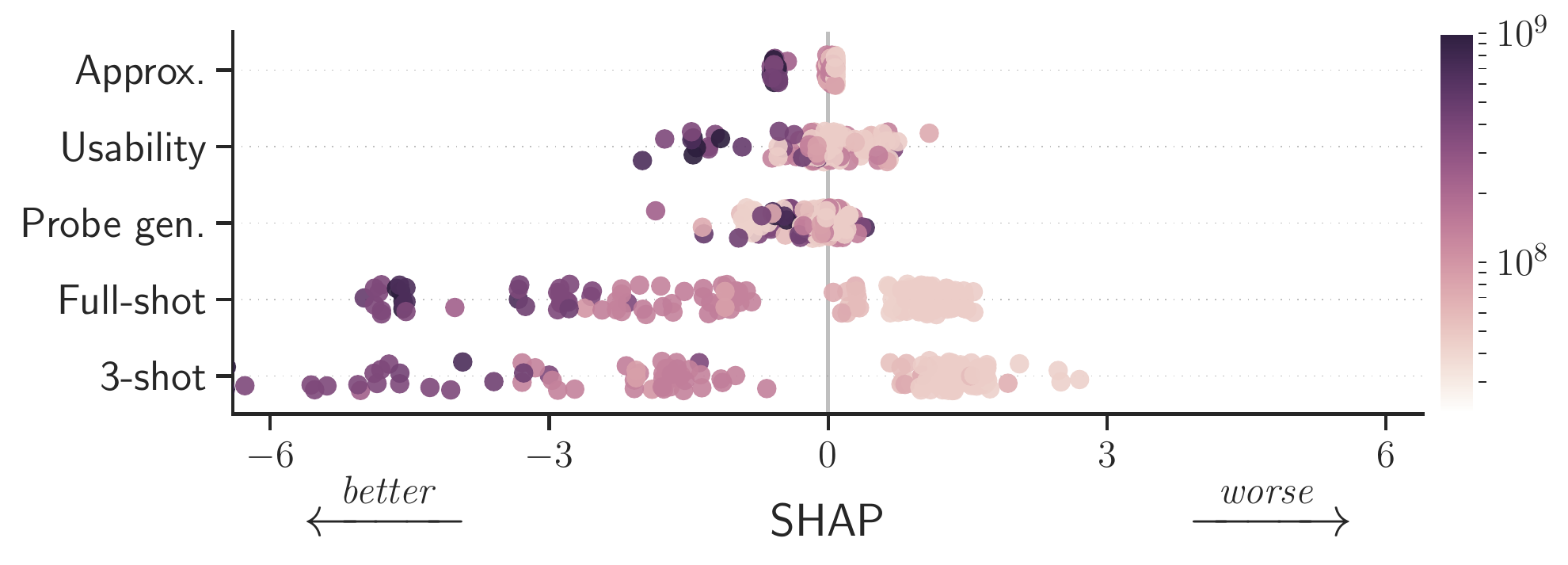}
\caption{Number of parameters}
\label{fig:architecture:size_shap}
\end{subfigure}
\caption{
Impact of the (a) architecture's family, and (b) number of parameters (color) on risk components and aggregated full- or few-shot risk.
Lower SHAP values (x-axis) are better as Y-axis are errors.
}    
\vspace{-1em}
\end{figure}

{}

\paragraph{ViTs improve probe generalization}
\Cref{fig:architecture:family_shap} shows that ViTs are significantly better than ResNets for probe generalization (GLA p-value = $9\sci{8}$) and do not worsen usability.
This thus translates to few- and full-shot improvements.

\paragraph{Larger encoders improve usability and approximation}
\Cref{fig:architecture:size_shap} shows that increasing the number of parameters improves the usability and approximation (GLA p-value = $4\sci{17}$), without impacting generalization.
Those gains improve full- and few-shot performance.
In \cref{appx:sec:res:architecture} we show that smaller ViT patch sizes lead to similar gains.

Now let us analyze the impact of projection heads in SSL, which are known to improve overall full-shot performance 
\cite{bachman_learning_2019,chen_simple_2020,chen_big_2020}.

\begin{figure}[h]
    \centering
    \includegraphics[width=1\linewidth]{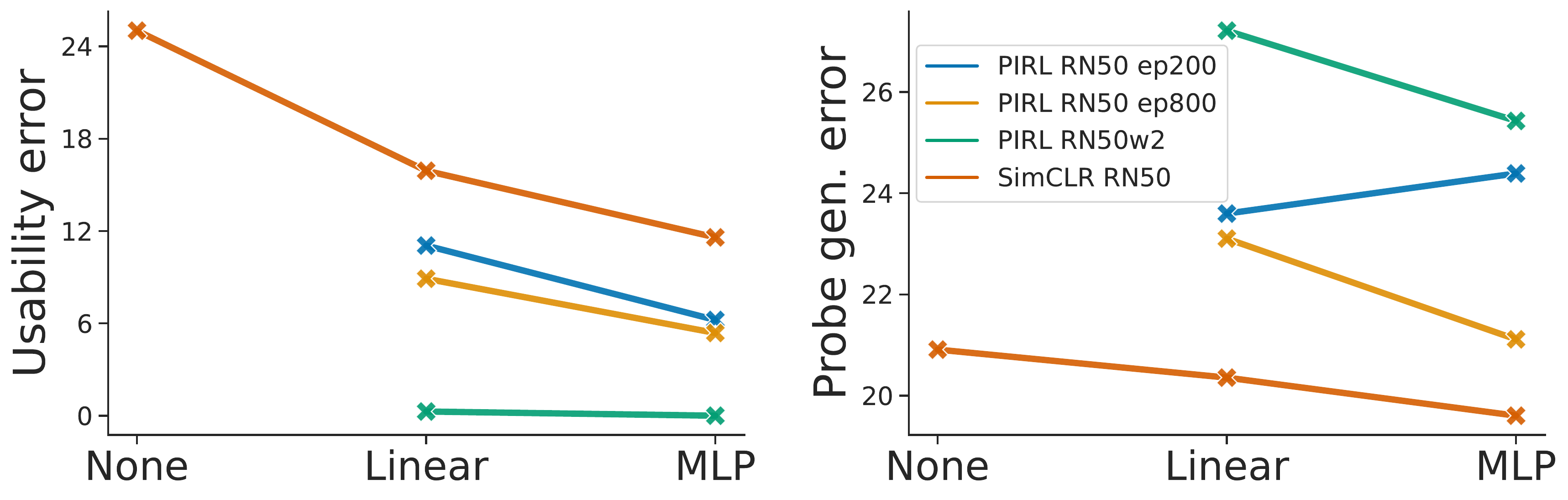}
    \caption{
    Effect of the projection head on usability and probe generalization error, when all other hyperparameters are kept the same. 
    Each color shows a specific model.
}    \label{fig:projection:controlled}
\end{figure}

\paragraph{Large projection heads improve usability}
\Cref{fig:projection:controlled} shows that MLP projections improve usability (CA p-value = $9\sci{12}$) and often also probe generalization.
In \cref{appx:sec:res:architecture}
we show that increasing the capacity (number of parameters) of an MLP projection head further improves usability.

Many works have tried to explain why projection heads improve SSL. 
For example, \citet{jing_understanding_2022} suggests that projections avoid dimensionality collapse.
In \cref{appx:sec:res:architecture}, we show that projection heads indeed improve effective dimensionality and thus usability (\cref{sec:res:hparams:dimensionality}) but that the increase in effective dimensionality is not larger for non-linear projection heads.
This suggests that we still do not completely understand the impact of non-linear projections.

\subsubsection{Objective}
\label{sec:res:hparams:objectives}

\begin{figure}[h]
    \centering
    \includegraphics[width=\linewidth]{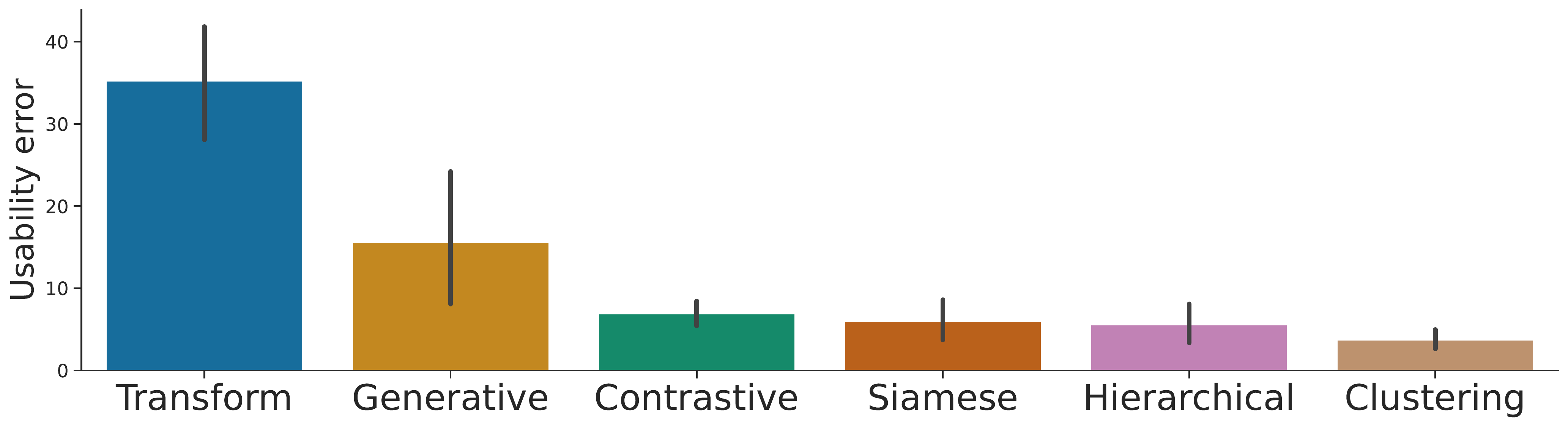}
    \caption{
    Impact of objective type on usability. 
Each bar shows the average usability error for all encoders pretrained with that type of SSL objective.
Type details in 
\cref{appx:sec:details:hparam_impact}.
%as described in \cref{appx:sec:details:hparam_impact}.
}    \label{fig:objective:usability_global}
\end{figure}

We now analyze the effect that the objective has on the representation. 
To simplify the analysis we aggregate all ($\Nobj$) objectives into 6 types 
(x-axis of \cref{fig:objective:usability_global}).

\paragraph{Generative and transformation-predicting objectives suffer from high usability error}
 \Cref{fig:objective:usability_global} shows that representations learned using objectives that are generative (\eg MAE or BEiT) or predict the data augmentation (\eg RotNet or LocNet) are less usable (GLA p-value = $3\sci{4}$). 
The other objectives give similar usability, with a slight edge for clustering objectives (\eg DISSL, DINO, or SwAV). 

The lack of usability explains why generative encoders such as MAE do not give a good linear probing performance, despite their strong fine-tuning performance \cite{he_masked_2022}. 
Intuitively, generative objectives preserve all information about the input but do not ensure that this information is usable by linear probes \cite{xu_theory_2020,dubois_learning_2020}.
%Indeed, if the decoder is powerful enough it can re-generate the input from non-linearly separable representations  \cite{xu_theory_2020,dubois_learning_2020}.
In comparison, contrastive objectives  ensure linear usability because they maximize dot-product similarity \cite{saunshi_theoretical_2019,tosh_contrastive_2021,haochen_provable_2021}. 
More generally, 
\citet{dubois_improving_2022} shows that many existing SSL losses explicitly optimize for usability. 
%That objective consists of linearly predicting the equivalence classes induced by data augmentations.
%That object
%Namely, linearly predicting the equivalence classes induced by the data augmentation.

\begin{figure}[h]
\centering
%\vspace*{-1em}
\includegraphics[width=\linewidth]{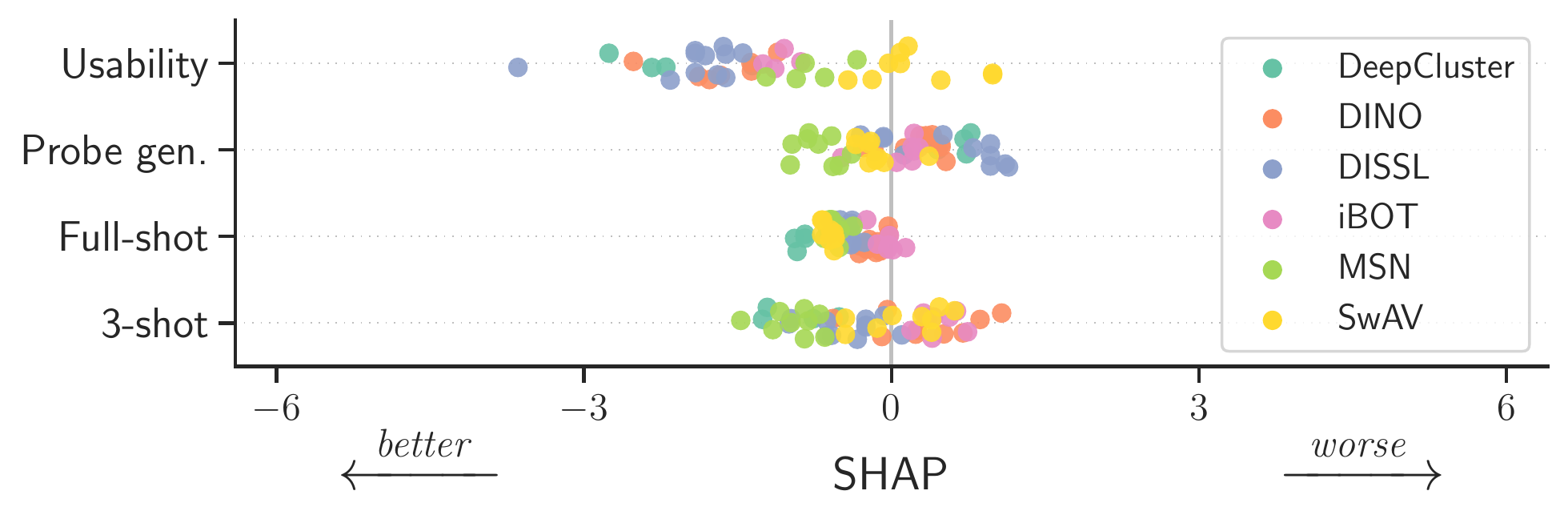}
%\vspace*{-2em}
\caption{Comparison between clustering objectives.
}    \label{fig:objective:main_clustering_objectives}
\end{figure}
\paragraph{The exact objective has little impact}
\Cref{fig:objective:main_clustering_objectives} compares different clustering objectives and shows that the impact of the exact objective is relatively minor. %compared to the other design choices we analyzed. 
For example, the impact on the aggregated risk is at most $1$ percentage point.
This suggests that one should choose a simple and easy-to-tune objective and focus on other components.
%This contrasts with what most of the SSL research community spends time on. 

% \subsubsection{Optimization}
% \label{sec:res:hparams:optimization}

% In \cref{appx:sec:res:optimization} we analyze the impact of optimization hyperparameters.
% Most notably, we show longer training generally improves usability and probe generalization but can worsen the encoder's generalization.
% Furthermore, we show that Adam(W) tends to perform better than SGD.
% In general, however, we found optimization hyperparameters to give less clear trends than other design choices.

% \subsubsection*{Acknowledgments}
% Use unnumbered third-level headings for the acknowledgments. All
% acknowledgments, including those to funding agencies, go at the end of the paper.
% Only add this information once your submission is accepted and deanonymized. 

\section{Related work}

\paragraph{Risk decomposition}
The estimation/approximation or the bias/variance decomposition has been very useful for practitioners and theoreticians to focus on specific risk components \cite{kohavi_bias_1996,pedro_unified_2000,valentini_bias_2004}.
Such decomposition has nevertheless rarely been extended beyond classical supervised learning.
Notable exceptions include \cite{wu_representation_2020} and \cite{zhou_adversarial_2022} in the context of domain adaptation and federated learning respectively.
To our knowledge, we are the first to provide an exact decomposition for SSL, but some theoretical works, \eg, \citet{bansal_for_2021}, have decomposed bounds on the risk (rather than the risk).

\paragraph{Benchmarking SSL}
One of our secondary contributions is a thorough benchmark of many SSL models (5 settings, \Nhparam{} design choices, \Nobj{} objective, and \Npre{} models).
There have been previous SSL benchmarks but those are either much smaller or use a different evaluation pipeline for each model.
For example, \citet{goyal_scaling_2019} provides a thorough but small benchmark (3 design choices and 2 objectives).
While \citet{goyal_vissl_2021} 
 and \citet{mmselfsup2021} evaluate more models (\Nvissl{} and \Nmmss{} respectively) but use different evaluation pipelines as their goal is to replicate previous work rather than to provide a fair benchmarking.

 \paragraph{Understanding SSL}
 There is a growing literature of work that tries to explain the effect of specific SSL design choices, \eg projections heads \cite{gupta_understanding_2022,appalaraju_towards_2020,jing_understanding_2022} or augmentations \cite{tsai_self-supervised_2021,tian_what_2020,federici_learning_2020,mitrovic_representation_2021,wu_on_2021,dubois_lossy_2021}, or provide a conceptual framework to think about design choices \cite{dubois_improving_2022}.
Sometimes those explanations agree with one another but other times they are orthogonal or even in contradiction.
Our work does not provide explanations but rather a new tool to empirically verify previous hypotheses and suggest new ones. 
For example, in \cref{sec:res:hparams} we highlight previous explanations that are supported by our empirical results.\looseness=-1

\section{Summary and outlook}

We present an SSL risk decomposition to provide a fine-grained understanding of the type of errors made by a linear probe predicting from SSL representations.
Our risk decomposition generalizes the supervised approximation/estimation decomposition by considering errors arising from the representation learning process.
We provide computationally efficient estimators for each risk component, akin to the training and validation errors in supervised learning.
Using those estimators, we analyze $\Npre$ pretrained SSL models and the effect of $\Nhparam$ design choices.
Our findings suggest that the two primary sources of errors are the usability of the representation, resulting from linear separability issues, and the probe's generalization error, due to finite training data.
Furthermore, we show that there is often a tradeoff between these two sources of errors, which translates into a performance tradeoff between few- and full-shot probing.
Some design choices, such as the dimensionality of the representation and the SSL objective, can control this tradeoff and thus improve performance in certain settings at the expense of others. 
Meanwhile, other choices, such as the use of large projection heads and ViT encoders, overcome the tradeoff and thus improve performance in all settings.%\looseness=-1

Our risk decomposition and in particular our estimators have limitations that should be addressed to improve their applicability. 
Most notably, they require the probe's training data to be a subset of the encoder's pretraining data, limiting their application in common out-of-distribution settings.
We hope that our findings will inspire further research in this direction, and, more generally, the use of risk decompositions for analyzing sources of errors in machine learning.

\section*{Acknowledgements}
We thank Rohan Taori, Niladri Chatterji, Shibani Santurkar, Ananya Kumar for helpful feedback.
YD is supported by a Knights-Hennessy Scholarship.
The work is supported by an Open Philanthropy Project Award.

\bibliography{bib_tatsu}

\newcommand*{\annalstat}{Annals of Statistics} \newcommand*{\jasa}{Journal of
  the American Statistical Association (JASA)} \newcommand*{\tacl}{Transactions
  of the Association for Computational Linguistics (TACL)}
  \newcommand*{\coling}{International Conference on Computational Linguistics
  (COLING)} \newcommand*{\acl}{Association for Computational Linguistics (ACL)}
  \newcommand*{\naacl}{North American Association for Computational Linguistics
  (NAACL)} \newcommand*{\aclijcnlp}{Association for Computational Linguistics
  and International Joint Conference on Natural Language Processing
  (ACL-IJCNLP)} \newcommand*{\emnlpconll}{Empirical Methods in Natural Language
  Processing and Computational Natural Language Learning (EMNLP/CoNLL)}
  \newcommand*{\emnlpijnlp}{Empirical Methods in Natural Language Processing
  and International Joint Conference on Natural Language Processing
  (EMNLP-IJCNLP)} \newcommand*{\emnlp}{Empirical Methods in Natural Language
  Processing} \newcommand*{\hltnaacl}{Human Language Technology and North
  American Association for Computational Linguistics (HLT/NAACL)}
  \newcommand*{\eacl}{European Association for Computational Linguistics
  (EACL)}  \newcommand*{\icml}{International Conference on Machine Learning
  (ICML)} \newcommand*{\neurips}{Advances in Neural Information Processing
  Systems (NeurIPS)} \newcommand*{\nips}{Advances in Neural Information
  Processing Systems} \newcommand*{\iclr}{International Conference on Learning
  Representations (ICLR)} \newcommand*{\iclrworkshop}{International Conference
  on Learning Representations Workshop (ICLR)} \newcommand*{\jmlr}{Journal of
  Machine Learning Research (JMLR)} \newcommand*{\fatml}{Conference on
  Fairness, Accountability, and Transparency} \newcommand*{\aistats}{Artificial
  Intelligence and Statistics (AISTATS)} \newcommand*{\cvpr}{Conference on
  Computer Vision and Pattern Recognition (CVPR)}
  \newcommand*{\iccv}{International Conference on Computer Vision (ICCV)}
  \newcommand*{\icpr}{International Conference on Pattern Recognition (ICPR)}
  \newcommand*{\eccv}{European Conference on Computer Vision (ECCV)}
  \newcommand*{\uai}{Conference on Uncertainty in Artificial Intelligence
  (UAI)}  \newcommand*{\ecai}{European Conference on Artificial Intelligence}
  \newcommand*{\aaai}{AAAI Conference on Artificial Intelligence}
  \newcommand*{\packdd}{Pacific-Asia Conference on Knowledge Discovery and Data
  Mining} \newcommand*{\kdd}{International Conference on Knowledge Discovery
  and Data Mining (KDD)} \newcommand*{\neurcom}{Neural Computation}
  \newcommand*{\msml}{Mathematical and Scientific Machine Learning Conference
  (MSML)} \newcommand*{\ijcnn}{International Joint Conference on Neural
  Networks (IJCNN)} \newcommand*{\ieeesigproc}{IEEE Transactions on Signal
  Processing} \newcommand*{\ieeeec}{IEEE Transactions on Electronic Computers}
  \newcommand*{\procieee}{Proceedings of the IEEE}
  \newcommand*{\pnas}{Proceedings of the National Academy of Sciences}
  \newcommand*{\chiconf}{Conference on Human Factors in Computing Systems
  (CHI)} \newcommand*{\ieeecp}{IEEE Symposium on Security and Privacy (SP)}
  \newcommand*{\stoc}{Symposium on Theory of Computing (STOC)}
  \newcommand*{\pods}{Symposium on Principles of Database Systems (PODS)}
  \newcommand*{\colt}{Conference on Learning Theory (COLT)}
  \newcommand*{\www}{The World Wide Web Conference (WWW)}
  \newcommand*{\soda}{Symposium on Discrete Algorithms (SODA)}
  \newcommand*{\focs}{Symposium on Foundations of Computer Science (FOCS)}
  \newcommand*{\acm}{Communications of the Association for Computing Machinery
  (ACM)} \newcommand*{\ieeeaccess}{IEEE Access}
  \newcommand*{\ijcv}{International Journal of Computer Vision (IJCV)}
  \newcommand*{\ieeetpami}{IEEE Transactions on Pattern Analysis and Machine
  Intelligence} \newcommand*{\ieeetit}{IEEE Transactions on Information Theory}
  \newcommand*{\alt}{Conference on Algorithmic Learning Theory (ALT)}
  \newcommand*{\cocoon}{International Computing and Combinatorics Conference
  (COCOON)} \newcommand*{\arxiv}[1]{arXiv preprint arXiv:#1}
\begin{thebibliography}{94}
\providecommand{\natexlab}[1]{#1}
\providecommand{\url}[1]{\texttt{#1}}
\expandafter\ifx\csname urlstyle\endcsname\relax
  \providecommand{\doi}[1]{doi: #1}\else
  \providecommand{\doi}{doi: \begingroup \urlstyle{rm}\Url}\fi

\bibitem[Appalaraju et~al.(2020)Appalaraju, Zhu, Xie, and
  Feh{\'e}rv{\'a}ri]{appalaraju_towards_2020}
Appalaraju, S., Zhu, Y., Xie, Y., and Feh{\'e}rv{\'a}ri, I.
\newblock Towards good practices in self-supervised representation learning.
\newblock \emph{\arxiv{2012.00868}}, 2020.

\bibitem[Asano et~al.(2020)Asano, Rupprecht, and
  Vedaldi]{asano_self-labelling_2020}
Asano, Y.~M., Rupprecht, C., and Vedaldi, A.
\newblock Self-labelling via simultaneous clustering and representation
  learning.
\newblock In \emph{\iclr}, 2020.

\bibitem[Assran et~al.(2022)Assran, Caron, Misra, Bojanowski, Bordes, Vincent,
  Joulin, R., and Ballas]{assran_masked_2022}
Assran, M., Caron, M., Misra, I., Bojanowski, P., Bordes, F., Vincent, P.,
  Joulin, A., R., M., and Ballas, N.
\newblock Masked siamese networks for label-efficient learning.
\newblock In \emph{\eccv}, 2022.

\bibitem[Bachman et~al.(2019)Bachman, Hjelm, and
  Buchwalter]{bachman_learning_2019}
Bachman, P., Hjelm, R.~D., and Buchwalter, W.
\newblock Learning representations by maximizing mutual information across
  views.
\newblock In \emph{\neurips}, 2019.

\bibitem[Bansal et~al.(2021)Bansal, Kaplun, and Barak]{bansal_for_2021}
Bansal, Y., Kaplun, G., and Barak, B.
\newblock For self-supervised learning, rationality implies generalization,
  provably.
\newblock In \emph{\iclr}, 2021.

\bibitem[Bao et~al.(2022)Bao, Dong, Piao, and Wei]{bao_beit_2022}
Bao, H., Dong, L., Piao, S., and Wei, F.
\newblock Beit: {BERT} pre-training of image transformers.
\newblock In \emph{\iclr}, 2022.

\bibitem[Bardes et~al.(2022{\natexlab{a}})Bardes, Ponce, and
  LeCun]{bardes_vicreg_2022}
Bardes, A., Ponce, J., and LeCun, Y.
\newblock {VICR}eg: Variance-invariance-covariance regularization for
  self-supervised learning.
\newblock In \emph{\iclr}, 2022{\natexlab{a}}.

\bibitem[Bardes et~al.(2022{\natexlab{b}})Bardes, Ponce, and
  LeCun]{bardes_vicregl_2022}
Bardes, A., Ponce, J., and LeCun, Y.
\newblock {VICR}egl: Self-supervised learning of local visual features.
\newblock In \emph{\neurips}, 2022{\natexlab{b}}.

\bibitem[Barron(1994)]{barron_approximation_1994}
Barron, A.~R.
\newblock Approximation and estimation bounds for artificial neural networks.
\newblock \emph{Machine Learning}, 14:\penalty0 115--133, 1994.

\bibitem[Belkin et~al.(2019)Belkin, Hsu, Ma, and
  Mandal]{belkin_reconciling_2019}
Belkin, M., Hsu, D., Ma, S., and Mandal, S.
\newblock Reconciling modern machine-learning practice and the classical
  bias--variance trade-off.
\newblock \emph{\pnas}, 116:\penalty0 15849--15854, 2019.

\bibitem[Bergstra et~al.(2011)Bergstra, Bardenet, Bengio, and
  K{\'e}gl]{bergstra2011algorithms}
Bergstra, J., Bardenet, R., Bengio, Y., and K{\'e}gl, B.
\newblock Algorithms for hyper-parameter optimization.
\newblock \emph{Advances in neural information processing systems}, 24, 2011.

\bibitem[Bottou \& Bousquet(2007)Bottou and Bousquet]{bottou_tradeoffs_2007}
Bottou, L. and Bousquet, O.
\newblock The tradeoffs of large scale learning.
\newblock In \emph{\neurips}, 2007.

\bibitem[Bousquet et~al.(2022)Bousquet, Daniely, Kaplan, Mansour, Moran, and
  Stemmer]{bousquet_monotone_2022}
Bousquet, O.~J., Daniely, A., Kaplan, H., Mansour, Y., Moran, S., and Stemmer,
  U.
\newblock Monotone learning.
\newblock In \emph{\colt}, 2022.

\bibitem[Caron et~al.(2018)Caron, Bojanowski, Joulin, and
  Douze]{caron_deep_2018}
Caron, M., Bojanowski, P., Joulin, A., and Douze, M.
\newblock Deep clustering for unsupervised learning of visual features.
\newblock In \emph{\eccv}, 2018.

\bibitem[Caron et~al.(2020)Caron, Misra, Mairal, Goyal, Bojanowski, and
  Joulin]{caron_unsupervised_2020}
Caron, M., Misra, I., Mairal, J., Goyal, P., Bojanowski, P., and Joulin, A.
\newblock Unsupervised learning of visual features by contrasting cluster
  assignments.
\newblock In \emph{\neurips}, 2020.

\bibitem[Caron et~al.(2021)Caron, Touvron, Misra, J{'e}gou, Mairal, Bojanowski,
  and Joulin]{caron_emerging_2021}
Caron, M., Touvron, H., Misra, I., J{'e}gou, H., Mairal, J., Bojanowski, P.,
  and Joulin, A.
\newblock Emerging properties in self-supervised vision transformers.
\newblock In \emph{\iccv}, 2021.

\bibitem[Chen et~al.(2021{\natexlab{a}})Chen, Gan, Li, Guo, Chen, Gao, Chung,
  Xu, Zeng, Lu, Li, Carin, and Tao]{chen_simpler_2021}
Chen, J., Gan, Z., Li, X., Guo, Q., Chen, L., Gao, S., Chung, T., Xu, Y., Zeng,
  B., Lu, W., Li, F., Carin, L., and Tao, C.
\newblock Simpler, faster, stronger: Breaking the log-k curse on contrastive
  learners with flatnce.
\newblock \emph{\arxiv{2107.01152}}, 2021{\natexlab{a}}.

\bibitem[Chen \& Guestrin(2016)Chen and Guestrin]{chen_xgboost_2016}
Chen, T. and Guestrin, C.
\newblock Xgboost: A scalable tree boosting system.
\newblock In \emph{SIGKDD}, pp.\  785--794, 2016.

\bibitem[Chen et~al.(2020{\natexlab{a}})Chen, Kornblith, Norouzi, and
  Hinton]{chen_simple_2020}
Chen, T., Kornblith, S., Norouzi, M., and Hinton, G.
\newblock A simple framework for contrastive learning of visual
  representations.
\newblock In \emph{\icml}, 2020{\natexlab{a}}.

\bibitem[Chen et~al.(2020{\natexlab{b}})Chen, Kornblith, Swersky, Norouzi, and
  Hinton]{chen_big_2020}
Chen, T., Kornblith, S., Swersky, K., Norouzi, M., and Hinton, G.
\newblock Big self-supervised models are strong semi-supervised learners.
\newblock In \emph{\neurips}, 2020{\natexlab{b}}.

\bibitem[Chen et~al.(2020{\natexlab{c}})Chen, Fan, Girshick, and
  He]{chen_improved_2020}
Chen, X., Fan, H., Girshick, R., and He, K.
\newblock Improved baselines with momentum contrastive learning.
\newblock \emph{\arxiv{2003.04297}}, 2020{\natexlab{c}}.

\bibitem[Chen et~al.(2021{\natexlab{b}})Chen, Xie, and He]{chen_empirical_2021}
Chen, X., Xie, S., and He, K.
\newblock An empirical study of training self-supervised vision transformers.
\newblock In \emph{\iccv}, 2021{\natexlab{b}}.

\bibitem[Cherti et~al.(2022)Cherti, Beaumont, Wightman, Wortsman, Ilharco,
  Gordon, Schuhmann, Schmidt, and Jitsev]{cherti_reproducible_2022}
Cherti, M., Beaumont, R., Wightman, R., Wortsman, M., Ilharco, G., Gordon, C.,
  Schuhmann, C., Schmidt, L., and Jitsev, J.
\newblock Reproducible scaling laws for contrastive language-image learning.
\newblock \emph{\arxiv{2212.07143}}, 2022.

\bibitem[Dar et~al.(2021)Dar, Muthukumar, and Baraniuk]{dar_farewell_2021}
Dar, Y., Muthukumar, V., and Baraniuk, R.~G.
\newblock A farewell to the bias-variance tradeoff? an overview of the theory
  of overparameterized machine learning.
\newblock \emph{\arxiv{2109.02355}}, 2021.

\bibitem[Devlin et~al.(2019)Devlin, Chang, Lee, and
  Toutanova]{devlin_bert_2019}
Devlin, J., Chang, M., Lee, K., and Toutanova, K.
\newblock {BERT}: Pre-training of deep bidirectional transformers for language
  understanding.
\newblock In \emph{\naacl}, pp.\  4171--4186, Minneapolis, Minnesota, June
  2019. Association for Computational Linguistics.

\bibitem[Doersch et~al.(2015)Doersch, Gupta, and
  Efros]{doersch_unsupervised_2015}
Doersch, C., Gupta, A., and Efros, A.
\newblock Unsupervised visual representation learning by context prediction.
\newblock In \emph{\iccv}, 2015.

\bibitem[Domingos(2000)]{pedro_unified_2000}
Domingos, P.
\newblock A unified bias-variance decomposition and its applications.
\newblock In \emph{\icml}, 2000.

\bibitem[Dosovitskiy et~al.(2021)Dosovitskiy, Beyer, Kolesnikov, Weissenborn,
  Zhai, Unterthiner, Dehghani, Minderer, Heigold, Gelly, Uszkoreit, and
  Houlsby]{dosovitskiy_image_2021}
Dosovitskiy, A., Beyer, L., Kolesnikov, A., Weissenborn, D., Zhai, X.,
  Unterthiner, T., Dehghani, M., Minderer, M., Heigold, G., Gelly, S.,
  Uszkoreit, J., and Houlsby, N.
\newblock An image is worth 16x16 words: Transformers for image recognition at
  scale.
\newblock In \emph{\iclr}, 2021.

\bibitem[Dubois et~al.(2020)Dubois, Kiela, Schwab, and
  Vedantam]{dubois_learning_2020}
Dubois, Y., Kiela, D., Schwab, D.~J., and Vedantam, R.
\newblock Learning optimal representations with the decodable information
  bottleneck.
\newblock In \emph{\neurips}, 2020.

\bibitem[Dubois et~al.(2021)Dubois, Bloem-Reddy, Ullrich, and
  Maddison]{dubois_lossy_2021}
Dubois, Y., Bloem-Reddy, B., Ullrich, K., and Maddison, C.~J.
\newblock Lossy compression for lossless prediction.
\newblock \emph{\neurips}, 2021.

\bibitem[Dubois et~al.(2022)Dubois, Hashimoto, Ermon, and
  Liang]{dubois_improving_2022}
Dubois, Y., Hashimoto, T., Ermon, S., and Liang, P.
\newblock Improving self-supervised learning by characterizing idealized
  representations.
\newblock In \emph{\neurips}, 2022.

\bibitem[Ericsson et~al.(2021)Ericsson, Gouk, and
  Hospedales]{ericsson_why_2021}
Ericsson, L., Gouk, H., and Hospedales, T.~M.
\newblock Why do self-supervised models transfer? investigating the impact of
  invariance on downstream tasks.
\newblock \emph{\arxiv{abs/2111.11398}}, 2021.

\bibitem[Federici et~al.(2020)Federici, Dutta, Forr{\'{e}}, Kushman, and
  Akata]{federici_learning_2020}
Federici, M., Dutta, A., Forr{\'{e}}, P., Kushman, N., and Akata, Z.
\newblock Learning robust representations via multi-view information
  bottleneck.
\newblock In \emph{\iclr}, 2020.

\bibitem[Foster et~al.(2021)Foster, Pukdee, and
  Rainforth]{foster_improving_2021}
Foster, A., Pukdee, R., and Rainforth, T.
\newblock Improving transformation invariance in contrastive representation
  learning.
\newblock In \emph{\iclr}, 2021.

\bibitem[Geman et~al.(1992)Geman, Bienenstock, and Doursat]{geman_neural_1992}
Geman, S., Bienenstock, E., and Doursat, R.
\newblock Neural networks and the bias/variance dilemma.
\newblock \emph{Neural computation}, 4\penalty0 (1):\penalty0 1--58, 1992.

\bibitem[Gidaris et~al.(2018)Gidaris, Singh, and
  Komodakis]{gidaris_unsupervised_2018}
Gidaris, S., Singh, P., and Komodakis, N.
\newblock Unsupervised visual representation learning by context prediction.
\newblock In \emph{\iclr}, 2018.

\bibitem[Goyal et~al.(2019)Goyal, Mahajan, Gupta, and
  Misra]{goyal_scaling_2019}
Goyal, P., Mahajan, D., Gupta, A., and Misra, I.
\newblock Scaling and benchmarking self-supervised visual representation
  learning.
\newblock In \emph{\iccv}, 2019.

\bibitem[Goyal et~al.(2021)Goyal, Duval, Reizenstein, Leavitt, Xu, Lefaudeux,
  Singh, Reis, Caron, Bojanowski, Joulin, and Misra]{goyal_vissl_2021}
Goyal, P., Duval, Q., Reizenstein, J., Leavitt, M., Xu, M., Lefaudeux, B.,
  Singh, M., Reis, V., Caron, M., Bojanowski, P., Joulin, A., and Misra, I.
\newblock {VISSL}, 2021.

\bibitem[Grill et~al.(2020)Grill, Strub, Altch{\'{e}}, Tallec, Richemond,
  Buchatskaya, Doersch, Pires, Guo, Azar, Piot, Kavukcuoglu, Munos, and
  Valko]{grill_bootstrap_2020}
Grill, J.~B., Strub, F., Altch{\'{e}}, F., Tallec, C., Richemond, P.~H.,
  Buchatskaya, E., Doersch, C., Pires, B.~A., Guo, Z., Azar, M.~G., Piot, B.,
  Kavukcuoglu, K., Munos, R., and Valko, M.
\newblock Bootstrap {Your} {Own} {Latent} - a new approach to self-supervised
  learning.
\newblock In \emph{\neurips}, 2020.

\bibitem[Gupta et~al.(2022)Gupta, Ajanthan, Hengel, and
  Gould]{gupta_understanding_2022}
Gupta, K., Ajanthan, T., Hengel, A. v.~d., and Gould, S.
\newblock Understanding and improving the role of projection head in
  self-supervised learning.
\newblock \emph{arXiv preprint arXiv:2212.11491}, 2022.

\bibitem[HaoChen et~al.(2021)HaoChen, Wei, Gaidon, and
  Ma]{haochen_provable_2021}
HaoChen, J.~Z., Wei, C., Gaidon, A., and Ma, T.
\newblock Provable guarantees for self-supervised deep learning with spectral
  contrastive loss.
\newblock In \emph{\neurips}, 2021.

\bibitem[He et~al.(2016)He, Zhang, Ren, and Sun]{he_deep_2016}
He, K., Zhang, X., Ren, S., and Sun, J.
\newblock Deep residual learning for image recognition.
\newblock In \emph{\cvpr}, 2016.

\bibitem[He et~al.(2020)He, Fan, Wu, Xie, and Girshick]{he_momentum_2020}
He, K., Fan, H., Wu, Y., Xie, S., and Girshick, R.
\newblock Momentum contrast for unsupervised visual representation learning.
\newblock In \emph{\cvpr}, 2020.

\bibitem[He et~al.(2022)He, Chen, Xie, Li, Doll{\'{a}}r, and
  Girshick]{he_masked_2022}
He, K., Chen, X., Xie, S., Li, Y., Doll{\'{a}}r, P., and Girshick, R.~B.
\newblock Masked autoencoders are scalable vision learners.
\newblock In \emph{\cvpr}, 2022.

\bibitem[Hua et~al.(2021)Hua, Wang, Xue, Ren, Wang, and Zhao]{hua_on_2021}
Hua, T., Wang, W., Xue, Z., Ren, S., Wang, Y., and Zhao, H.
\newblock On feature decorrelation in self-supervised learning.
\newblock In \emph{\iccv}, 2021.

\bibitem[Jing et~al.(2022)Jing, Vincent, LeCun, and
  Tian]{jing_understanding_2022}
Jing, L., Vincent, P., LeCun, Y., and Tian, Y.
\newblock Understanding dimensional collapse in contrastive self-supervised
  learning.
\newblock In \emph{\iclr}, 2022.

\bibitem[Kaplan et~al.(2020)Kaplan, McCandlish, Henighan, Brown, Chess, Child,
  Gray, Radford, Wu, and Amodei]{kaplan_scaling_2020}
Kaplan, J., McCandlish, S., Henighan, T., Brown, T., Chess, B., Child, R.,
  Gray, S., Radford, A., Wu, J., and Amodei, D.
\newblock Scaling laws for neural language models.
\newblock \emph{arXiv preprint arXiv:2001.08361}, 2020.

\bibitem[Kohavi \& Wolpert(1996)Kohavi and Wolpert]{kohavi_bias_1996}
Kohavi, R. and Wolpert, D.~H.
\newblock Bias plus variance decomposition for zero-one loss functions.
\newblock In \emph{\icml}, 1996.

\bibitem[Lundberg \& Lee(2017)Lundberg and Lee]{lundberg_unified_2017}
Lundberg, S.~M. and Lee, S.
\newblock A unified approach to interpreting model predictions.
\newblock In \emph{\neurips}, 2017.

\bibitem[Miao et~al.(2022)Miao, Mathieu, Dubois, Rainforth, Teh, Foster, and
  Kim]{miao_learning_2022}
Miao, N., Mathieu, E., Dubois, Y., Rainforth, T., Teh, Y.~W., Foster, A., and
  Kim, H.
\newblock Instance-specific augmentation: Capturing local invariances.
\newblock \emph{\arxiv{2206.00051}}, 2022.

\bibitem[Misra \& van~der Maaten(2020)Misra and van~der
  Maaten]{misra_self-supervised_2020}
Misra, I. and van~der Maaten, L.
\newblock Self-supervised learning of pretext-invariant representations.
\newblock In \emph{\cvpr}, 2020.

\bibitem[Mitrovic et~al.(2021)Mitrovic, McWilliams, Walker, Buesing, and
  Blundell]{mitrovic_representation_2021}
Mitrovic, J., McWilliams, B., Walker, J., Buesing, L., and Blundell, C.
\newblock Representation learning via invariant causal mechanisms.
\newblock In \emph{\iclr}, 2021.

\bibitem[MMSelfSup(2021)]{mmselfsup2021}
MMSelfSup.
\newblock {MMSelfSup}: Openmmlab self-supervised learning toolbox and
  benchmark.
\newblock \url{https://github.com/open-mmlab/mmselfsup}, 2021.

\bibitem[Mukherjee et~al.(2006)Mukherjee, Niyogi, Poggio, and
  Rifkin]{mukherjee_learning_2006}
Mukherjee, S., Niyogi, P., Poggio, T., and Rifkin, R.
\newblock Learning theory: stability is sufficient for generalization and
  necessary and sufficient for consistency of empirical risk minimization.
\newblock \emph{Advances in Computational Mathematics}, 25:\penalty0 161--193,
  2006.

\bibitem[Nakkiran et~al.(2020)Nakkiran, Kaplun, Bansal, Yang, Barak, and
  Sutskever]{nakkiran_deep_2020}
Nakkiran, P., Kaplun, G., Bansal, Y., Yang, T., Barak, B., and Sutskever, I.
\newblock Deep double descent: Where bigger models and more data hurt.
\newblock In \emph{\iclr}, 2020.

\bibitem[Neal(2019)]{neal_on_2019}
Neal, B.
\newblock On the bias-variance tradeoff: Textbooks need an update.
\newblock \emph{\arxiv{1912.08286}}, 2019.

\bibitem[Neal et~al.(2018)Neal, Mittal, Baratin, Tantia, Scicluna,
  Lacoste{-}Julien, and Mitliagkas]{neal_modern_2018}
Neal, B., Mittal, S., Baratin, A., Tantia, V., Scicluna, M., Lacoste{-}Julien,
  S., and Mitliagkas, I.
\newblock A modern take on the bias-variance tradeoff in neural networks.
\newblock \emph{\arxiv{1810.08591}}, 2018.

\bibitem[Neyshabur et~al.(2015)Neyshabur, Tomioka, and
  Srebro]{neyshabur_in_2015}
Neyshabur, B., Tomioka, R., and Srebro, N.
\newblock In search of the real inductive bias: On the role of implicit
  regularization in deep learning.
\newblock In \emph{iclrworkshop}, 2015.

\bibitem[Noroozi \& Favaro(2016)Noroozi and Favaro]{noroozi_unsupervised_2016}
Noroozi, M. and Favaro, P.
\newblock Unsupervised learning of visual representations by solving jigsaw
  puzzles.
\newblock In \emph{\eccv}, 2016.

\bibitem[Paszke et~al.(2019)Paszke, Gross, Massa, Lerer, Bradbury, Chanan,
  Killeen, Lin, Gimelshein, Antiga, Desmaison, Kopf, Yang, DeVito, Raison,
  Tejani, Chilamkurthy, Steiner, Fang, Bai, and Chintala]{paszke_pytorch_2019}
Paszke, A., Gross, S., Massa, F., Lerer, A., Bradbury, J., Chanan, G., Killeen,
  T., Lin, Z., Gimelshein, N., Antiga, L., Desmaison, A., Kopf, A., Yang, E.,
  DeVito, Z., Raison, M., Tejani, A., Chilamkurthy, S., Steiner, B., Fang, L.,
  Bai, J., and Chintala, S.
\newblock Pytorch: An imperative style, high-performance deep learning library.
\newblock In \emph{\neurips}, 2019.

\bibitem[Pedregosa et~al.(2011)Pedregosa, Varoquaux, Gramfort, Michel, Thirion,
  Grisel, Blondel, Prettenhofer, Weiss, Dubourg, Vanderplas, Passos,
  Cournapeau, Brucher, Perrot, and Duchesnay]{pedregosa_scikit-learn_2011}
Pedregosa, F., Varoquaux, G., Gramfort, A., Michel, V., Thirion, B., Grisel,
  O., Blondel, M., Prettenhofer, P., Weiss, R., Dubourg, V., Vanderplas, J.,
  Passos, A., Cournapeau, D., Brucher, M., Perrot, M., and Duchesnay, E.
\newblock Scikit-learn: Machine learning in {P}ython.
\newblock \emph{Journal of Machine Learning Research (JMLR)}, 12, 2011.

\bibitem[Radford et~al.(2021)Radford, Kim, Hallacy, Ramesh, Goh, Agarwal,
  Sastry, Askell, Mishkin, Clark, Krueger, and
  Sutskever]{radford_learning_2021}
Radford, A., Kim, J.~W., Hallacy, C., Ramesh, A., Goh, G., Agarwal, S., Sastry,
  G., Askell, A., Mishkin, P., Clark, J., Krueger, G., and Sutskever, I.
\newblock Learning transferable visual models from natural language
  supervision.
\newblock In \emph{\icml}, 2021.

\bibitem[Rosenfeld et~al.(2020)Rosenfeld, Rosenfeld, and
  Belinkov]{rosenfeld_constructive_2020}
Rosenfeld, J., Rosenfeld, A., and Belinkov, Y.
\newblock A constructive prediction of the generalization error across scales.
\newblock In \emph{\iclr}, 2020.

\bibitem[Rosenfeld(2021)]{rosenfeld_scaling_2021}
Rosenfeld, J.~S.
\newblock \emph{Scaling laws for deep learning}.
\newblock PhD thesis, Massachusetts Institute of Technology, 2021.

\bibitem[Ruan et~al.(2022)Ruan, Dubois, and Maddison]{ruan_optimal_2022}
Ruan, Y., Dubois, Y., and Maddison, C.~J.
\newblock Optimal representations for covariate shift.
\newblock In \emph{\iclr}, 2022.

\bibitem[Santurkar et~al.(2022)Santurkar, Dubois, Taori, Liang, and
  Hashimoto]{santurkar2022caption}
Santurkar, S., Dubois, Y., Taori, R., Liang, P., and Hashimoto, T.
\newblock Is a caption worth a thousand images? a controlled study for
  representation learning.
\newblock \emph{arXiv preprint arXiv:2207.07635}, 2022.

\bibitem[Saunshi et~al.(2019)Saunshi, Plevrakis, Arora, Khodak, and
  Khandeparkar]{saunshi_theoretical_2019}
Saunshi, N., Plevrakis, O., Arora, S., Khodak, M., and Khandeparkar, H.
\newblock A theoretical analysis of contrastive unsupervised representation
  learning.
\newblock In \emph{\icml}, 2019.

\bibitem[Saunshi et~al.(2022)Saunshi, Ash, Goel, Misra, Zhang, Arora, Kakade,
  and Krishnamurthy]{saunshi_understanding_2022}
Saunshi, N., Ash, J.~T., Goel, S., Misra, D., Zhang, C., Arora, S., Kakade,
  S.~M., and Krishnamurthy, A.
\newblock Understanding contrastive learning requires incorporating inductive
  biases.
\newblock In \emph{\icml}, 2022.

\bibitem[Shalev-Shwartz \& Ben-David(2014)Shalev-Shwartz and
  Ben-David]{shalevshwartz_understanding_2014}
Shalev-Shwartz, S. and Ben-David, S.
\newblock \emph{Understanding Machine Learning: From Theory to Algorithms}.
\newblock Cambridge University Press, 2014.

\bibitem[Tian et~al.(2020{\natexlab{a}})Tian, Krishnan, and
  Isola]{tian_contrastive_2020}
Tian, Y., Krishnan, D., and Isola, P.
\newblock Contrastive multiview coding.
\newblock In \emph{\eccv}, 2020{\natexlab{a}}.

\bibitem[Tian et~al.(2020{\natexlab{b}})Tian, Sun, Poole, Krishnan, Schmid, and
  Isola]{tian_what_2020}
Tian, Y., Sun, C., Poole, B., Krishnan, D., Schmid, C., and Isola, P.
\newblock What makes for good views for contrastive learning?
\newblock In \emph{\neurips}, 2020{\natexlab{b}}.

\bibitem[Tosh et~al.(2021)Tosh, Krishnamurthy, and Hsu]{tosh_contrastive_2021}
Tosh, C., Krishnamurthy, A., and Hsu, D.
\newblock Contrastive learning, multi-view redundancy, and linear models.
\newblock In \emph{\alt}, 2021.

\bibitem[Tsai et~al.(2021)Tsai, Wu, Salakhutdinov, and
  Morency]{tsai_self-supervised_2021}
Tsai, Y.~H., Wu, Y., Salakhutdinov, R.~R., and Morency, L.
\newblock Self-supervised learning from a multi-view perspective.
\newblock In \emph{\iclr}, 2021.

\bibitem[Valentini \& Dietterich(2004)Valentini and
  Dietterich]{valentini_bias_2004}
Valentini, G. and Dietterich, T.~G.
\newblock Bias-variance analysis of support vector machines for the development
  of svm-based ensemble methods.
\newblock \emph{\jmlr}, 5:\penalty0 725--775, 2004.

\bibitem[van~den Oord et~al.(2019)van~den Oord, Li, and
  Vinyals]{oord_representation_2019}
van~den Oord, A., Li, Y., and Vinyals, O.
\newblock Representation learning with contrastive predictive coding.
\newblock \emph{\arxiv{2110.02796}}, 2019.

\bibitem[Vapnik(2000)]{vapnik_nature_2000}
Vapnik, V.~N.
\newblock \emph{The Nature of Statistical Learning Theory}.
\newblock Springer-Verlag, 2000.

\bibitem[Vapnik \& Chervonenkis(1971)Vapnik and Chervonenkis]{vapnik_on_1971}
Vapnik, V.~N. and Chervonenkis, A.~Y.
\newblock On uniform convergence of the frequencies of events to their
  probabilities.
\newblock \emph{Teoriya Veroyatnostei i ee Primeneniya}, 16\penalty0
  (2):\penalty0 264--279, 1971.

\bibitem[Viering et~al.(2019)Viering, Mey, and Loog]{viering_open_2017}
Viering, T., Mey, A., and Loog, M.
\newblock Open problem: Monotonicity of learning.
\newblock In \emph{\colt}, 2019.

\bibitem[Wang \& Liu(2021)Wang and Liu]{wang_understanding_2021}
Wang, F. and Liu, H.
\newblock Understanding the behaviour of contrastive loss.
\newblock In \emph{\cvpr}, 2021.

\bibitem[Wang \& Isola(2020)Wang and Isola]{wang_understanding_2020}
Wang, T. and Isola, P.
\newblock Understanding contrastive representation learning through alignment
  and uniformity on the hypersphere.
\newblock In \emph{\icml}, 2020.

\bibitem[Wang et~al.(2021)Wang, Zhang, Shen, Kong, and Li]{wang_dense_2021}
Wang, X., Zhang, R., Shen, C., Kong, T., and Li, L.
\newblock Dense contrastive learning for self-supervised visual pre-training.
\newblock In \emph{\cvpr}, 2021.

\bibitem[Wang et~al.(2022)Wang, Zhang, Wang, Yang, and Lin]{wang_chaos_2022}
Wang, Y., Zhang, Y., Wang, Y., Yang, J., and Lin, Z.
\newblock Chaos is a ladder: A new theoretical understanding of contrastive
  learning via augmentation overlap.
\newblock In \emph{\iclr}, 2022.

\bibitem[Wu et~al.(2021)Wu, Zhuang, Mosse, Yamins, and Goodman]{wu_on_2021}
Wu, M., Zhuang, C., Mosse, M., Yamins, D. L.~K., and Goodman, N.~D.
\newblock On mutual information in contrastive learning for visual
  representations.
\newblock \emph{\arxiv{2005.13149}}, 2021.

\bibitem[Wu et~al.(2020)Wu, Guo, Chen, Liang, Jha, and
  Chalasani]{wu_representation_2020}
Wu, X., Guo, Y., Chen, J., Liang, Y., Jha, S., and Chalasani, P.
\newblock Representation bayesian risk decompositions and multi-source domain
  adaptation.
\newblock \emph{\arxiv{2004.10390}}, 2020.

\bibitem[Wu et~al.(2018)Wu, Xiong, Yu, and Lin]{wu_unsupervised_2018}
Wu, Z., Xiong, Y., Yu, S.~X., and Lin, D.
\newblock Unsupervised feature learning via non-parametric instance
  discrimination.
\newblock In \emph{\cvpr}, 2018.

\bibitem[Xu et~al.(2020)Xu, Zhao, Song, Stewart, and Ermon]{xu_theory_2020}
Xu, Y., Zhao, S., Song, J., Stewart, R., and Ermon, S.
\newblock A theory of usable information under computational constraints.
\newblock In \emph{\iclr}, 2020.

\bibitem[Yan et~al.(2020)Yan, Misra, Gupta, Ghadiyaram, and
  Mahajan]{yan_clusterfit_2020}
Yan, X., Misra, I., Gupta, A., Ghadiyaram, D., and Mahajan, D.
\newblock {ClusterFit}: Improving generalization of visual representations.
\newblock In \emph{\cvpr}, 2020.

\bibitem[Yang et~al.(2020)Yang, Yu, You, Steinhardt, and
  Ma]{yang_rethinking_2020}
Yang, Z., Yu, Y., You, C., Steinhardt, J., and Ma, Y.
\newblock Rethinking bias-variance trade-off for generalization of neural
  networks.
\newblock In \emph{\icml}, 2020.

\bibitem[Zbontar et~al.(2021)Zbontar, Jing, Misra, LeCun, and
  Deny]{zbontar_barlow_2021}
Zbontar, J., Jing, L., Misra, I., LeCun, Y., and Deny, S.
\newblock Barlow {Twins}: Self-supervised learning via redundancy reduction.
\newblock In \emph{\icml}, 2021.

\bibitem[Zhan et~al.(2020)Zhan, Xie, Liu, Ong, and Loy]{zhan_online_2020}
Zhan, X., Xie, J., Liu, Z., Ong, Y.-S., and Loy, C.~C.
\newblock Online deep clustering for unsupervised representation learning.
\newblock In \emph{\cvpr}, 2020.

\bibitem[Zhiliang et~al.(2022)Zhiliang, Li, Bao, Ye, and
  Wei]{zhiliang_beit_2022}
Zhiliang, P., Li, D., Bao, H., Ye, Q., and Wei, F.
\newblock Beit v2: Masked image modeling with vector-quantized visual
  tokenizers.
\newblock \emph{\arxiv{2208.06366}}, 2022.

\bibitem[Zhou et~al.(2021)Zhou, Wei, Wang, Shen, Xie, Yuille, and
  Kong]{zhou_ibot_2021}
Zhou, J., Wei, C., Wang, H., Shen, W., Xie, C., Yuille, A.~L., and Kong, T.
\newblock ibot: Image {BERT} pre-training with online tokenizer.
\newblock \emph{\arxiv{2111.07832}}, 2021.

\bibitem[Zhou et~al.(2022{\natexlab{a}})Zhou, Zhou, Si, Yu, Ng, and
  Yan]{zhou_mugs_2022}
Zhou, P., Zhou, Y., Si, C., Yu, W., Ng, T.~K., and Yan, S.
\newblock Mugs: {A} multi-granular self-supervised learning framework.
\newblock \emph{\arxiv{2203.14415}}, 2022{\natexlab{a}}.

\bibitem[Zhou et~al.(2022{\natexlab{b}})Zhou, Wu, Wang, and
  He]{zhou_adversarial_2022}
Zhou, Y., Wu, J., Wang, H., and He, J.
\newblock Adversarial robustness through bias variance decomposition: A new
  perspective for federated learning.
\newblock In \emph{Conference on Information and Knowledge Management (CIKM)},
  2022{\natexlab{b}}.

\end{thebibliography}
\bibliographystyle{icml2022}

%%%%%%%%%%%%%%%%%%%%%%%%%%%%%%%%%%%%%%%%%%%%%%%%%%%%%%%%%%%%%%%%%%%%%%%%%%%%%%%
%%%%%%%%%%%%%%%%%%%%%%%%%%%%%%%%%%%%%%%%%%%%%%%%%%%%%%%%%%%%%%%%%%%%%%%%%%%%%%%
% APPENDIX
%%%%%%%%%%%%%%%%%%%%%%%%%%%%%%%%%%%%%%%%%%%%%%%%%%%%%%%%%%%%%%%%%%%%%%%%%%%%%%%
%%%%%%%%%%%%%%%%%%%%%%%%%%%%%%%%%%%%%%%%%%%%%%%%%%%%%%%%%%%%%%%%%%%%%%%%%%%%%%%
\newpage
\appendix
\onecolumn

%%%%%%%%%%%%%%%%%%%%%%%%%%%%%%%%%%%%%%%%%%%%%%%%%%%%%%%%%%%%

% \addcontentsline{toc}{section}{Appendix} % Add the appendix text to the document TOC
% \part{Appendix} % Start the appendix part

% \parttoc % Insert the appendix TOC
% %\tableofcontents

% \clearpage
% \newpage

\section{Risk decompositions}
\label{appx:sec:risk_dec}
\subsection{Supervised decomposition}
\label{appcs:sec:risk_dec:supervised}

The goal of supervised learning is to predict targets $Y$ from inputs $X$ sampled from a distribution $\ptask{}(X,Y)$.
The predictor is selected from a desired functional family $\Qx{} \subseteq \{f : \Xc \to \actspace{} \}$ by an algorithm $ \algsup{} : \Pc{\Xc, \Yc} \to \Qx$. 
%that takes as input an input-label distribution.
For example, empirical risk minimization (ERM) maps the empirical distribution $\hp{S}(X,Y)$ of a training set $\Stask \iidsim \ptask{}$ to risk minimizer $\predS{} \defeq \algsup{}(\hp{S}) \in \Qx$.
The selected predictor $\predS{}$ is then evaluated using the risk $\Lp{\predS{}} \defeq \Ep{\ptask{}}{\ell(Y, \predS{}(X))}$ with respect to a desired evaluation loss $\ell$, \eg,  0-1 loss for classification error.
Let us denote the best possible predictor in the desired functional family as $\predF \in \argmin_{f \in \Qx{}} \Lp{f}$, the Bayes (irreducible) risk by $\brisk{} \defeq \min_{f: \Xc \to \Yc } \Lp{f}$, and the $\hp{S}$-empirical risk of any predictor $f$  by $\hLp{f; \hp{\Stask}}$.\footnote{
For notational convenience, we assume throughout the paper that minimizers are achievable and algorithms are deterministic.
}
For conciseness, we use subscripts to denote the risk $\rF \defeq \Lp{\predF{}}$ and $\rS \defeq \Lp{\predS{}}$.

The risk $\rS{}$ of the selected predictor is ultimately the value that we care about. 
But when designing, empirically evaluating, and theoretically analyzing a model, it is often helpful to understand the types of errors made by $\predS{}$.  
For example, it is useful to monitor both the generalization gap and the training error to know which pipeline component to improve (regularization, architecture, etc). 
This can be formalized by the standard excess risk decomposition \cite{barron_approximation_1994}:
%\begin{small}
\begin{equation}\label{eq:supervised_risk_decomposition}
\underbrace{\vphantom{\int } \rS - \brisk{}}_{\vphantom{\int x^2} \text{excess risk}}
 =  
 \underbrace{\vphantom{\int } \rS{} - \rF{}}_{\vphantom{\int x^2}\color{mygreen} \text{estimation error}} + 
 \underbrace{\vphantom{\int } \rF{} - \brisk{},}_{\vphantom{\int x^2} \color{myred}  \text{approximation error}}   
\end{equation}
%\end{small}
where the approximation error measures the error due to searching over a constrained family $\Qz$ and the estimation error quantifies the impact of using finite samples and a non-optimal learning algorithm.
Typically, the algorithm is universally consistent so the estimation error does not depend on the algorithm because the predictor $\predA{} = \algsup{}(\ptask{})$ chosen on the population distribution is the best in the family $\rF{} = \rA$, where $\rA \defeq \Lp{\predA{}}$.
If this is not the case, one can further separate estimation error between generalization ($\rS - \rA$) and algorithmic error ($\rA - \rF$).

\begin{equation}\label{eq:supervised_risk_decomposition_further}
\underbrace{\vphantom{\int } \rS - \brisk{}}_{\vphantom{\int x^2} \text{excess risk}}
 =  
  \underbrace{\vphantom{\int } \rS{} - \rA{}}_{\vphantom{\int x^2}\color{mygreen} \text{generalization error}} + 
    \underbrace{\vphantom{\int } \rA{} - \rF{}}_{\vphantom{\int x^2}\color{mybrown} \text{algorithmic error}} + 
 \underbrace{\vphantom{\int } \rF{} - \brisk{}.}_{\vphantom{\int x^2} \color{myred}  \text{approximation error}}
\end{equation}
To derive the decomposition we order the expected risk of predictors $\Ep{S}{\rS} \geq \rA \geq \rF 
\geq \brisk{}$
%predictors by their expected risk $\predS{} \leq \predA{} \leq \predF{}$ 
and write the excess risk as a telescoping sum.
 By construction, the resulting error components are thus non-negative in expectation.
 The ordering holds if the algorithm trained on the population data learns a predictor that is at least as good than on any finite samples $S$, \eg, if the algorithm is a monotonic \cite{shalevshwartz_understanding_2014,viering_open_2017,bousquet_monotone_2022}.
Note that the decomposition could be further expanded by considering other potential sources of errors such as optimization errors.

\subsection{Alternative decompositions for representation learning}
\label{appcs:sec:risk_dec:alternatives}

\begin{figure}[h]
    \centering
    \includegraphics[width=0.5\linewidth]{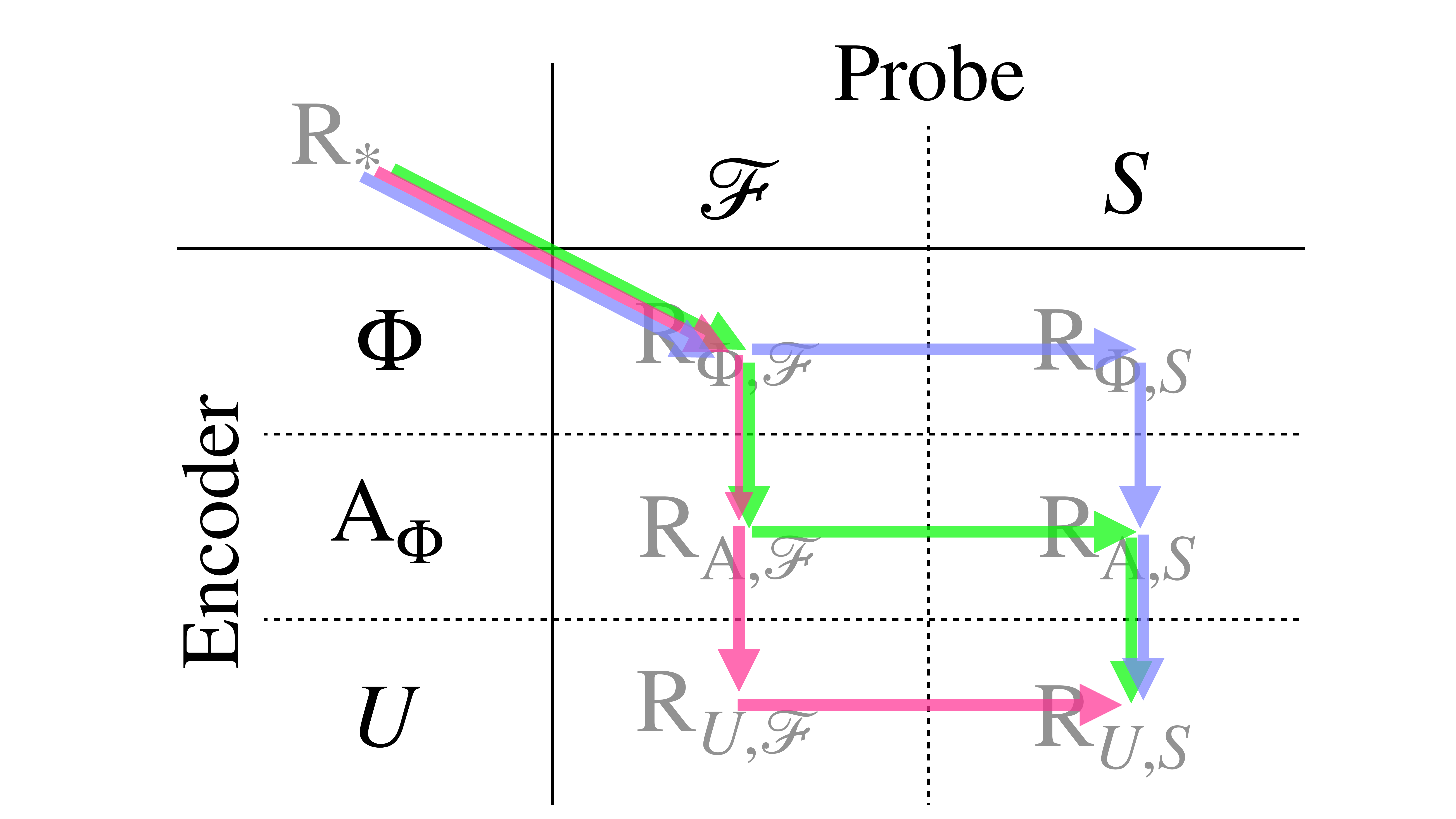}
    \caption{Illustration of the possible loss decompositions corresponding to different ways of traversing the encoder/probe training matrix.
    In green we see our proposed decomposition, in purple the generalization errors are switched, 
    in pink the usability and probe's generalization are switched.
    }
    \label{fig:matrix}
\end{figure}

In the main paper, we saw one possible excess risk decomposition for representation learning. 
This decomposition is not unique, and we now briefly discuss other possible decompositions.
To understand those, it is important to ask ourselves what are the properties of a good risk decomposition.
We consider three specific properties, namely, each risk component should ideally:
\begin{inlinelist}
\item be positive;
\item highlight important representation learning errors; and
\item have an efficient estimator.
\end{inlinelist}

For positivity to hold in expectation, one simply has to find a sequence of predictors that are ordered by expected risk and then write the final excess risk as a telescoping sum by adding and subtracting respective risks in order.
For representation learning, we consider three potential sources of errors $(\Spre,\algrep,\Phi)$: the functional family $\Phi$ (\eg ResNet50), the SSL algorithm $\algrep$ (\eg SimCLR optimized with SGD), and the training set (\eg ImageNet training).
For the supervised probe, we essentially have the same choices $(\Stask,\algsup,\Qz)$, but follow the standard supervised excess risk and remove the algorithm choice as it is typically universally consistent.
Altogether we have 3 choices for the encoder and 2 for the probe, which can be represented as the matrix $(\Spre,\algrep,\Phi) \times (\Stask,\Qz)$.
The question then becomes what ordered sequence to use, \ie, which path to take to traverse the matrix as seen in \cref{fig:matrix}.
 We thus have the three following possible (positive) decompositions.

\begin{equation}\label{appcs:eq:representation_risk_decomposition}
\underbrace{\vphantom{\int } \rUS - \brisk}_{\vphantom{\int x^2} \text{excess risk}}
 =  
 \underbrace{ \vphantom{\int } \rUS - \rAS}_{\vphantom{\int x^2} \color{myblue} \text{encoder generalization}} + 
  \underbrace{ \vphantom{\int } \rAS - \rAF}_{\vphantom{\int x^2} \color{mygreen} 
 \text{probe generalization}} + 
  \underbrace{ \vphantom{\int }   \rAF - \rFF}_{\vphantom{\int x^2}\color{myorange} \text{representation usability}} + 
 \underbrace{  \vphantom{\int } \rFF - \brisk}_{\vphantom{\int x^2}\color{myred} \text{ approximation}} 
\end{equation}

\paragraph{Our decomposition} 
First, there is \cref{appcs:eq:representation_risk_decomposition} (\textcolor{mylime}{green path} in \cref{fig:matrix}), which is the decomposition whose interpretation we discuss extensively in \cref{sec:excess_risk_representation}.
The only difference here is that we use start the path from the Bayes Risk $\brisk{}$ instead of zero.
We are thus decomposing the excess risk instead of the total risk, as is common in supervised learning (see \cref{appcs:sec:risk_dec:supervised}).
As discussed in \cref{sec:practical_estimation}, each of our risk components admits practical estimators.
Our risk decomposition thus satisfies our three desired properties (positivity, highlight representation learning errrors, and estimation).

\begin{equation}\label{appcs:eq:dec_switch_gen_error}
\underbrace{\vphantom{\int } \rUS - \brisk{}}_{\vphantom{\int x^2} \text{excess risk}}
 =  
 \underbrace{ \vphantom{\int } \rUS - \rUF}_{\vphantom{\int x^2}\color{mygreen} \txtcirc{1}} + 
  \underbrace{ \vphantom{\int } \rUF - \rAF}_{\vphantom{\int x^2}\color{myblue} \txtcirc{2}} + 
  \underbrace{ \vphantom{\int } \rAF - \rFF}_{\vphantom{\int x^2}\color{myorange} \text{representation usability}} + 
 \underbrace{  \vphantom{\int } \rFF - \brisk{}}_{\vphantom{\int x^2}\color{myred} \text{approximation}} 
\end{equation}

\paragraph{Switching generalization errors}
Another possible decomposition is \cref{appcs:eq:dec_switch_gen_error} (\textcolor{mypurple}{purple path} in \cref{fig:matrix}), which replaces $\rAS$ with $\rUF$.
Looking more carefully at $\txtcirc{1}$ and $\txtcirc{2}$ we see that both risk components have a similar interpretation as in \cref{appcs:eq:representation_risk_decomposition}; they are generalization errors.
The difference is that it first considers the generalization errors of the predictor $\txtcirc{1}$ and then that of the encoder $\txtcirc{2}$.
The choice is thus arbitrary in terms of highlighting important representation learning errors. 
The reason we favored the other decomposition (\cref{appcs:eq:representation_risk_decomposition}) is due to estimation. 
Indeed, the natural estimator for $\rUF$ would be to train and evaluate the probe on the test set $\Ste$ so that only the probe has to generalize, \ie, $ \hrUF{} \defeq \min_{f \in \Qz} \hLp{\encA{} \circ f; \hp{\Ste}}$.
The problem here is that $\Ste$ is relatively small ($50$K for ImageNet) and so the $\hrUF{}$ would greatly underestimate $\rUF$ as the probe can overfit $\Ste$. 
In contrast, $\hrAS$ is a better estimator as it trains a probe on the much larger $\Stask \setminus \Ssub$.
In \cref{appx:sec:results:alternative} we use this second decomposition and show that it would make little impact on our experimental results, despite the worse estimator.
This is reassuring as it suggests that our interpretation is robust to the choice of decomposition.

\begin{equation}\label{appcs:eq:dec_switch_usability}
\underbrace{\vphantom{\int } \rUS - \brisk{}}_{\vphantom{\int x^2} \text{excess risk}}
 =  
 \underbrace{ \vphantom{\int } \rUS - \rAS}_{\vphantom{\int x^2}\color{myblue} \text{encoder generalization}} + 
  \underbrace{ \vphantom{\int } \rAS - \rFS}_{\vphantom{\int x^2}\color{myorange} \txtcirc{3}} + 
  \underbrace{ \vphantom{\int } \rFS - \rFF}_{\vphantom{\int x^2}\color{mygreen} \txtcirc{4}} + 
 \underbrace{  \vphantom{\int } \rFF - \brisk{}}_{\vphantom{\int x^2}\color{myred} \text{approximation}} 
\end{equation}

\paragraph{Switching representation usability and probe generalization error}
The second possible decomposition is
\cref{appcs:eq:dec_switch_usability} (\textcolor{mypink}{pink path} in \cref{fig:matrix}), which replaces $\rAF$ with $\rFS$.
As a result, the representation usability$\txtcirc{3}$ would be considered before the probe generalization $\txtcirc{4}$.
The main downside is that the encoder generalization error does not depend on the pretraining algorithm $\algrep$ and so one would not be able to quantify how much the representation helps downstream sample efficiency.
In other words, given that we want to understand representation learning, we would like to have as many terms as possible that  depend on the representations.
\Cref{appcs:eq:dec_switch_usability}  does not highlight/distinguish between important representation learning errors as the probe generalization error does not consider the effect of representations.

\subsection{Alternative representation of our decomposition }
\label{appcs:sec:risk_dec:illustration}
\begin{figure}[h!]
    \centering
    \vspace{-1em}
    \includegraphics[width=0.55\linewidth]{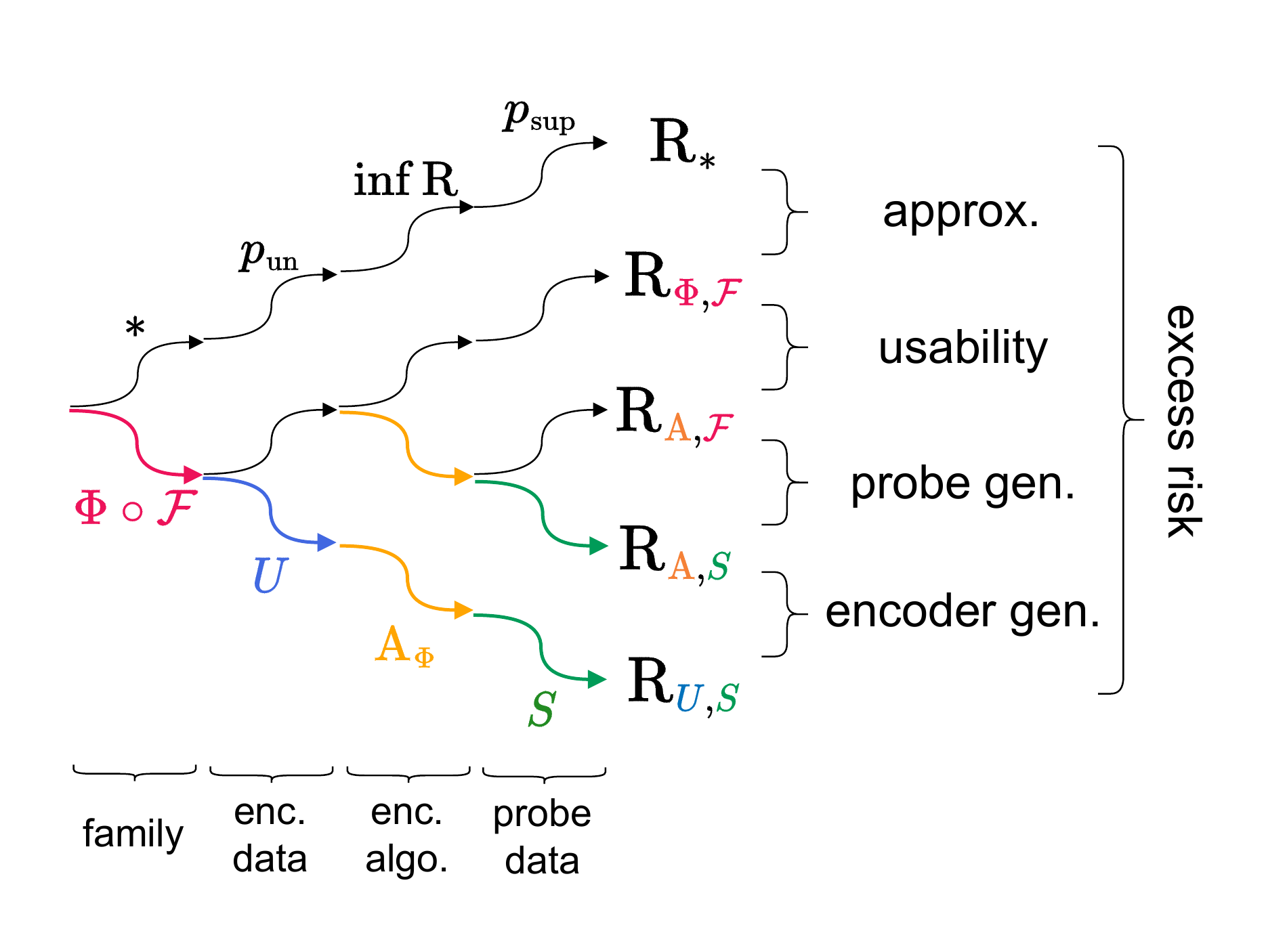}
    \vspace{-1em}
    \caption{
    Our excess risk decomposition consists of the difference between risks in settings of increasing difficulty (going down).
    In particular, we consider 4 potential approximations: 
\begin{inlinelist}
 \item constrained functional families $\color{OrangeRed}\Phi\circ\mathcal{F}$ instead of unconstrained $*$;
 \item finite pretraining data $\color{RoyalBlue}U$ instead of the population pretraining distribution $\ppre$; 
\item non-optimal representation learning algorithm ${\color{Orange}\mathrm{A}_{\small \Phi}}$ instead of end-to-end risk minimization $\inf \mathrm{R}$; 
\item finite training data $\color{ForestGreen}S$ instead of the population training distribution $\ptask$.
\end{inlinelist}}
    \label{fig:loss_dec_schema}
\end{figure}

In the main paper and in \cref{appcs:sec:risk_dec:alternatives} we illustrate our risk decomposition as the path in the $(\Spre,\algrep,\Phi) \times (\Stask,\Qz)$ matrix.
Another potentially useful illustration is \cref{fig:loss_dec_schema}, which shows that our excess risk decomposition consists of the difference between risks in settings of increasing difficulty (more approximation).
Changing the order in which we consider different approximations give rise to alternative decompositions (\cref{appcs:sec:risk_dec:alternatives}).

\subsection{Relationship with the supervised decomposition}
\label{appcs:sec:risk_dec:recovering}

A natural question to ask is how does our risk decomposition for representation learning relates the standard supervised decomposition.
The answer is that the former trivially generalizes the latter.
In particular, if we define the family of predictors in a supervised setting as the family of composed encoders and probes $\Phi \circ \Qz \defeq \{ \enc \circ f \cond \enc \in \Phi, f \in \Qz \}$ and the new supervised algorithm $\algcirc$ as a two step algorithm that first fits the encoder using $\algrep$ (after dropping labels) and then fits the probe with the desired supervised algorithm $\algsup{}$, then we have the following equivalences between risk components.
On the left we show representation learning components and on the right we show supervised learning components.
\begin{alignat}{3}
&\underbrace{\vphantom{\int } \rUS - \brisk{}}_{\vphantom{\int x^2} \text{rep. excess risk}}
 &{}={}& \Lp{(\phi\circ f)_{\scriptscriptstyle S}} - \brisk{} 
  &{}={}& \underbrace{\vphantom{\int } \rS - \brisk{}}_{\vphantom{\int x^2} \text{sup. excess risk}}  \\
 &\underbrace{ \vphantom{\int } \rUS - \rAS}_{\vphantom{\int x^2}\color{myblue} \text{encoder generalization}} + 
  \underbrace{ \vphantom{\int } \rAS - \rAA}_{\vphantom{\int x^2}\color{mygreen} \text{probe generalization}} 
 &{}={}& \rUS - \rAA
 =   \Lp{(\phi\circ f)_{\scriptscriptstyle S}} - \Lp{(\phi\circ f)_{\scriptscriptstyle A}} 
  &{}={}&  \underbrace{\vphantom{\int } \rS - \rA }_{\vphantom{\int x^2}\color{mygreen} \text{sup. generalization error}} \label{appx:eq:recovering:probe_gen_error}\\
  &\underbrace{\vphantom{\int x^2} \rAA - \rAF }_{\vphantom{\int x^2}\color{mybrown} \text{probe sup. algorithm}} + 
  \underbrace{\vphantom{\int x^2} \rAF - \rFF}_{\vphantom{\int x^2}\color{myorange} \text{representation usability}} 
 &{}={}&  \rAA - \rFF
 = \Lp{(\phi\circ f)_{\scriptscriptstyle A}} - \Lp{(\phi\circ f)_{\scriptscriptstyle \Phi \circ \mathcal{F}}}
 &{}={}&\underbrace{\vphantom{\int } \rA - \rCirc}_{\vphantom{\int x^2}\color{mybrown} \text{sup. algorithmic error}} \label{appx:eq:recovering:probe_alg_error} \\
 &\underbrace{  \vphantom{\int } \rFF - \brisk{}}_{\vphantom{\int x^2}\color{myred} \text{approximation}} 
 &{}={}& \Lp{(\phi\circ f)_{\scriptscriptstyle \Phi \circ \mathcal{F}}} - \brisk{}  
 &{}={}&  \underbrace{\vphantom{\int } \rCirc - \brisk{}}_{\vphantom{\int x^2}\color{myred} \text{sup. approximation error}} \label{appx:eq:recovering:approx_error} 
 \end{alignat}
In \cref{appx:eq:recovering:probe_alg_error},  we introduced the probes supervised algorithmic error, which is natural when recovering the standard risk decomposition with an algorithmic error. 
As discussed in \cref{appcs:sec:risk_dec:alternatives} we typically drop this term as it is zero if the supervised algorithm is universally consistent (\eg ERM) in which case $\rAA = \rAF$ so the probe's generalization recovers in \cref{appx:eq:recovering:probe_gen_error} recovers the definition from the main paper.

We thus see that our risk decomposition recovers the standard supervised decomposition and is a natural extension of it.
Note that in the case when we use identity encoders $\Phi$ then the encoder's generalization and representation usability become zero.
Then, as we would expect, the probe generalization, probe sup. algorithm, and approximation error respectively recover the sup. generalization, sup. algorithmic  and sup. approximation error from \cref{appcs:sec:risk_dec:supervised}.

\subsection{Tradeoffs}
\label{appcs:sec:risk_dec:tradeoffs}

One of the advantages of using the standard supervised risk decomposition is that it highlights a potential tradeoff between estimation and approximation error \cite{bottou_tradeoffs_2007,shalevshwartz_understanding_2014}.
Such a conceptual tradeoff can be very useful to train and develop supervised models, \eg, when using larger models it is often useful to increase the training data or regularization.
In the following we discuss three such tradeoffs in representation learning that directly arise from the standard estimation-approximation tradeoff.
But first, let us briefly remind that the standard tradeoff (and by extension our tradeoffs) is a conceptual framework rather than a universal theorem.

\paragraph{The approximation-estimation and related tradeoffs are not universal}
Although the approximation-estimation tradeoff (or the related bias-vias tradeoff) is typically stated as a universal fact that arises from the decomposition, this is not actually the case. 
There are usually three arguments given to support those intuitive tradeoffs.
The first common argument is the risk decomposition.
For example, \citet{shalevshwartz_understanding_2014} state after providing the decomposition that ``these two [approximation and estimation] terms imply a tradeoff between choosing a more complex [hypothesis class] $\mathcal{H}$''.
But this is only true assuming that the total aggregated risk is constant.
An other common argument for the tradeoff is typically given by theoretical bounds on each term. 
The issue with those bounds is that they typically consider (upper bounds) on the worst-case scenario for constrained predictors rather than what actually happens in practice.
In fact, recent theoretical work have argued that this tradeoff does not hold in the over-parameterized regime \cite{yang_rethinking_2020,dar_farewell_2021}.
Finally, the trade-off is often supported using empirical evidence.
This is for example done by \citet{geman_neural_1992}, which is typically cited when discussing such tradeoff.
But the empirical evidence does not universally support such tradeoff.
In fact, there is growing empirical evidence that increasing the size of some models (\eg neural networks) can improve both the approximation and the estimation error \cite{neyshabur_in_2015,belkin_reconciling_2019,nakkiran_deep_2020}.
For a more detailed discussion about the non-universality of the approximation/estimation or bias/variance tradeoffs see
\citet{neal_modern_2018,neal_on_2019}. 

Now that we have discussed what the standard approximation-estimation tradeoff is (not), let us see how it gives rise to the following three tradeoffs in our representation learning framework.

\begin{itemize}
    \item Approximation vs probe generalization
    \item Approximation vs encoder generalization
    \item Usability vs probe generalization
\end{itemize}

\textbf{Approximation vs probe generalization} and \textbf{approximation vs encoder generalization.}
The first two tradeoffs are direct consequences of the standard approximation-estimation tradeoff.
Indeed, as discussed in \cref{appcs:sec:risk_dec:recovering}, representation learning with probing can be written as a standard supervised setting.
In this case, the supervised approximation-estimation tradeoff becomes a tradeoff between the approximation error (\cref{appx:eq:recovering:approx_error}) and the sum of encoder and probe generalization (\cref{appx:eq:recovering:probe_gen_error}).
By fixing either the encoder or the probe we then directly get the first two tradeoffs.
In the main paper, we do not discuss those two tradeoffs as they are relatively obvious and both contain the approximation error term, which is typically negligible in SSL \cref{fig:trends_main}.

\paragraph{Usability vs probe generalization}
To understand the last tradeoff, consider the downstream probing task.
For a given encoder, this corresponds to standard supervised learning and we thus know that there is an approximation vs estimation tradeoff. 
In standard supervised learning, one typically considers the underlying data distribution and the supervised learning algorithm fixed and so the only factor that affects the tradeoff is the predictive family.\footnote{If we do the same in the probing case, then we recover the aforementioned approximation vs probe generalization tradeoff.}
Holding the data distribution fixed makes sense in standard supervised learning, but for the case of probes we can actually change this distribution by using a different encoder.
Indeed, the inputs to the probes are the encoded examples and thus changing the encoder will change the underlying data distribution. 
The usability-probe generalization tradeoff then corresponds to the probe's supervised tradeoff if we keep the probing family fixed (\eg linear probe) but modify the data distribution by changing the encoder.
Changing the data distribution can indeed change the effective complexity of the probing family, which can be seen by standard data-dependent complexity measures such as the Rademacher Complexity \cite{shalevshwartz_understanding_2014}.
We thus have a trade-off between the probe's training error and the generalization that is due solely to the pretraining algorithm $\algrep$ rather than the probing family $\Qz$. 
On the one hand, if the encoder does not allow the probe to extract any input information (\eg, the representation is a constant) then the representation is not usable (large probe's training error) but the probe generalizes .
On the other hand, if the encoder allows the probe to extract all input information (\eg, the representation is a one-hot encoding of the input) then the representation is usable  but the probe will overfit.

Given that all aforementioned tradeoffs are directly derived from the standard supervised tradeoffs they are also not universal tradeoffs.
For example, it is possible to simultaneously achieve the minimal probe generalization and usability error \cite{dubois_learning_2020,dubois_improving_2022} despite the U-P tradeoff.

\clearpage
\newpage

\section{Estimators}
\label{appx:sec:estimators}
% For this section, we mostly emphasize whether estimators clearly do not satisfy some mathematical property (\eg bias or consistency). 
% We do not emphasize the exact assumptions required 

\subsection{Supervised decompositon}
\label{appx:sec:estimators:supervised}

First, let us review how risk components are estimated in practice when comparing and analyzing supervised learning models.
To estimate \cref{eq:supervised_risk_decomposition} we need the following 3 estimators.
The main challenge is that the risk components are defined using population risk, but we do not have access to the population distribution $\ptask$.
The typical way to overcome this challenge is to use plug-in empirical estimators with the data we have $\Str$ and $\Ste$.

\paragraph{$\bm{\hrS{}}$} We want to estimate the risk when the predictor is trained on finite samples. 
Using the empirical distribution $\hp{\Ste} \approx \ptask$, we get the plugin estimator corresponding to the standard evaluation loss:
$
\hrS{} \defeq \hLp{\predS; \hp{\Ste}}  \approx  \Lp{\predS}   \eqdef \rS
$.
$\hrS{}$ is unbiased and consistent under standard technical assumptions (\eg $\Ste,\Str \iidsim \ptask{}$) by the law of large numbers.

\paragraph{$\bm{\hrF{}}$} We want to estimate the risk on the population data. 
Using the empirical distribution $\hp{\Str} \approx \ptask$, we get the plugin estimator corresponding to the training loss:
$
\hrF{} \defeq \min_{f \in \Qx}   \hLp{f; \hp{\Str}} \approx  \Lp{\predF}   \eqdef \rF. 
$
It can be shown to be consistent under technical assumptions \cite{vapnik_nature_2000,mukherjee_learning_2006} but it underestimates the true risk (biased).

\paragraph{$\bm{\hbrisk{}}$} Bayes risk is hard to estimate but is actually not necessary when comparing models as it is only a function of the task.

\subsection{Decomposition for representation learning}
\label{appx:sec:estimators:supervised}

\vspace*{-1em}
\newcommand{\vspacing}{$\vphantom{\int_{i}^t}$} \begin{algorithm}[H]
\caption{Estimating risk components in the standard SSL setting}
\label{alg:components}
\begin{algorithmic}[1]
\Require{Encoder family $\Phi$, probe family $\Qz$, training ${\color{coltr}\Str}$ and testing ${\color{colte}\Ste}$ sets, SSL algorithm $\algrep$, evaluation loss $\ell$. }
\Function{risk}{$\Q, \ \mathcal{D}_{tr}, \ \mathcal{D}_{te}$ }
\State \vspacing $\hat{f} \leftarrow \inf_{f \in \Q}  \sum_{(x,y) \in \mathcal{D}_{tr}} \ell(y, f(x)) $ \Comment{Risk minimization}
\State \Return{$\frac{1}{|\mathcal{D}_{te}|} \sum_{(x,y) \in \mathcal{D}_{te}} \ell(y, \hat{f}(x))$} \Comment{Test risk}
 \EndFunction
 %
% \Procedure{decomposition}{}
\State \vspacing $ \hrFF \leftarrow  \Call{risk}{\Phi \circ \Qz, {\color{coltr}\Str}, {\color{coltr}\Str}}$ \Comment{Supervised train performance}
\State \vspacing $\phi \leftarrow  \algrep(\Phi,{\color{coltr}\Str})$ \Comment{Pretrain SSL encoder}
\State \vspacing ${\color{coltr}S^\phi_{\text{tr}}} \leftarrow  $ [($\phi$(x),y) \textbf{for} x,y \textbf{in} ${\color{coltr}\Str}$] \Comment{Featurize data}
\State \vspacing ${\color{colte}S^\phi_{\text{te}}} \leftarrow  $ [($\phi$(x),y) \textbf{for} x,y \textbf{in} ${\color{colte}\Ste}$] 
\State \vspacing ${\color{colsub}S^\phi_{\text{sub}}} \leftarrow \mathrm{subset}({\color{coltr}S^\phi_{\text{tr}}}, n=\mathrm{len}({\color{colte}\Ste}))$ 
\State \vspacing  $\hrAF \leftarrow  \Call{risk}{\Q, {\color{coltr}S^\phi_{\text{tr}}}, {\color{coltr}S^\phi_{\text{tr}}}}$ \Comment{Risk without generalization}
\State \vspacing  $\hrAS \leftarrow  \Call{risk}{\Q, {\color{coltr}S^\phi_{\text{tr}}} \setminus {\color{colsub}S^\phi_{\text{sub}}}, {\color{colsub}S^\phi_{\text{sub}}}}$ \Comment{Risk with only probe gen.}
\State \vspacing  $ \hrUS \leftarrow  \Call{risk}{\Q, {\color{coltr}S^\phi_{\text{tr}}}, {\color{colte}S^\phi_{\text{te}}}}$ \Comment{Risk with enc. and probe gen.}
%\State \vspacing test\smallunderscore risk $ \leftarrow  \Call{risk}{\Q, {\color{coltr}S^\phi_{\text{tr}}} \setminus {\color{colsub}S^\phi_{\text{sub}}}, {\color{colte}S^\phi_{\text{te}}} }$ \Comment{Risk with encoder and predictor gen.}
\State \vspacing approx\smallunderscore error $\leftarrow \hrFF $ 
\State \vspacing usability\smallunderscore error $\leftarrow \hrAF - \hrFF$ 
\State \vspacing probe\smallunderscore gen $\leftarrow \hrAS - \hrAF$ 
\State \vspacing encoder\smallunderscore gen $\leftarrow \hrUS - \hrAS$ 
%\State \vspacing aggregated\smallunderscore risk $\leftarrow \hrUS $ 
\State  \Return approx\smallunderscore error, usability\smallunderscore error, probe\smallunderscore gen, encoder\smallunderscore gen 
 %\EndProcedure
\end{algorithmic}
\end{algorithm}

\begin{figure}[H]
    \centering
    \vspace*{-2em}\includegraphics[width=0.45\linewidth]{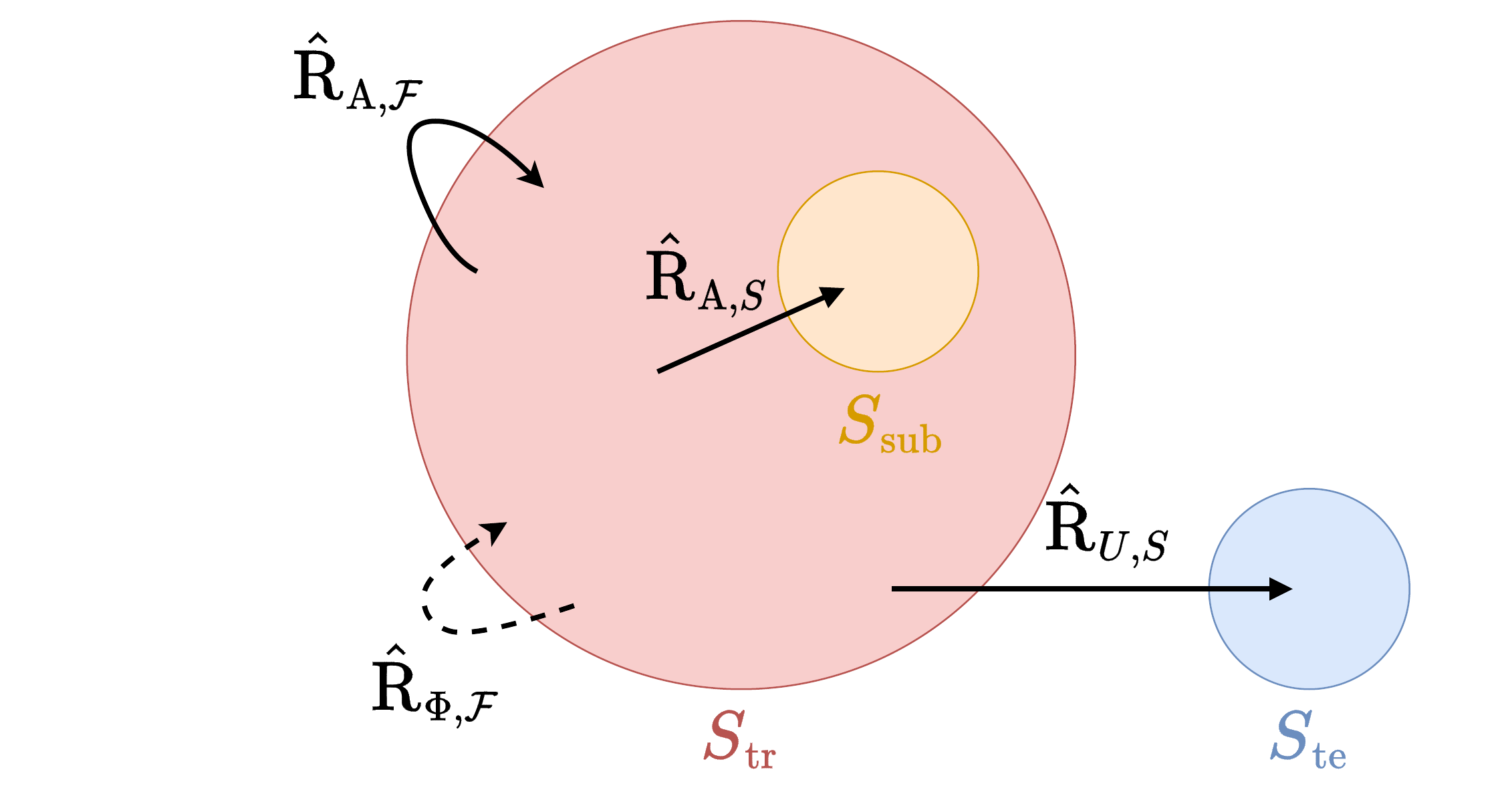}
    \caption{Estimators of our risk components in the standard SSL setting. 
    (Top) Pseudocode.
    (Bottom) Illustration of the estimators as arrows from the probe's train set to the evaluation set. 
    Full lines mean that we are only training the probe using supervised learning.
    Dashed line means that we are training both the encoder and the probe using supervised learning.
    }
    \label{fig:estimation}
\end{figure}

\Cref{fig:estimation} provides an illustration and algorithm of the estimators we proposed in \cref{sec:practical_estimation}.
%One detail that we omitted in the algorithm is that in practice we cannot train the empirical risk 
Let us now discuss each estimator in more detail.
As a reminder in the standard SSL setting we are in, the pretraining and training data distribution is the same (besides the labels) \ie, $\ppre=\ptask$.

\paragraph{$\bm{\hrUS{}}$} we want to
estimate the risk when the families $\Phi,\Qz$ are constrained, the encoder is pretrained using the algorithm $\algrep$, and both the probe and the encoder are trained on finite samples .
Using the empirical distributions $\hp{\Ste} \approx \ptask$ and $\hp{\Str} \approx \ppre$, we get the plugin estimator corresponding to the standard evaluation loss:
\begin{alignat}{3}
\rUS{} 
&\defeq \Lp{\predS{} \circ \encS{}} \qquad 
&& \text{where } \encS{} \defeq \algrep(\hp{\Stask}) \text{ and } \predS{} \defeq \algsup(\hp{\Stask}) \text{ and } \Stask{} \iidsim \ptask{} \\
&\approx  \hLp{\hpredS{} \circ \hencA{}; \hp{\Ste}} \quad 
&& \text{where } \hencA{} \defeq \algrep(\hp{\Ste}) \text{ and } \hpredS{} \defeq \algsup(\hp{\Ste}) 
&&  \hp{\Ste} \approx  \ptask \\
&\eqdef \hrUS{}
\end{alignat}
Similarly to the supervised case (\cref{appx:sec:estimators:supervised}), $\hrUS{}$ is unbiased and consistent under standard technical assumptions by the law of large numbers.

\paragraph{$\bm{\hrAS{}}$} we want to
estimate the risk when the families $\Phi,\Qz$ are constrained, the encoder is pretrained using the algorithm $\algrep$ on the population distribution, but the probe is now trained on finite samples $\Stask{} \iidsim \ptask{}$.
We will again use  the empirical distributions $\hp{\Str} \approx \ptask$ as a plug in estimate for the population distribution.
This means that the finite training data for the probes will need to be sampled from the empirical distribution  $\hp{\Str}$ to emulate the fact that the probe has to generalize to unseen data.
To do so we partition the training data into a small subset $\Ssub$ on which we train the probe and its complement $\Str \setminus \Ssub$ for evaluation.
The final estimator is:
\begin{alignat}{3}
\rAS{} 
&\defeq \Lp{\predS{} \circ \encA{}} \qquad 
&& \text{where } \encA{} \defeq \algrep(\ppre) \text{ and } \predS{} \defeq \algsup(\hp{\Stask}) \text{ and } \Stask{} \iidsim \ptask{} \\
&\approx  \Lp{\predS{} \circ \hencA{}} \quad 
&& \text{where } \hencA{} \defeq \algrep(\hp{\Str}) \text{ and } \predS{} \defeq \algsup(\hp{\Stask}) \text{ and } \Stask \iidsim \ptask{} \qquad
&& \hp{\Str} \approx  \ppre \\
&\approx  \hLp{\predS{} \circ \hencA{}; \hp{\Ssub}} \quad 
&& \text{where }\hencA{} \defeq \algrep(\hp{\Str}) \text{ and } \predS{} \defeq \algsup(\hp{\Stask}) \text{ and } \Stask \iidsim \ptask{} \qquad
&& \hp{\Ssub} \approx  \ptask\\ 
&\approx  \hLp{\hpredS{} \circ \hencA{}; \hp{\Ssub}} \quad 
&& \text{where } \hencA{} \defeq \algrep(\hp{\Str}) \text{ and } \hpredS{} \defeq \algsup(\hp{\Str \setminus \Ssub}) 
&&  \Str \setminus \Ssub \approx \Stask \\
&\eqdef \hrAS{}
\end{alignat}
The estimator can be shown to be consistent for the training set $S \defeq \Str \setminus \Ssub$ in the case where $\Str \setminus \Ssub$ is fixed but $|\Str|,|\Ssub| \to \infty$.
The estimator is generally biased.
One other issue with the estimator is that it is consistent for the training set $S \defeq \Str \setminus \Ssub$ instead of $S \defeq \Str$.
In the case where $|\Str| \gg |\Ssub|$ this should be negligible as $\hp{\Str \setminus \Ssub}$ will be close to $\hp{\Str}$.
This is why in practice we use a very small $\Ssub$.
In particular, for ImageNet we have $|\Ssub| = 5\scip{4}$ and $|\Str| > 1\scip{6}$.

\paragraph{$\bm{\hrAF{}}$} we want to
estimate the risk when the families $\Phi,\Qz$ are constrained, the encoder is pretrained using the algorithm $\algrep$, but the probe and encoder are pretrained on the population distribution.
The challenge is that we do not have access to the population distribution.
Using the empirical distributions $\hp{\Str} \approx \ptask$, we get a plugin estimator that corresponds to (pre)training the encoder and probe on the same distribution as they are being evaluate it on.
This is the standard training error of the probe:
\begin{alignat}{3}
\rAF{} 
&\defeq  \inf_{f\in \Qz}  \Lp{f \circ \encA{}} \quad &&\text{where } \encA{} \defeq \algrep(\ppre) \\
&\approx \inf_{f\in \Qz} \hLp{f \circ \hencA{}; \hp{\Str}} \quad &&\text{where } \hencA{} \defeq \algrep(\hp{\Str}) \qquad && \hp{\Str} \approx \ptask = \ppre \\
&\eqdef \hrAF{}
\end{alignat}
The estimator is similar to $\hrFF{}$ in that we use $\hp{\Str}$ as a plug in estimate for the pretraining/training/evaluation set.
$\hrAF{}$ can thus also be shown to be consistent (as $|\Str| \to \infty$) under the technical assumptions but it is biased (typically underestimates the true risk).

\paragraph{$\bm{\hrFF{}}$} we want to
estimate the best achievable risk for a given encoder and probe family $\Phi \circ \Qz$. 
The problem is that we do not have access to the population distribution.
Using the empirical distributions $\hp{\Str} \approx \ptask$, we get a plugin estimator that corresponds to the empirical risk minima (\ie the training loss of a supervised model):
\begin{align}
 \rFF{} 
 &\defeq \inf_{f\in \Qz} \inf_{\phi \in \Phi} \Lp{f \circ \phi}  \\
 &\approx \inf_{f\in \Qz} \inf_{\phi \in \Phi} \hLp{f \circ \phi; \hp{\Str}} & \hp{\Str} \approx \ptask = \ppre \\ 
  &\eqdef \hrFF{}
\end{align}
Just as with the supervised case (\cref{appx:sec:estimators:supervised}) 
it can be shown to be consistent (as $|\Str| \to \infty$) under the technical assumptions but it underestimates the true risk (biased).
Indeed, this is the supervised empirical risk minima for predictors in $\Phi \circ \Qz$.
Note that $\hrFF{}$ requires training a supervised model (empirical risk minimizer).
This can be computationally prohibitive for large $\Phi{}$, but is only required once per architecture and such pretrained model can often be found online.
One issue with online models is that their empirical risk typically overestimate the desired minimal risk, as they are typically regularized.

\paragraph{$\bm{\hbrisk{}}$} Just as in the supervised case (\cref{appx:sec:estimators:supervised}), the Bayes risk is unknown but it only depends on the task so we can disregard it is the same for all compared models.

The properties of the estimators are summarized in \cref{appx:tab:properties_estimaton}

\begin{table}[h]
\centering
\begin{threeparttable}
\caption{Properties of each estimator.}
\label{appx:tab:properties_estimaton}
\begin{tabular}{lrrr}
\toprule
estimator &  consistent & unbiased & computationally efficient \\
\midrule
$\bm{\hrUS{}}$ & $\checkmark$  & $\checkmark$ & $\checkmark$ \\
$\bm{\hrAS{}}$ & $\checkmark$  & \xmark & $\checkmark$ \\
$\bm{\hrAF{}}$ & $\checkmark$\tnote{*}  & \xmark & $\checkmark$\\
$\bm{\hrFF{}}$ & $\checkmark$  & \xmark & \xmark\tnote{\textdagger} \\
\bottomrule
\end{tabular}
\begin{tablenotes}
\item[*] The estimator is consistent for the training set $S = \Str \setminus \Ssub$ rather than $S= \Str$.
\item[\textdagger] The estimator requires training a supervised model of architecture $\Phi \circ \Qz$, which can be inefficient.
This is only required once per architecture and thus becomes efficient when comparing multiple models of the same architecture. 
Furthermore, such supervised model can often be found online.
\end{tablenotes}
 \end{threeparttable}
\end{table}

\clearpage
\newpage

\section{Experimental details}
\label{appx:sec:reproducibility}

\subsection{Open source API}
\label{appx:sec:details:open_source}

All the pretraining encoders, their associated metadata, and the results discussed below are available via a simple and unified API using respectively:
\begin{description}[noitemsep]
    \item[Models] \texttt{torch.hub.load("\github:main", encoder)} returns a pretrained pytorch \texttt{encoder} and the \texttt{preprocessing} pipeline. 
    For all our models a \texttt{PIL} image \texttt{x} can be encoded using \texttt{encoder(preprocessing(x).unsqueeze(0))}. 
    A list of available models can be found using \texttt{torch.hub.list("\github:main")}. 
    Each model's name is \texttt{<objective>\_<architecture>\_<other>}, where \texttt{other} is some compressed metadata that we use to distinguish models (it is the same as the ``other'' column in \cref{appx:tab:all_results}).
    
    \item[Metadata] \texttt{torch.hub.load("\github:main", "metadata\_df")} returns a pandas dataframe of all metadata. For a nested dictionary use \texttt{"metadata\_dict"} instead.
    \item[Results] \texttt{torch.hub.load("\github:main", "results")} returns a dataframe of all evaluated metrics (corresponding to \cref{appx:tab:all_results}).
\end{description}
More details and our evaluation code can also be found at \codeurl{}.

\subsection{Pretrained models and metadata}
\label{appx:sec:details:pretrained}

Aside from the \NOurPre{} SSL models we pretrained, all others were taken from:
\href{https://pytorch.org/hub}{torch hub}, \href{https://github.com/pytorch/vision}{torchvision}, 
\href{https://github.com/facebookresearch/vissl}{VISSL}, 
\href{https://github.com/rwightman/pytorch-image-models}{timm},
\href{https://huggingface.co/models}{Hugging Face },
\href{https://github.com/open-mmlab/mmselfsup}{MMSelfSup},
\href{https://github.com/HobbitLong/PyContrast}{PyContrast}, or from the official GitHub repository of the considered model.

In total, we consider \Npre{} pretrained encoders that we broadly categorize in the following categories:
\begin{description}
\item[Predicting transformations] First, there are the encoders that are pretraiend by essentially predicting the augmented transformation. 
In particular, LocNet \cite{doersch_unsupervised_2015}, Jigsaw \cite{noroozi_unsupervised_2016}, RotNet \cite{gidaris_unsupervised_2018}.
\item[Contrastive] We use contrastive to mean any methods that use some derivative of InfoNCE \cite{oord_representation_2019}.
Specifically, we consider 
NPID \cite{wu_unsupervised_2018}, 
NPID++ \cite{misra_self-supervised_2020}, 
PIRL \cite{misra_self-supervised_2020}, 
MoCo \cite{he_momentum_2020}, 
MoCov2 \cite{chen_improved_2020}, 
MoCov3 \cite{chen_empirical_2021}, 
SimCLR \cite{chen_simple_2020}, 
CLIP \cite{radford_learning_2021},
Lossyless \cite{dubois_lossy_2021},
SpecCL \cite{haochen_provable_2021}.
\item[Hierarchical] We use hierarchical  to mean methods that have a local and global component of the loss.
Specifically, we consider 
DenseCL \cite{wang_dense_2021}, 
MUGS \cite{zhou_mugs_2022},
VICRegL \cite{bardes_vicregl_2022}.
.
\item[Clustering] We use clustering to mean any method where representations are learned by predicting clusters of the data (\eg via a clustering step or jointly learned by a teacher).
Specifically, we consider 
DeepCluster \cite{caron_deep_2018}, 
ClusterFit \cite{yan_clusterfit_2020},
SwAV \cite{caron_unsupervised_2020},
DeepClusterv2 \cite{caron_unsupervised_2020},
Selav2 \cite{asano_self-labelling_2020,caron_unsupervised_2020},
ODC \cite{zhan_online_2020},
iBOT \cite{zhou_ibot_2021},
DINO \cite{caron_emerging_2021},
DISSL \cite{dubois_improving_2022},
MSN \cite{assran_masked_2022}.
\item[Siamese] We call ``siamese'' models that do not nicely fall in the previous categories but still use siamese networks. 
This includes 
BYOL \cite{grill_bootstrap_2020},
SimSiam \cite{chen_simpler_2021},
Barlow Twins \cite{zbontar_barlow_2021},
VICReg \cite{bardes_vicreg_2022}.
\item[Generative] We consider models that were pretrained with variants of Bert-style  \citep{devlin_bert_2019} masking for vision. 
Specifically, we consider BEiT \citep{bao_beit_2022},
BEiTv2 \citep{zhiliang_beit_2022},
and MAE \citep{he_masked_2022}.
\item[Supervised] Finally, we also download and evaluate (with linear probing) pretrained supervised models. 
The reason is two-fold.
First, supervised models of the same architecture are an important baseline to understand the performance of SSL encoders.
Second, those models are used to estimate the approximation error as discussed in \cref{sec:practical_estimation}.
In particular, we considered supervised ViTs \cite{dosovitskiy_image_2021} and ResNets \cite{he_deep_2016} of various architecture. 
\end{description}
Note that for each of the SSL models we consider different hyperparameters, such as the encoder's architecture or the number of training epochs).
For each of the pretrained model we also collected (to the best of our ability) metadata including information about the SSL objective, the architecture, the pretraining data, the representation, the pretraining optimization, and the compute budget.
In particular, we collected the following information when applicable and available.
 \begin{multicols}{3}
    \begin{itemize}
        \item SSL objective
        \item SSL category 
        \item version of the objective
        \item number of negatives
        \item number of classes
        \item uses stop-gradients?
        \item uses EMA encoder?
        \item output dim. of proj.
        \item width of proj. head
        \item depth of proj. head
        \item architecture
        \item architecture family
        \item patch size
        \item architecture of proj. head 1
        \item architecture of proj. head 2
        \item weight tying between proj. head?
        \item \# of parameters for encoder 
        \item \# of param. for proj. 
        \item dim. of representation
        \item representation layer
        \item epochs
        \item batch size
        \item optimizer
        \item learning rate
        \item weight decay
        \item learning rate scheduler
        \item pretraining data
        \item finetuning data
        \item image size
        \item number of views
        \item invariant to aug?
        \item list of augmentations
        \item publication date
        \item license of weights
        \item official weights?
        \item model trained in industry?
        \item pretraining time
        \item type of pretraining machine
        \item number of pretraining machines
    \end{itemize}
    \end{multicols}

\subsection{Evaluating all metrics}
\label{appx:sec:details:metrics}

One of the contributions of our paper is to provide a thorough and fair linear probing evaluation of \Npre{} pretrained models in 5 different label settings (100\%, 30-shot, 1\%, 5-shot, 3-shot).
We now describe the evaluation pipeline for each of the models. 
The code is available online at \codeurl{}.

\paragraph{Featurization} For each pretrained model, we first featurize the entire ImageNet dataset (train and test) similarly to \citet{cherti_reproducible_2022,dubois_improving_2022,dubois_lossy_2021,santurkar2022caption}.
This differs from the standard SSL pipeline where images are featurized on-the-fly at every step \cite{caron_emerging_2021,caron_unsupervised_2020,chen_simple_2020,chen_simpler_2021}.
The advantage of prefeaturization is that training a probe becomes ${\sim}1000\times$ faster ( ${\sim} 100$ GPU hours $\to$ ${\sim} 10$ min).
The disadvantage is that we cannot use data augmentations to train the probe, which decreases accuracy by an average of $1$ percent point. 

For the following estimators, we essentially follow \cref{alg:components}.

\paragraph{Full-shot linear probing or $\hrUS$}
To evaluate full-shot linear probing we use PyTorch \cite{paszke_pytorch_2019} and tune the following hyperparameters: lr, weight decay, batch size, is batchnorm, optimizer, scheduler.
In particular, we see that the linear probe is potentially regularized.
The hyperparameters are tuned using 30 steps of the Tree Parzen Estimator algorithm (TPE; \cite{bergstra2011algorithms}) to minimize a validation error.
For computational efficiency, we only train the probe  on 10\% of ImageNet during tuning.
Once the hyperaparameters are tuned we train the linear probe on all of ImageNet and return the test error.
This corresponds to our desired full-shot metric as well as $\hrUS$.

\paragraph{Estimating $\hrFF$}
To compute $ \hrFF$ we need to train a supervised encoder of the desired architecture (\cref{alg:components}), which can be computationally prohibitive.
As there are many online available supervised model, we, instead, download the model of the desired architecture (\eg ResNet50) and evaluate its training performance. 
One issue with this strategy is that models available online are typically tuned to perform well on a validation set, rather than on a training set as desired.
This means that we actually overestimate  $\hrFF$ and thus the approximation error.
This should not be a major issue given that our results show that the approximation error is actually very small (see \cref{appx:sec:results:trends}), \eg, for a ResNet50 we get  $\hrFF=0.84$ and so we don't overestimate the error by much.

\paragraph{Estimating $\hrAS,\hrAF$}
For $\hrAS,\hrAF$ we follow the tuning pipeline used for $\hrUS$ (full-shot linear probing), the only difference being the train/validation/test data.
Specifically, we always tune the probe on a dataset that mirrors the evaluation set.
For example, for $\hrAF$ the probe is trained and tested on ImageNet's train set (\cref{alg:components}), and so tuning is performed on the training set.
For $\hrAS$ the probe is evaluated on $\Ssub$ (where $|\Ssub|=50K$) and evaluated on $\Str \setminus \Ssub$, for tuning we do the same but use a different $\Ssub$.

\paragraph{Risk components}
Once we have $\hrAS,\hrAF,\hrFF,\hrUS$ we compute the risk components by using their definitions (see last lines of \cref{alg:components})

\paragraph{Few-shot linear probing}
To compute the few shot linear probes, we the same high-level pipeline as for the full-shot probing but now use sklearn's \cite{pedregosa_scikit-learn_2011} logistic regression with the lbfgs solver, which we found to be more efficient than PyTorch.
We tune only the regularization parameter C using again 30 rounds of TPE.

% \subsection{Evaluating all decomposition}
% \label{appx:sec:details:decompositions}

% %\subsection{Issues with evaluating \approxerr}
% %\label{appx:sec:details:issues_approx}

% Note that $\hrFF{}$ requires training a supervised model.
% This can be computationally prohibitive for large $\Phi{}$, but is only required once per architecture and such pretrained model can typically already be found online.
% One issue with online models is that their empirical risk overestimate the desired minimal risk, as standard supervised algorithm typically add regularization (\ie $\algsup$ is not ERM).
% If $\algsup$ is ERM, then the estimator improves as $|\Stask|$ increases and is consistent but biased.

\subsection{Evaluating the impact of different hyperparameters}
\label{appx:sec:details:hparam_impact}

Given all the hyperparameters and metrics (performance in different settings and risk decomposition) that we have collected, we now want to evaluate the impact of each of the former on the latter.
We do so using three different methods:

\paragraph{Controlled analysis (CA) and linear model }
The most obvious way to analyze the impact of a hyperparameter on some metric is to consider models that differ only w.r.t. that hyperparameter. 
When such models are available, we train a linear model to predict the impact of that hyperparameter on the desired metric.
Specifically, we train $f(\text{metric}) = \alpha\cdot f(\text{hyperparam}) + \bm{\beta} \cdot [\text{model}]$ where ``metric"" denotes the metric we are predicting,$\alpha,\bm{\beta}$ are respectively a scalar and vector parameter fitted by least-squares, ``[\text{model}]'' is a one-hot encoding of the current model (models that differ in any other hyperparameter will have a different encoding), and $f()$ denotes either a log function or the identity whichever is best.

This controlled analysis has the advantage of removing the impact of any potential confounders.
The disadvantage is that it only quantifies (potentially log) linear relationships, and there are not that many models that only differ in a single hyperparameter so there is a coverage and statistical power issue. 

\begin{table}[h]
\centering
\caption{Percentage of explained test variance (estimated by 30-fold cross-validation) for our XGBoost models before and after filtering. Each column corresponds to a different model predicting the given metric.}
\label{appx:tab:xgboost_eval}
\begin{tabular}{lrrrrrr}
\toprule
 & Approx. & Usability & Probe gen. & Enc. gen. & Full-shot & 3-shot \\
\midrule
Pre-filtering & 96.10 & 65.46 & 86.41 & 43.52 & 85.28 & 92.69 \\
Post-filtering & 89.59 & 68.17 & 87.26 & 41.35 & 86.05 & 92.48 \\
\bottomrule
\end{tabular}
\end{table}

\paragraph{XGBoost + SHAP values}
We train one XGBoost model \cite{chen_xgboost_2016} for each metric that takes 51 available hyperparameters as inputs. 
We tune each of them separately using 50 runs of Bayesian hyperparameter tuning (Tree-structured Parzen Estimator) with 10-fold cross-validation.
We then use the XGBoost models to give us the importance of each hyperparameter on a specific metric using SHAP values \cite{lundberg_unified_2017}, which essentially estimates the impact of not using a certain hyperparameter for prediction.
One issue with the above strategy is that when the hyperparameters are highly correlated it is hard to quantify the impact of those hyperparameters.
To avoid such a problem, we filter features so as to decrease correlation without decreasing the cross-validation performance. 
This allows us to decrease the number of hyperparameters to 14 without decreasing the performance of the Xgboost model. The 14 hyperparameters that we retain are: \pyth{['objective', 'architecture', 'patch_size', 'epochs', 'pretraining_data', 'projection2_arch', 'nviews', 'z_dim', 'family', 'ssl_mode', 'n_parameters', 'n_augmentations', 'optimizer', 'projection_nparameters_hidden']}.
When evaluating hyperparameters that are in that list, we use those models trained after feature selection and we use the full model otherwise. 

\Cref{appx:tab:xgboost_eval} shows the percentage of test variance explained by the XGBoost model before and after features selection, we see that pos-filtering performs surprisingly well given that it needs to predict the performance on unseen models given only 14 hyperparameters and using less than 200 training examples.
The model does nevertheless struggle for encoder generalization and to a lesser extent usability, which suggests that we might have failed to consider an important hyperparamater. 

The main advantage of using XGBoost + SHAP values is that we can quantify non-linear relations and arbitrary interactions between hyperparameters, and that the output depends on all models (rather than only the ones that differ in a single hyperparameter). 
The disadvantage is that XGboost+SHAP values are harder to interpret and we cannot quantify statistical significance.

\paragraph{Global linear analysis (GLA)}
Finally, we also train a (potentially log) linear model to predict the metric using the desired hyperparameter while controlling for all other main hyperparameters that are not directly related to the desired hyperparameter. 
For example, when evaluating the impact of the ``architecture'' we do not condition on the ``z\_dim'' or the model ``family''  as those a directly related to the architecture. 
The advantage of this global linear model is that it does not suffer from the same coverage/statistical power issue than the controlled analysis. 
The issue is that the model is very simple (linearity without any interaction term) and we might not correctly control all confounders.

All of the above methods have some complementary advantages and disadvantages for interpreting the impact of a hyperparameter, which is why we consider the three simultaneously.

%The only hyperparameter for which some methods gave contradicting is the impact of \tofill{} on \tofill{} and we are thus relatively confident 

\clearpage
\newpage

\begin{figure}[h]
\centering
\begin{subfigure}[t]{0.44\linewidth}
\centering
\captionsetup{font=scriptsize,labelfont=scriptsize}
\includegraphics[width=\linewidth]{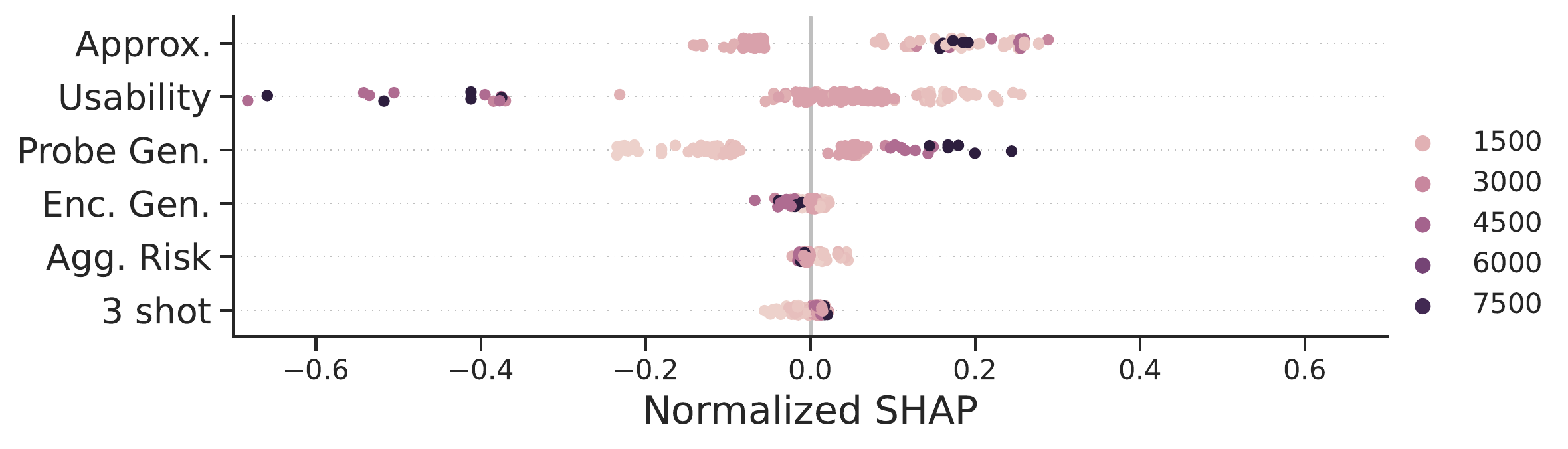}
\caption{Z Dim.}
\label{appx:fig:all_hparam_plots:z_dim}
\end{subfigure}
\hfill{}
\begin{subfigure}[t]{0.44\linewidth}
\centering
\captionsetup{font=scriptsize,labelfont=scriptsize}
\includegraphics[width=\linewidth]{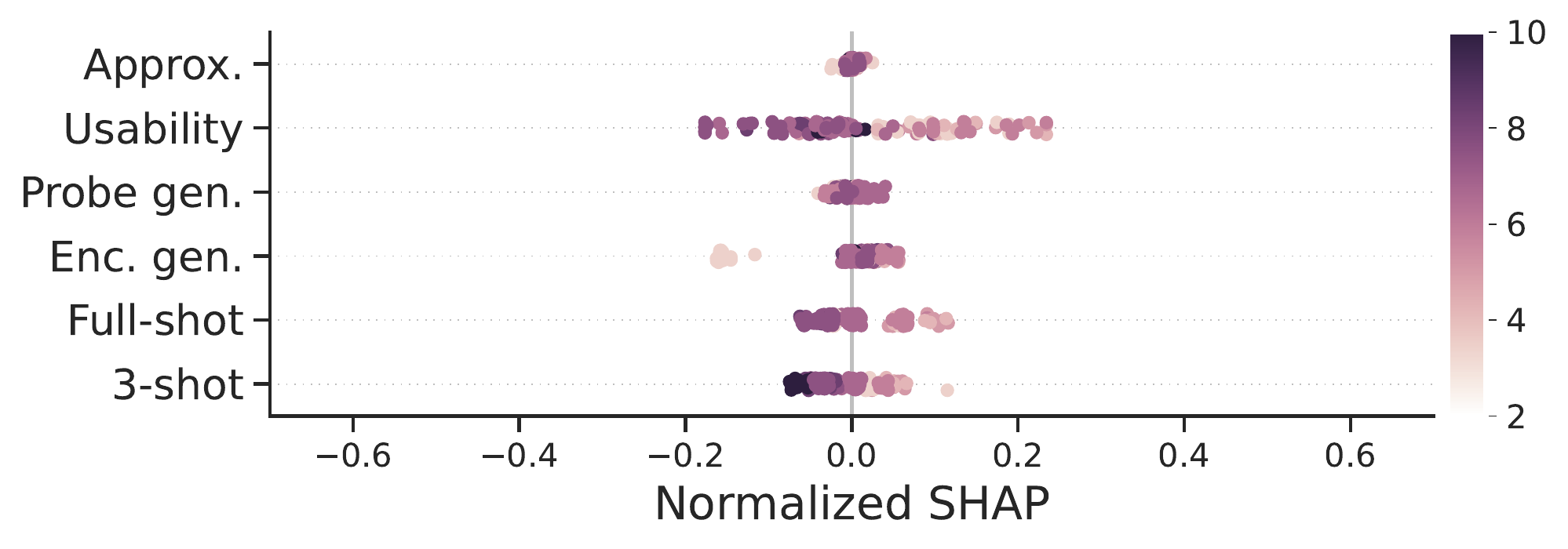}
\caption{Num. Augmetations}
\label{appx:fig:all_hparam_plots:n_augmentations}
\end{subfigure}
\begin{subfigure}[t]{0.44\linewidth}
\centering
\captionsetup{font=scriptsize,labelfont=scriptsize}
\includegraphics[width=\linewidth]{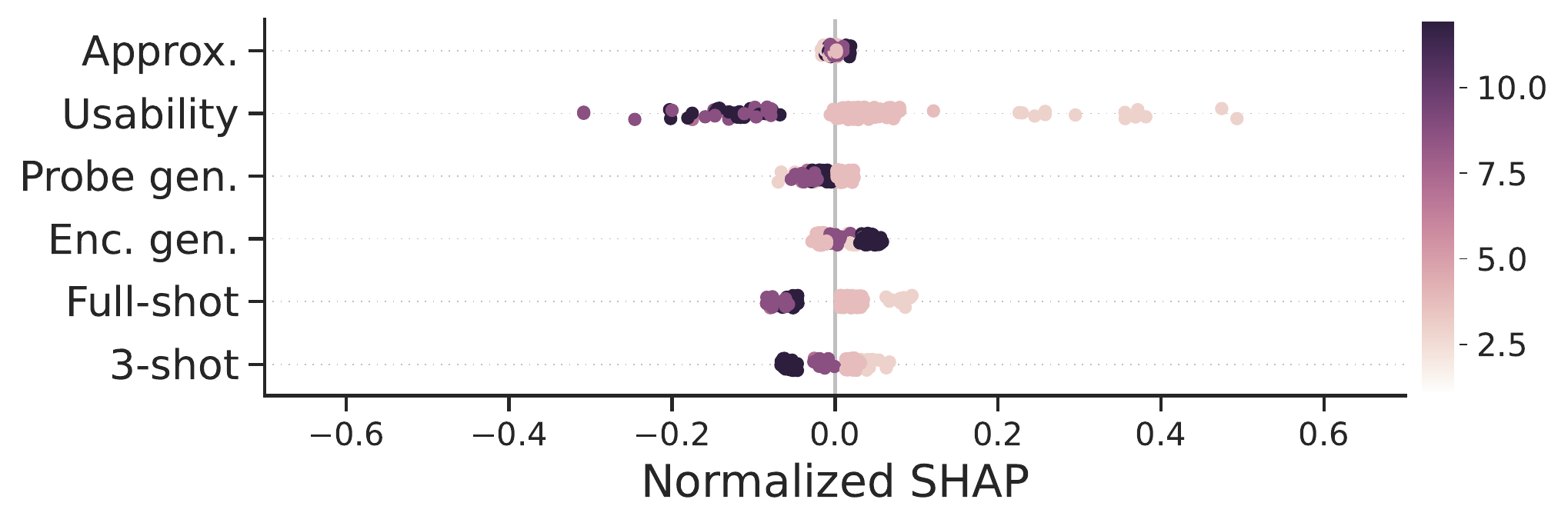}
\caption{Num. Views}
\label{appx:fig:all_hparam_plots:nviews}
\end{subfigure}
\hfill{}
\begin{subfigure}[t]{0.44\linewidth}
\centering
\captionsetup{font=scriptsize,labelfont=scriptsize}
\includegraphics[width=\linewidth]{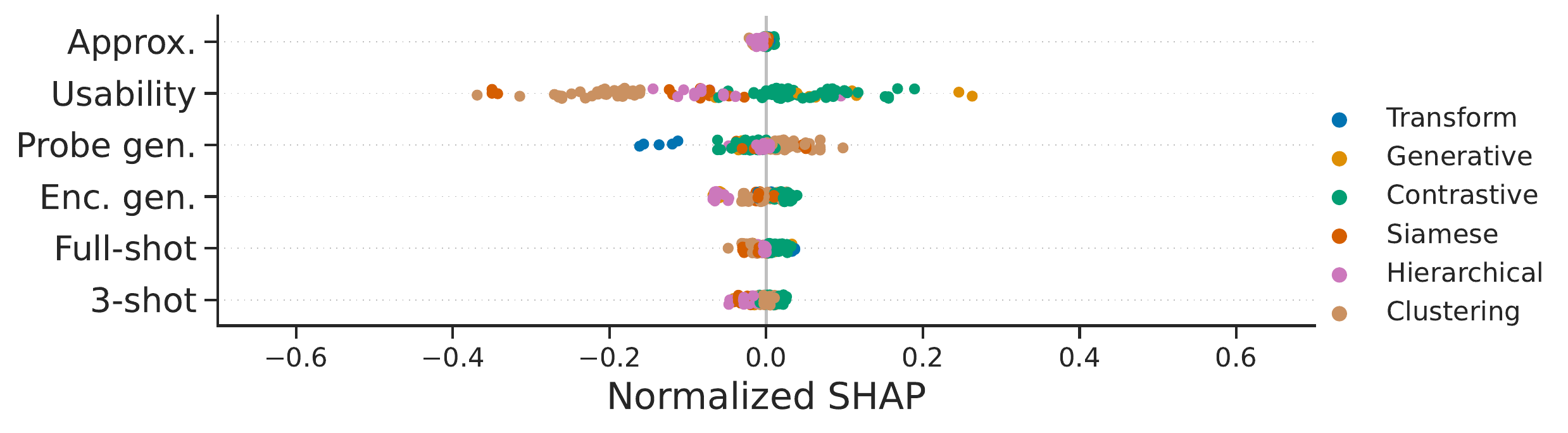}
\caption{SSL Mode}
\label{appx:fig:all_hparam_plots:ssl_mode}
\end{subfigure}
\begin{subfigure}[t]{0.44\linewidth}
\centering
\captionsetup{font=scriptsize,labelfont=scriptsize}
\includegraphics[width=\linewidth]{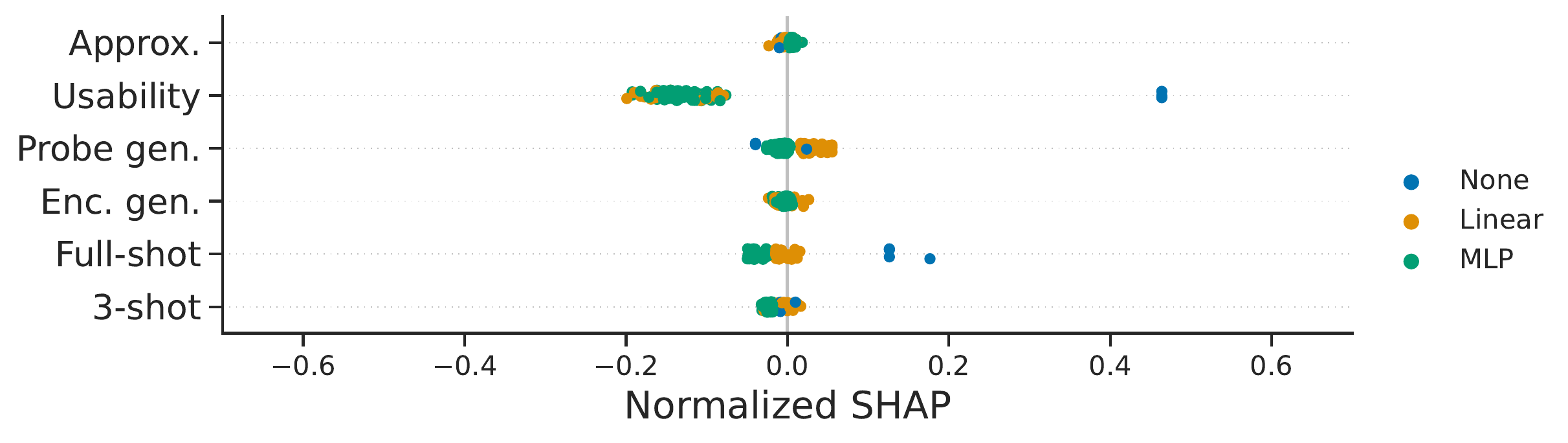}
\caption{Proj. Arch.}
\label{appx:fig:all_hparam_plots:projection2_arch}
\end{subfigure}
\hfill{}
\begin{subfigure}[t]{0.44\linewidth}
\centering
\captionsetup{font=scriptsize,labelfont=scriptsize}
\includegraphics[width=\linewidth]{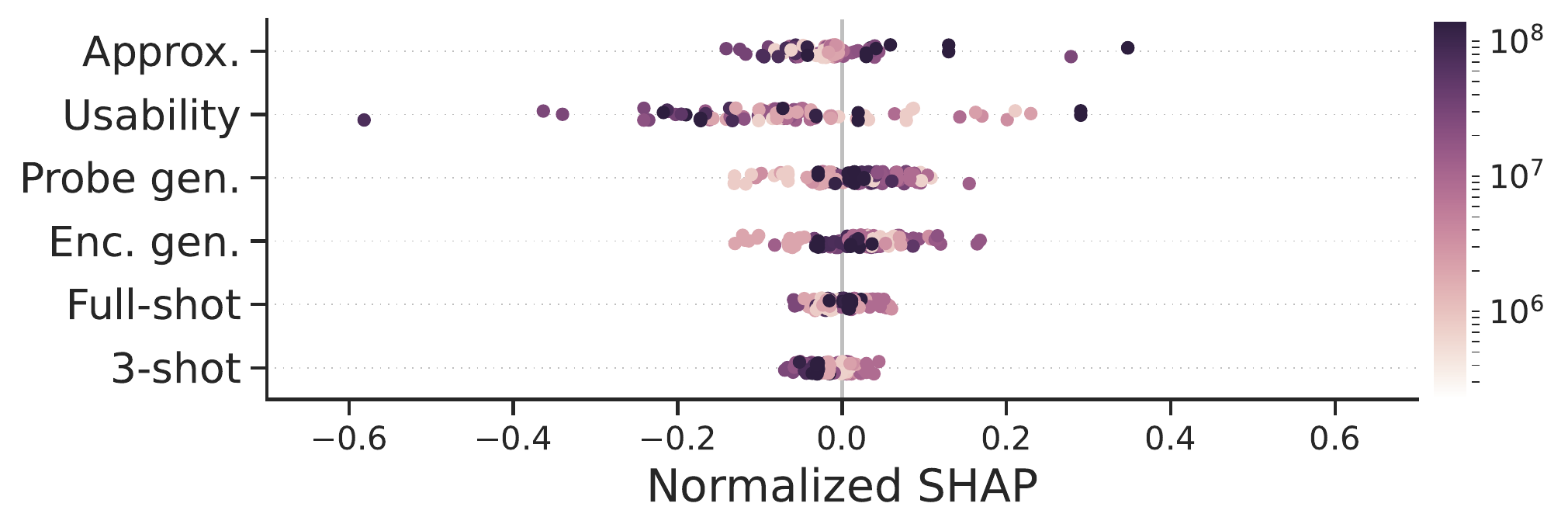}
\caption{Num. Projection Parameters}
\label{appx:fig:all_hparam_plots:projection_nparameters_hidden}
\end{subfigure}
\begin{subfigure}[t]{0.44\linewidth}
\centering
\captionsetup{font=scriptsize,labelfont=scriptsize}
\includegraphics[width=\linewidth]{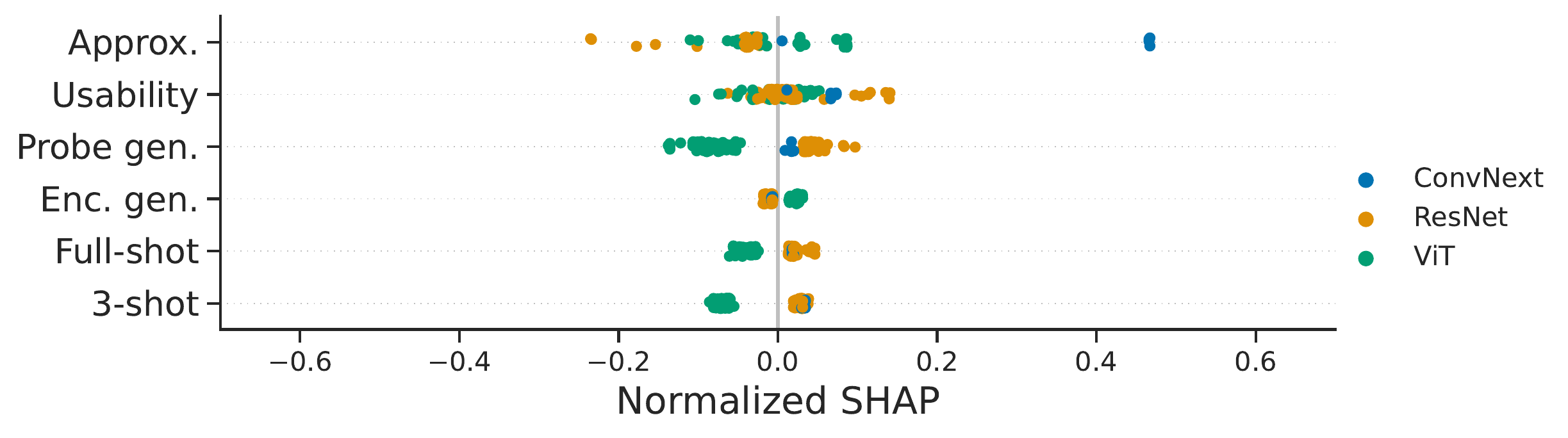}
\caption{Family}
\label{appx:fig:all_hparam_plots:family}
\end{subfigure}
\hfill{}
\begin{subfigure}[t]{0.44\linewidth}
\centering
\captionsetup{font=scriptsize,labelfont=scriptsize}
\includegraphics[width=\linewidth]{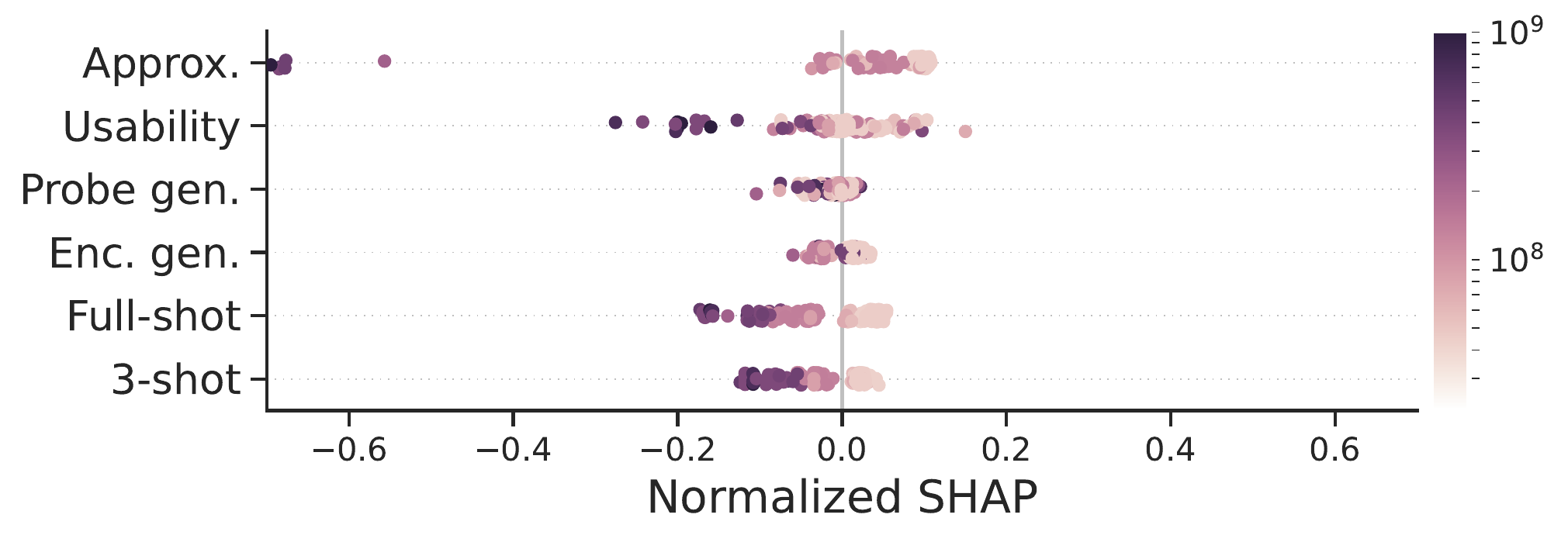}
\caption{Num. Parameters}
\label{appx:fig:all_hparam_plots:n_parameters}
\end{subfigure}
\begin{subfigure}[t]{0.44\linewidth}
\centering
\captionsetup{font=scriptsize,labelfont=scriptsize}
\includegraphics[width=\linewidth]{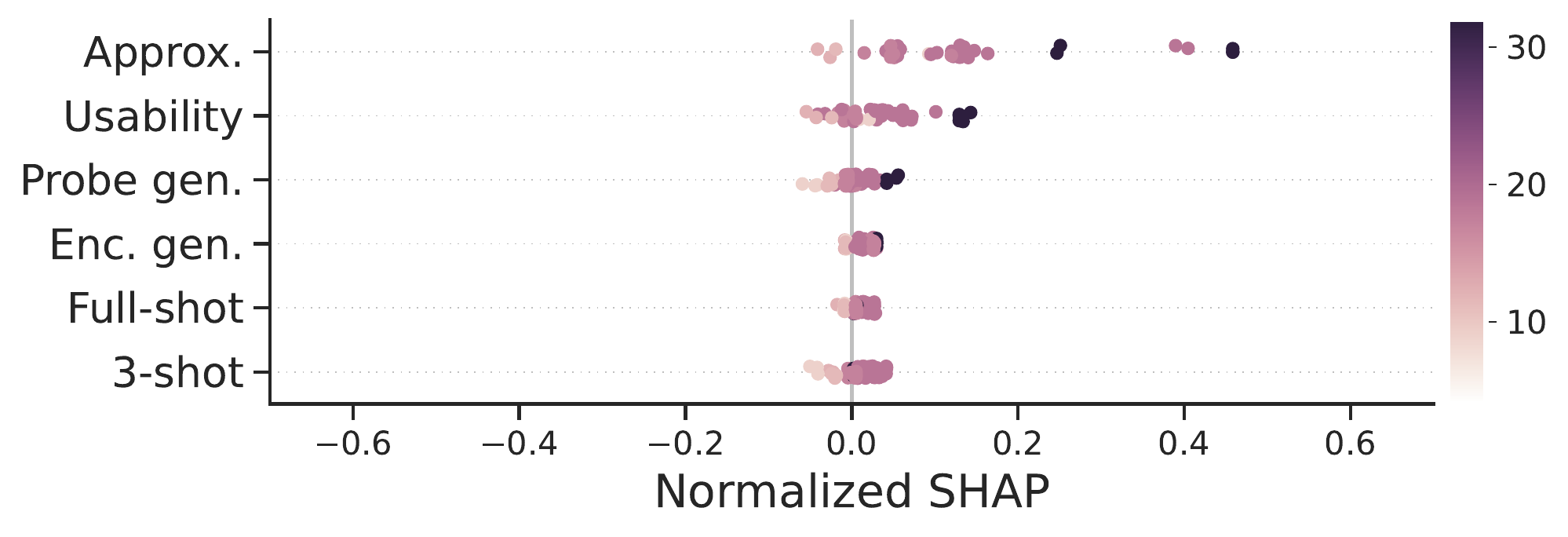}
\caption{Patch Size}
\label{appx:fig:all_hparam_plots:patch_size}
\end{subfigure}
\hfill{}
\begin{subfigure}[t]{0.44\linewidth}
\centering
\captionsetup{font=scriptsize,labelfont=scriptsize}
\includegraphics[width=\linewidth]{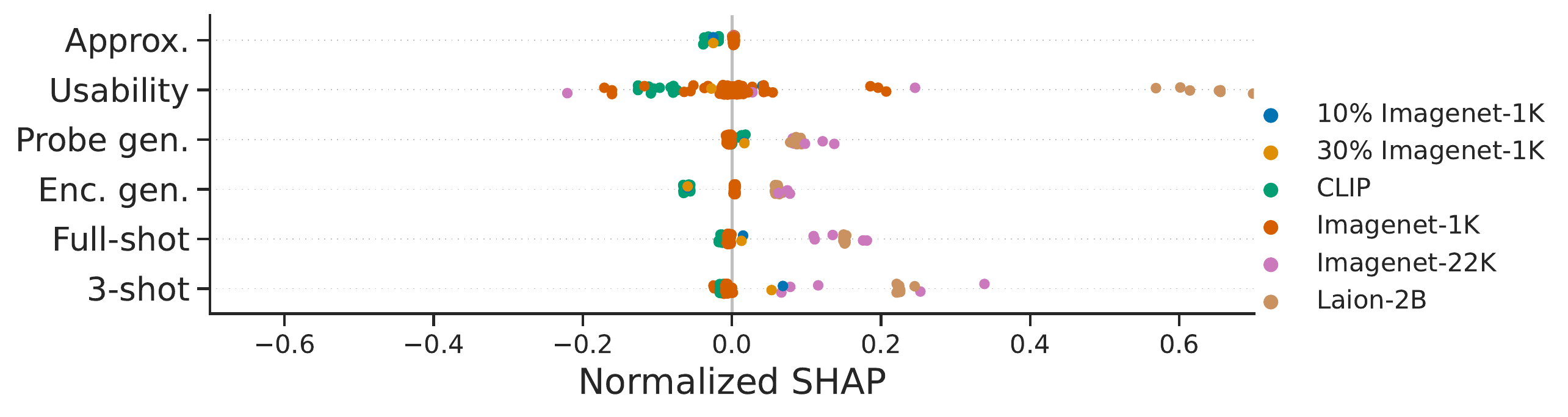}
\caption{Pretraining Data}
\label{appx:fig:all_hparam_plots:pretraining_data}
\end{subfigure}
\begin{subfigure}[t]{0.44\linewidth}
\centering
\captionsetup{font=scriptsize,labelfont=scriptsize}
\includegraphics[width=\linewidth]{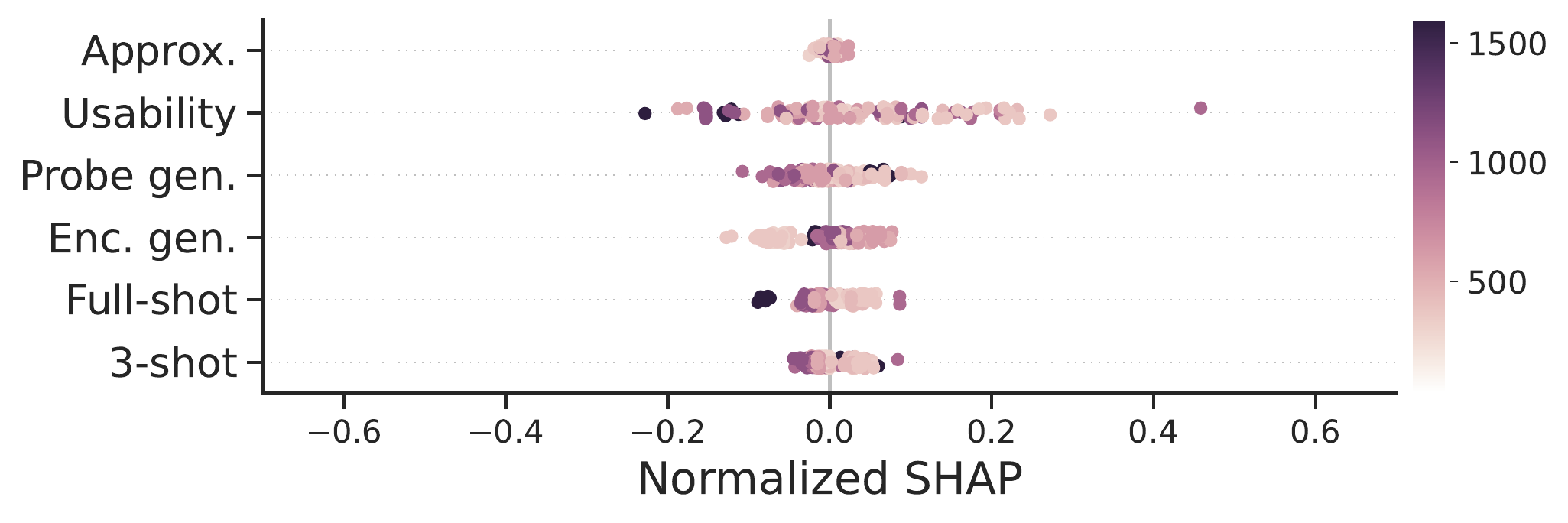}
\caption{Epochs}
\label{appx:fig:all_hparam_plots:epochs}
\end{subfigure}
\hfill{}
\begin{subfigure}[t]{0.44\linewidth}
\centering
\captionsetup{font=scriptsize,labelfont=scriptsize}
\includegraphics[width=\linewidth]{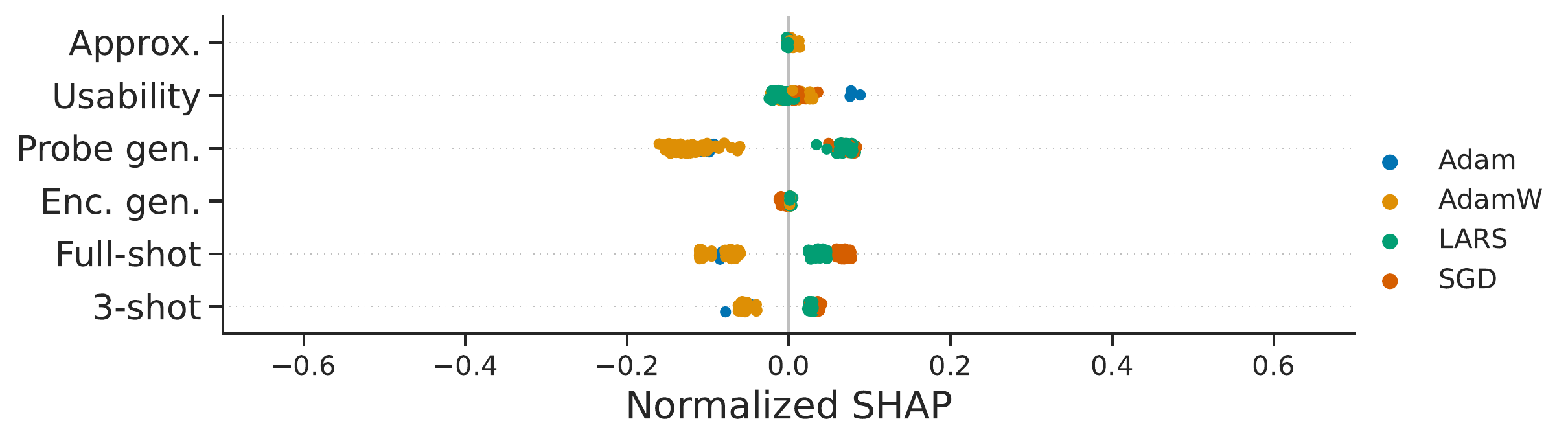}
\caption{Optimizer}
\label{appx:fig:all_hparam_plots:optimizer}
\end{subfigure}
% \begin{subfigure}[t]{0.44\linewidth}
% \centering
% \captionsetup{font=scriptsize,labelfont=scriptsize}
% \includegraphics[width=\linewidth]{figures/hparams_all/learning_rate.pdf}
% \caption{Learning Rate}
% \label{appx:fig:all_hparam_plots:learning_rate}
% \end{subfigure}
% \hfill{}
% \begin{subfigure}[t]{0.44\linewidth}
% \centering
% \captionsetup{font=scriptsize,labelfont=scriptsize}
% \includegraphics[width=\linewidth]{figures/hparams_all/weight_decay.pdf}
% \caption{Weight Decay}
% \label{appx:fig:all_hparam_plots:weight_decay}
% \end{subfigure}
\begin{subfigure}[t]{0.44\linewidth}
\centering
\captionsetup{font=scriptsize,labelfont=scriptsize}
\includegraphics[width=\linewidth]{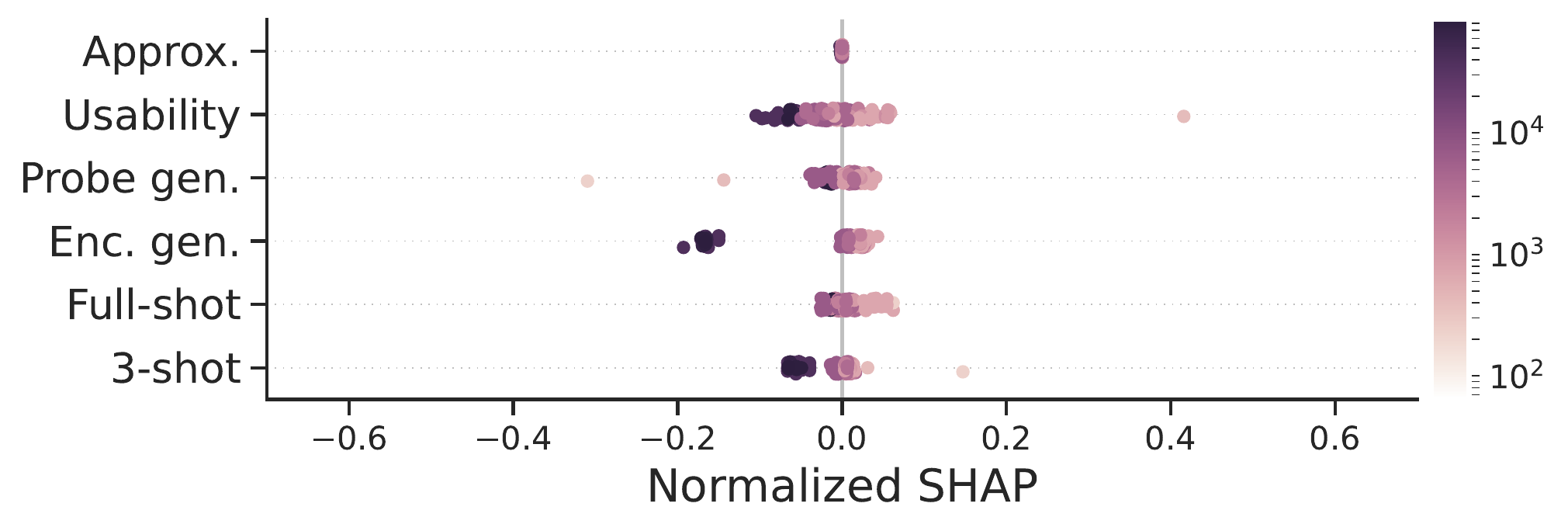}
\caption{Batch Size}
\label{appx:fig:all_hparam_plots:batch_size}
\end{subfigure}
\hfill{}
\begin{subfigure}[t]{0.44\linewidth}
\centering
\captionsetup{font=scriptsize,labelfont=scriptsize}
\includegraphics[width=\linewidth]{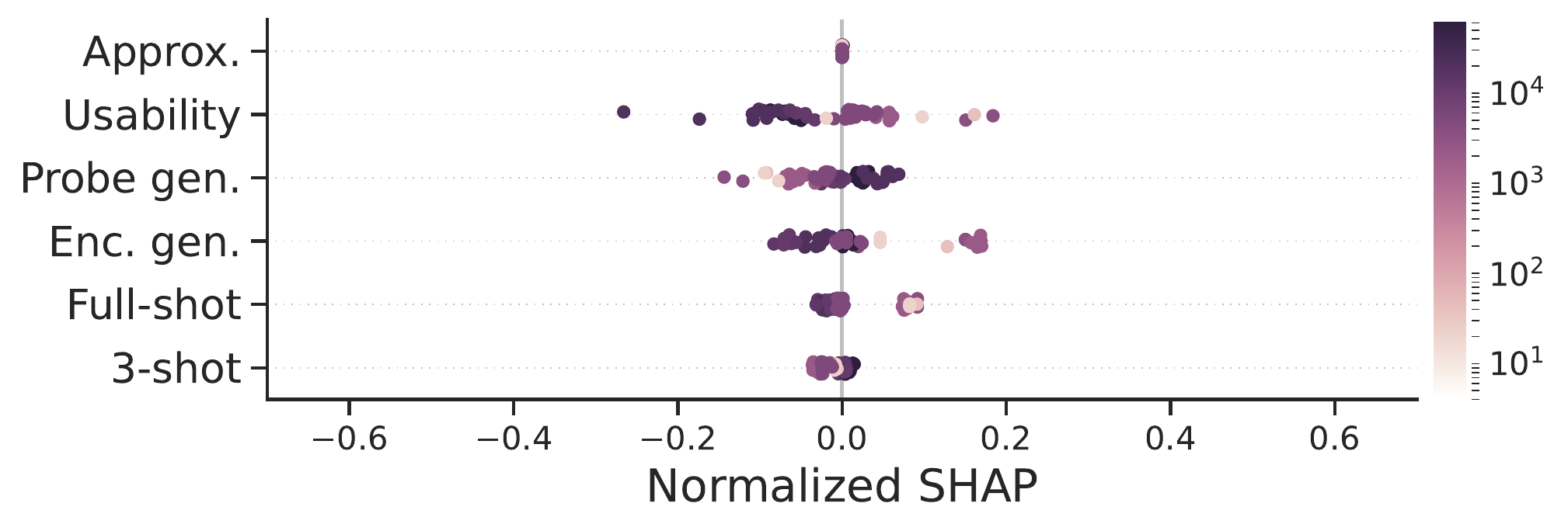}
\caption{N classes for clustering SSL}
\label{appx:fig:all_hparam_plots:n_classes}
\end{subfigure}
\vspace{-0.5em}
\captionsetup{font=small,labelfont=small}
\caption{Impact of important hyperparameters.
Each plot shows a hyperparameter.
Each point shows a different model.
The Y-axis shows the metric, either the risk component or the total risk in the full (``Agg. Risk') and few-shot regime (``3 shot'').
The X-axis shows the normalized SHAP value.
\textbf{Negative values mean that a hyperparameter is beneficial}: it decreases the risk.
Axes cut to $[-0.7,0.7]$.}
\label{appx:fig:all_hparam_plots}
\end{figure}
\pagebreak
\clearpage

\section{Impact of hyperparameters}
\label{appx:sec:res:hparams}

Throughout this section, we will analyze the impact of different hyperparameters on the following metrics: every decomposed risk component (approximation error, usability error, probe generalization error, encoder generalization), the aggregated risk of a linear probe trained on all of ImageNet, and the aggregated risk of a linear probe trained in a 3-shot setting. 
We evaluate the importance of each hyparameter using XGBoost+SHAP, linear models in a controlled setting, and linear models in general settings as described in \cref{appx:sec:details:hparam_impact}.

\paragraph{Impact of each hyperparameter}
A summary of how all hyperparameters impact each metric can be seen in \cref{appx:fig:all_hparam_plots}.
It shows, for each model (point in the scatter plot) how important the value of a certain hyperparameter (the color) is for each of the metrics (Y-axis) as measured by the SHAP value from the XGBoost model normalized by the average value of that metric (X-axis).
Note that every metric is a risk measure, so a lower SHAP value is better. 
For the rest of the section, we will discuss the impact of key hyperparameters on usability and probe generalization. 

\begin{figure}[h]
\centering
\includegraphics[width=0.99\linewidth]{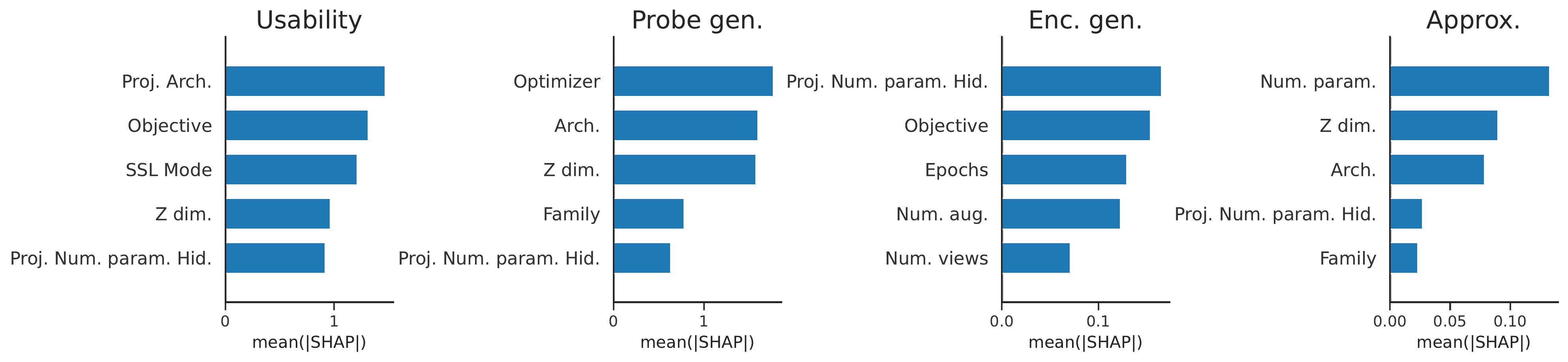}
\captionsetup{font=small,labelfont=small}
\caption{Most important parameters for each risk component as measured by the mean absolute SHAP value of an XGBoost model.}
\label{appcs:fig:param_importance_min}
\end{figure}

\paragraph{Most important  hyperparameter for each metric}
A summary of the most important hyperparameters for each metric can be seen in \cref{appcs:fig:param_importance_min}, which shows the average absolute SHAP value.
We see that usability is mostly impacted by the dimensionality, the projection head (``Proj. Arch.'' and ``Proj. \#param''), and the objective (``objective'' and ``SSL Mode'').
Probe generalization is mostly impacted by the dimensionality, the architecture (``Arch.'' and ``Family''), and the optimizer.
We will investigate each of those more carefully in the rest of the section. 
We see that the approximation error is mostly impacted by the architecture (``Num. param.'', ``Family'', `Z dim.`, and ``Arch.'') as one would expect given that SSL hyperparameters should not impact this error.
We also see that the encoder generalization depends on the augmentations (``augmentations'' and ``views'').
Overall we see that the dimensionality and the projection head seem to be important design choices for all components.

\subsection{Dimensionality}
\label{appx:sec:res:dimensionality}

\Cref{appcs:fig:param_importance_min} and \cref{appx:fig:all_hparam_plots:z_dim}   show that the dimensionality of the representation is a decisive hyperparameter for both the usability and the probe generalization error. 
Let us analyze this in more detail.

\begin{figure}[h]
\centering
\begin{subfigure}[t]{0.60\linewidth}
\centering
%\captionsetup{font=scriptsize,labelfont=scriptsize}
\includegraphics[width=\linewidth]{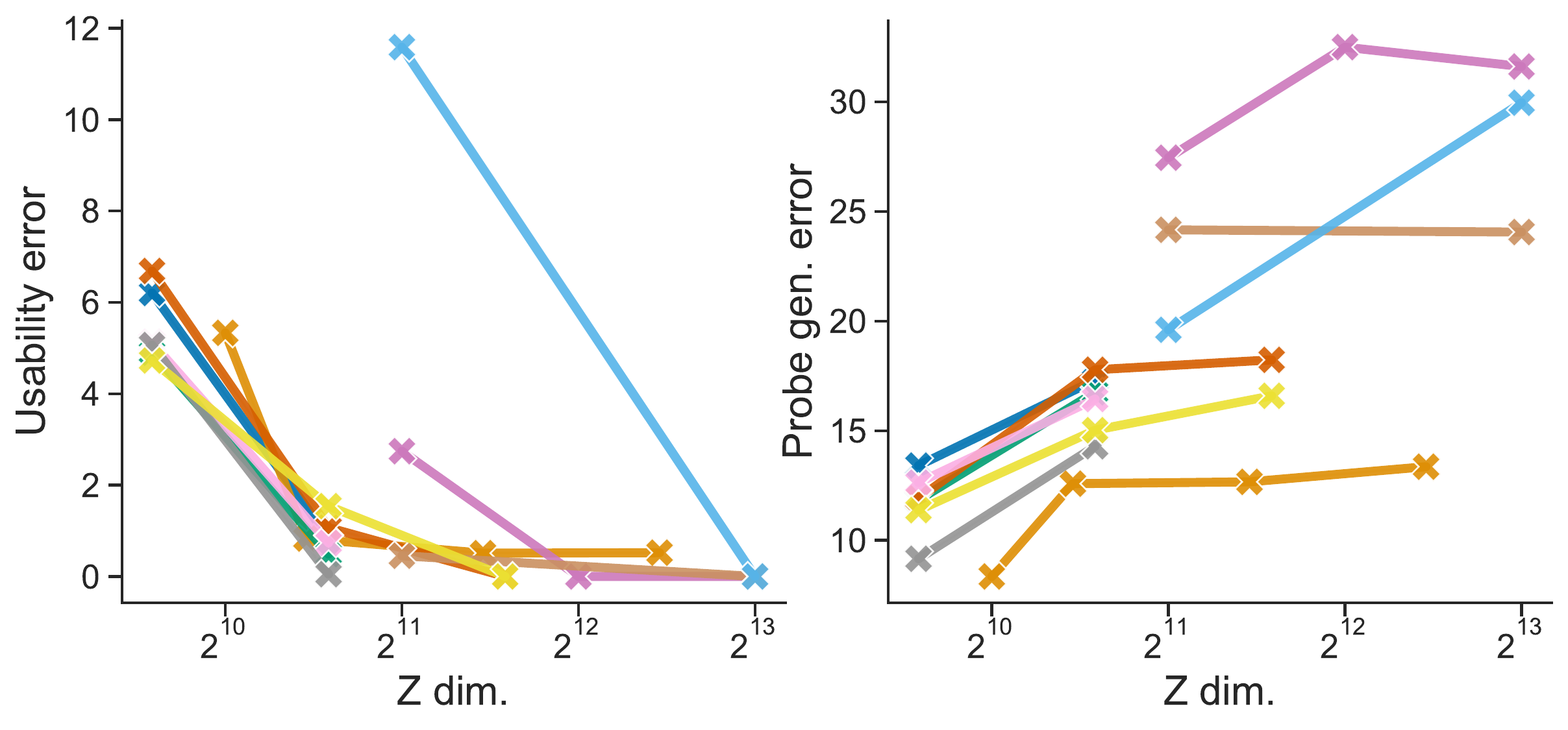}
\caption{Z dim. in the controlled setting}
\label{appx:fig:dim:controlled}
\end{subfigure}
\hfill{}
\begin{subfigure}[t]{0.30\linewidth}
\centering
%\captionsetup{font=scriptsize,labelfont=scriptsize}
\includegraphics[width=\linewidth]{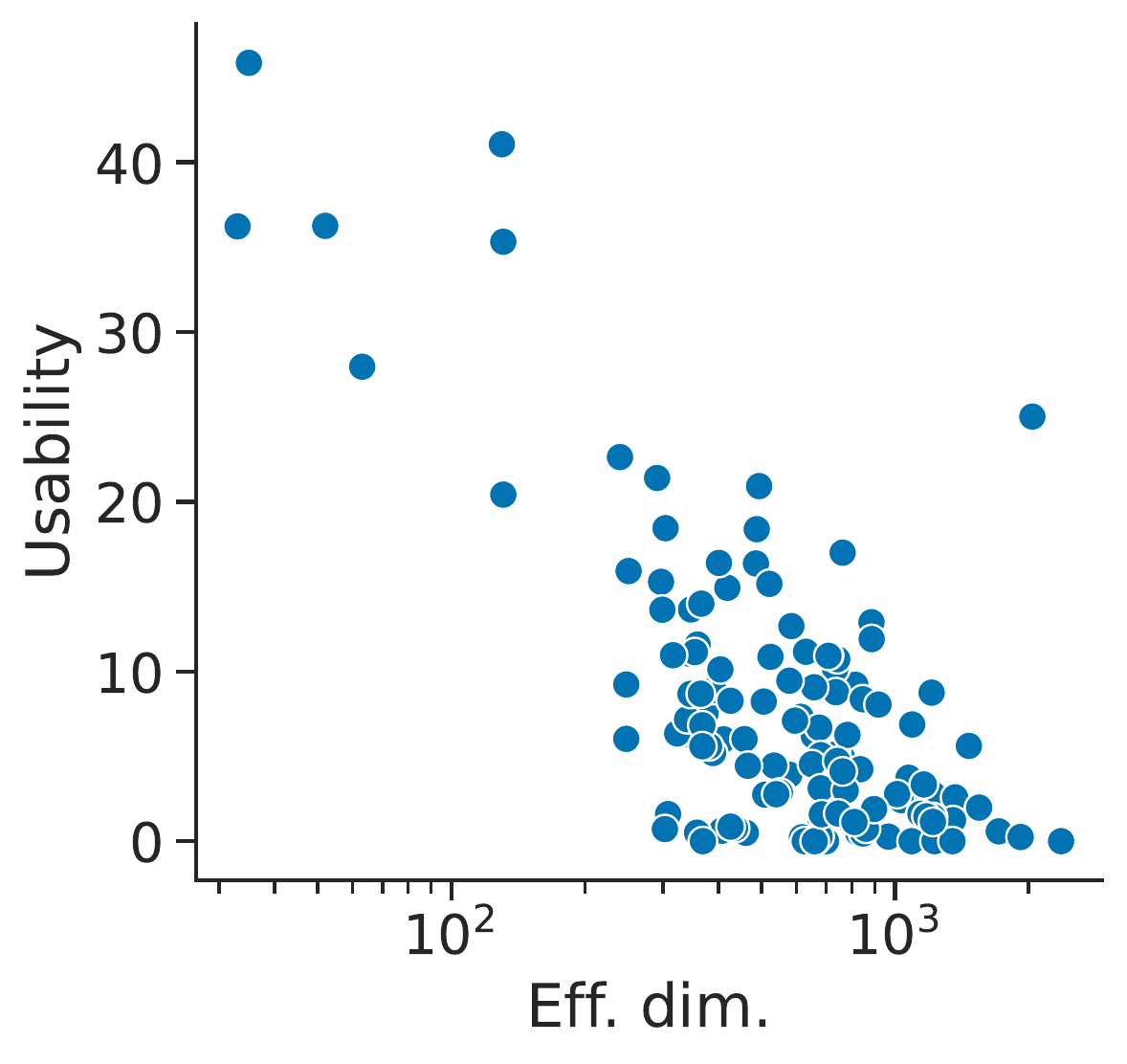}
\caption{Effective Z dim. vs usability}
\label{appx:fig:dim:eff_dim}
\end{subfigure}
\caption{
(a) Impact of Z dimensionality on usability and probe generalization error, when all other hyperparameters are kept the same. 
Each color shows a specific model and the effect that Z dimensionality has on that model. 
(b) Impact of the effective Z dimensionality (the rank of all the representations) on the usability error.
Each point corresponds to a different model with different hyperparameters.
}
\label{appx:fig:z_dim}
\end{figure}

\paragraph{Increasing dimensionality improves usability}
\Cref{appx:fig:all_hparam_plots:z_dim} shows that increasing dimensionality improves usability (decreases usability error).
This is further supported by the controlled analysis plotted in \cref{appx:fig:dim:controlled}.
The coefficient of log(dimensionality) for the controlled linear model is $-3.9$ (CA: pvalue=$4\sci{9}$) for usability.
The impact is also statistically significant for the global linear model.
Although the ambient dimensionality is important, what really matters is actually the effective dimensionality of the representation as shown in \cref{appx:fig:dim:eff_dim} (CA: pvalue=$6\sci{8}$).

The theory from \citet{dubois_improving_2022} suggests why increasing (effective) dimensionality is necessary and sufficient for good usability. 
Namely, they prove that SSL clusters representations by the equivalence classes induced by the training augmentations. 
From those clusters, one can then linearly predict any downstream label that is invariant to the augmentations if and only if the effective dimensionality of the representation is at least the number of classes minus one. 
This is because predicting any downstream labels is equivalent to shattering the $C$ clusters, which by standard statistical learning theory \cite{vapnik_on_1971} is only possible by linear models iff $d = C-1$.
Intuitively, increasing the input dimension increases the capacity of a linear model. 

\paragraph{Increasing dimensionality worsens probe generalization error}
The SHAP+XGBoost analysis (\cref{appx:fig:all_hparam_plots:z_dim}) and the controlled analysis (\cref{appx:fig:dim:controlled}) both show that increasing dimensionality leads to worse probe generalization error.
In particular, the  coefficient of log(dimensionality) for the controlled linear model is $3.8$ (CA: pvalue=$2\sci{9}$) for probe generalization error.
The impact is also statistically significant for the global linear model.

The negative effect that dimensionality error has on probe generalization can be understood in two different ways.
First, by standard statistical learning theory, we expect a smaller dimensionality of the input data to lead to better generalization given that the model can overfit on fewer components.
Second, due to the usability-probe generalization trade-off (\cref{sec:results:tradeoffs}) we expect dimensionality to have the opposite effect as it has on usability.

\begin{figure}[h]
    \centering
    \includegraphics[width=0.7\linewidth]{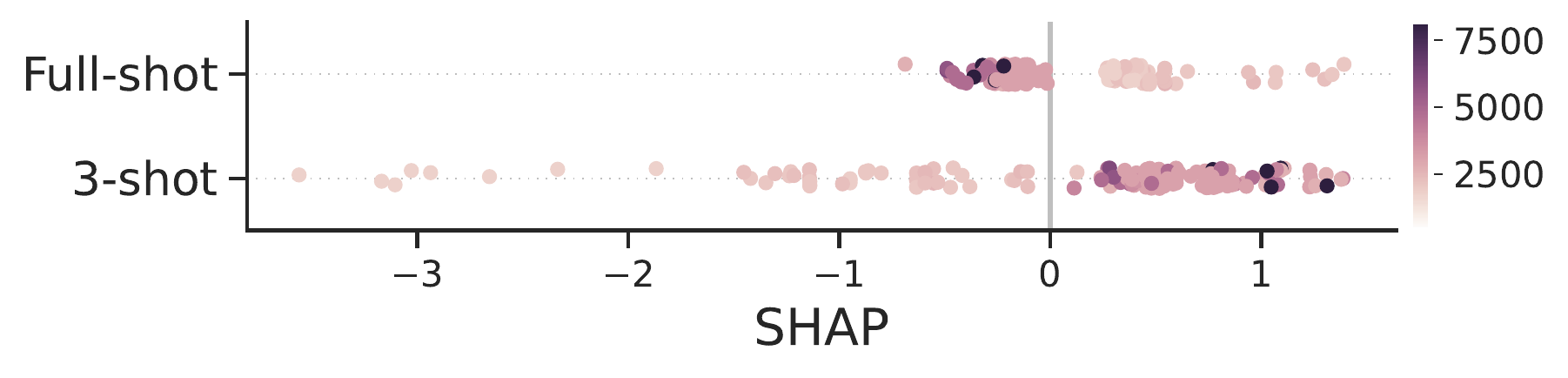}
    \caption{Z dimensionality has a significant impact on the performance in different settings.
    Every point corresponds to a model.
    The color shows the Z dimensionality.
    X-axis is the absolute SHAP value.
    Y-axis shows the performance in the full-shot (``Agg. Risk'') and few-shot (``3-shot'') setting.
    }    \label{appcs:fig:dimensionality:setting}
\end{figure}

\paragraph{Lower dimensional representations are better in few-shot settings}
Given the important impact that dimensionality has on usability and probe generalization, we expect it to also have an important impact on the performance of the representations in different settings due to \cref{sec:res:settings}.
In particular, we expect that lower dimensional representations will perform better in few-shot settings, while higher dimensional representations will perform better in full-shot settings.
\cref{appcs:fig:dimensionality:setting} shows that in the few-shot setting, using a low dimensionality can improve performance by up to 4 accuracy points, while it decreases full-shot performance by up to 1 accuracy point.

\subsection{Data and Augmentations}
\label{appx:sec:res:augmentations}

Let us analyze the impact that the choice of augmentations has on each metric.
One challenge is that there are many different augmentations and most models use the same ones, which makes it challenging to pin down the impact of a single augmentation.
To avoid this issue, we focus on two specific hyperparameters that are related to augmentations.
First, we consider the total number of augmentations used for training the model, which is coarser than the exact augmentations and thus easier to analyze.
Second, we consider the number of views/multicrops \cite{caron_unsupervised_2020} used to pretrain the model.
The advantage of multicrops is that it is the only augmentations for which we have many models that only differ with respect to it.

In \cref{sec:res:hparams:augmentations} we discuss the case of multicrops, here we focus on the total number of type of augmentations (\eg rotation, flipping, cropping, \dots)

\paragraph{Increasing the total number of augmentations likely improves usability}
\cref{appx:fig:all_hparam_plots:n_augmentations} suggests that increasing the number of augmentations might the usability of the representation.
Using the global linear model for quantifying the importance of the log number of augmentations, we have that the coefficient of the log number of augmentations is $-5.3$ (CA: pvalue=$4\sci{2}$). 
This high p-value compared to the effect of the number of views is likely due to the fact that increasing the number of augmentations does not monotonically decrease the number of equivalence classes because the augmentations are not comparable.
For example, a model that uses only auto-augment and cropping would be counted as having only 2 augmentations but those are likely much stronger than using small x- and y-translations and rotations, which would be counted as 3 augmentations.
We thus believe that the effect of increasing augmentation strength is similar to increasing the number of views, but that simply counting the number of augmentations is not an ideal way of quantifying the strength of augmentations.

% \subsection{Pretraining data}
% \label{appx:sec:res:data}

\begin{figure}[h]
    \centering
    \includegraphics[width=0.9\linewidth]{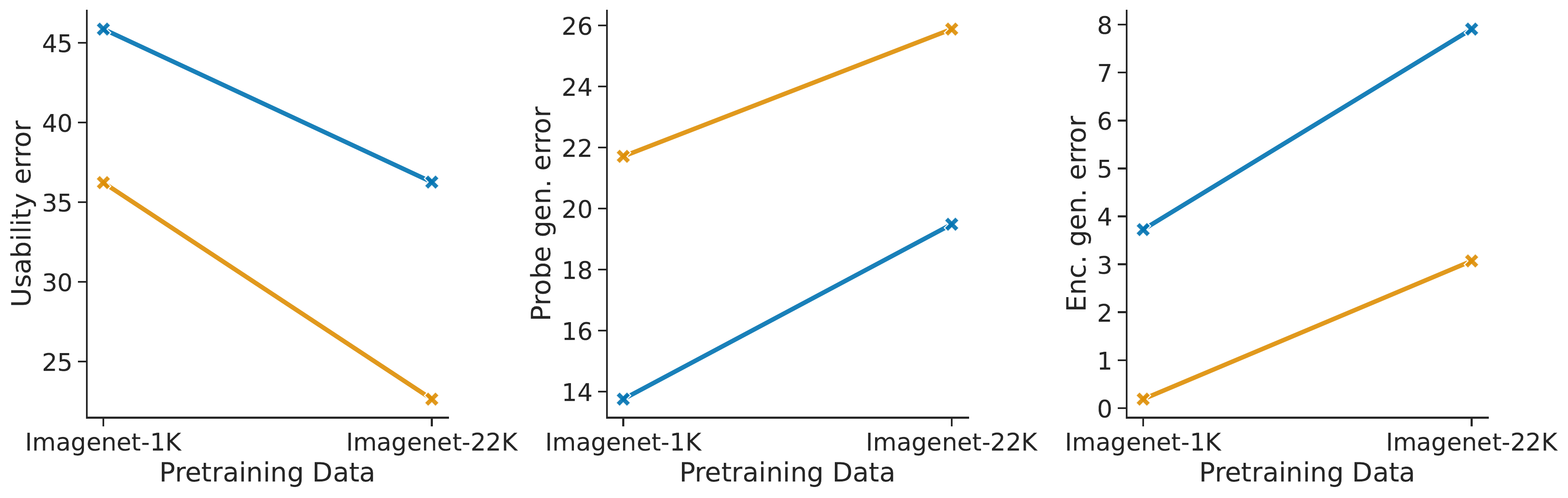}
    \caption{
        Effect of pretraining on ImageNet-22k on usability, probe generalization, and encoder generalization error.
         All other hyperparameters are kept the same. 
    Each color shows a specific model.
}    \label{appcs:fig:data:controlled}
\end{figure}

\paragraph{Pretraining on ImageNet-22k worsens generalization}
\cref{appx:fig:all_hparam_plots:pretraining_data} shows that pretraining on ImageNet-22k worsens both the encoder and the probe generalization error.
This can be seen also from the controlled setting in \cref{appcs:fig:data:controlled}.
This is interesting given that ImageNet-22k is a superset of the standard ImageNet-1k.
This shows that pretraining on additional data can be detrimental to generalization.

\subsection{Architecture}
\label{appx:sec:res:architecture}

% (\cref{fig:projection:projection_shap}).
% In \cref{appx:sec:res:architecture} we show that increasing the number of parameters in the projection also improves usability. %although it can worsen probe generalization.

% \input{figures/projection/projection_shap.tex}

It is well known that using large non-linear projection heads helps \cite{bachman_learning_2019,chen_simple_2020,chen_big_2020}, but it is not clear why it does work.
To our knowledge there are four explanations that have been proposed in the literature for why using at least one non-linear head can help:
\begin{inlinelist}
\item to avoid perfect invariance/alignment, which helps if the augmentations are stronger than desired \cite{chen_simple_2020,gupta_understanding_2022,appalaraju_towards_2020},
\item to avoid dimensionality collapse \cite{jing_understanding_2022},
\item to be able to learn the optimal pseudo-label that should be predicted to ensure linearly predictability \cite{dubois_improving_2022},
\item to avoid complete collapsing in non-contrastive learning \cite{chen_simpler_2021}.
\end{inlinelist}
All of those explanations suggest that adding a non-linear projection head would improve the usability of the representation. 
%\Cref{appcs:fig:param_importance_min} shows that the size of the projection head is crucial for usability as expected (both the architecture and the number of parameters), but is actually important for every component.

% \input{figures/projection/proj_controlledt}

\paragraph{Large projection heads improve usability}
\Cref{appcs:fig:param_importance_min} shows that the size of the projection head is crucial
for usability as expected (both the architecture and the number of parameters).
\Cref{appx:fig:all_hparam_plots:projection2_arch} and \cref{appx:fig:all_hparam_plots:projection_nparameters_hidden}  shows that using a large MLP projection head greatly improves usability.
Quantitatively, we have that the global linear model predicts a coefficient of $-8.6 \pm 2.6$ for using an MLP projection instead of no projection (GLA: p-value $1\sci{3}$) and a coefficient of $-0.68 \pm 0.28$ for the log of the number of projection parameters (GLA: p-value $2\sci{2}$).
The beneficial impact of using a larger projection head on usability is even more clear from the controlled setting seen in \cref{fig:projection:controlled} (CA: p-value $9\sci{12}$).

This empirically support our hypotheses that a larger projection should improve usability as suggested by previous literature. 
This still does not explain which of the four previous explanations is (more) correct.
As a partial answer to this question we consider the effect that projections heads have on effective dimensionality, and we have that using a linear projection head significantly improves effective dimensionality (GLA: pvalue=$3\sci{9}$) but a non-linear projection head is not significantly different from the linear one. 
This suggests that \citepos{jing_understanding_2022} hypothesis about dimensionality collapse explains some of the performance gains but not all.
Furthemore, we did not see any significant impact on alignment as suggested by \cite{gupta_understanding_2022} or gains from using one-linear projection head as suggested by \cite{dubois_improving_2022}.
This shows that our understanding of the impact of non-linear projection heads is still lacking.

\paragraph{MLP projection improves probe generalization}
\Cref{appx:fig:all_hparam_plots:projection2_arch} shows that using an MLP head is actually somewhat beneficial for \textbf{all} metrics. 
In particular, \cref{fig:projection:controlled} shows that MLP projection heads also typically improve probe generalization (CA p-value $5\sci{3}$). 
This shows that using an MLP projection head is one effective way to overcome the usability-probe generalization tradeoff.
The impact that a non-linear MLP projection head has on probe generalization cannot be predicted by the four previous hypotheses.
This further suggests that we do not completely understand why large non-linear projection heads improve performance.
% To try to improve this understanding we used our global linear model to predict the impact that larger projection heads (as measured by the log of the number of parameters) have on the statistics from \cref{appx:sec:results:statistics}, and found that it not only improves the effective dimensionality (p-value $4\sci{5}$) but also the alignment ($8\sci{3}$) of the representation. 
% Further investigation should be done to understand why this happens.

% \paragraph{MLP projection heads are never worse}
% Importantly, \cref{appx:fig:all_hparam_plots:projection2_arch} shows that using an MLP projection improves usability and probe generalization

% \subsection{Architecture}
% \label{appx:sec:res:architecture}

\Cref{appcs:fig:param_importance_min} shows that the architecture (family, number of parameters, and patch size) is really important for the probe generalization and approximation error.

\paragraph{Smaller patch sizes for ViTs is uniformly better}
\Cref{appx:fig:all_hparam_plots:patch_size} shows that smaller patch sizes for ViT are uniformly better but is especially important for the approximation and usability error.

% \paragraph{ResNets have better approximation error}
% \Cref{appx:fig:all_hparam_plots:family} shows that ResNets are better for the approximation error.
% Our global linear layer quantifies those gains $0.20\pm 0.08$ (p-value $1\sci{2}$). 

\subsection{Objective}
\label{appx:sec:res:objective}

Let us analyze the impact that the choice of SSL objective has on each metric.
One difficulty to do so is that there are many objectives and so (1) it is hard to analyze them simultaneously, and (2) there are only a few pretrained models for each objective.
To avoid both of those problems we aggregate the SSL objectives into the 6 coarser clusters described in \cref{appx:sec:details:pretrained} (transform, contrastive, clustering, 
 siamese, generative, hierarchical).

\begin{figure}[h]
\centering
%
%
%\captionsetup{font=scriptsize,labelfont=scriptsize}
\includegraphics[width=0.9\linewidth]{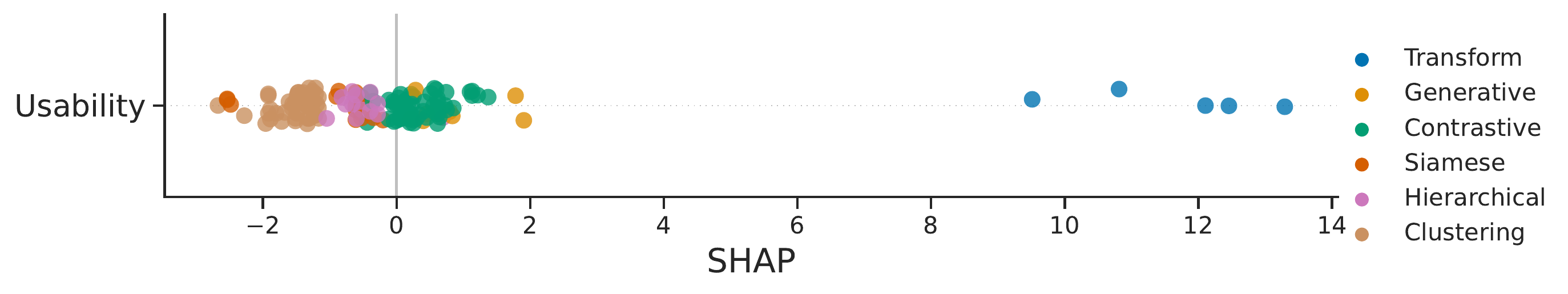}
\caption{Effect measured by SHAP}
\label{appx:fig:obj:usability_shap}
\caption{
SSL mode has an important impact on the usability error. 
(a) Average usability error for models of each SSL mode without considering potential confounders.
(b) SHAP values of each model color coded by the SSL mode.
}
\label{appx:fig:objective}
\end{figure}

\paragraph{Objectives that are generative or predict the transformation worsen the usability}
\Cref{appx:fig:all_hparam_plots:ssl_mode} shows that the SSL objective and the coarser SSL mode have an important impact on usability error.
\Cref{appx:fig:obj:usability_shap} shows more precisely the effect on usability. 
We see that the generative models and the ones that predict the transformation have much worse usability. 
The p-values as given by the global linear models are respectively $1\sci{4}$ and $1\sci{2}$.\footnote{
The impact of having an objective that predicts the transformation is not as significant as what we would expect from \cref{appx:fig:objective} because it is highly correlated with the publication year which we have to control for.
}
In contrast, clustering objectives significantly improve usability.

\begin{figure}[h]
    \centering
    \includegraphics[width=0.8\linewidth]{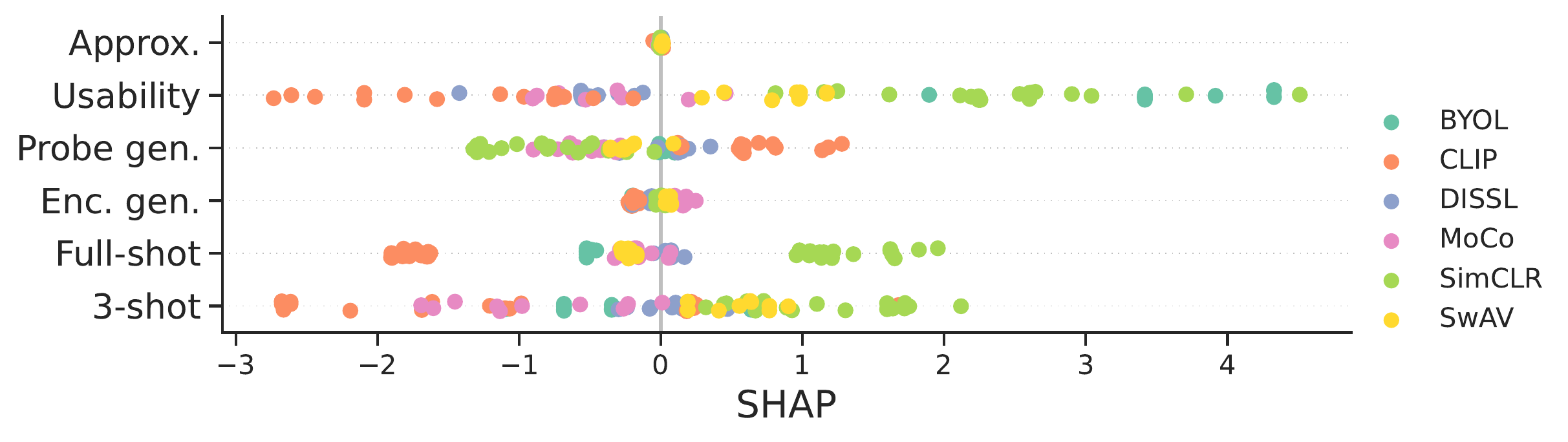}
    \caption{
    Effect of fine-grained objective functions on each risk component. 
    We only show objectives for which there are at least 7 models, to avoid over interpreting the results. 
      Every point corresponds to a model.
    The color shows the Z dimensionality.
    X-axis is the absolute SHAP value.
    The Y-axis shows the metric, either the risk component or the total risk in the full (``Agg. Risk') and few-shot regime (``3 shot'').
    }    \label{appcs:fig:obj:all_objectives}
\end{figure}

\paragraph{Finer grain analysis of objectives}
\Cref{appcs:fig:obj:all_objectives} shows the impact of the exact objective functions on each metric.
To make sure that the results are meaningful, we only show objectives for which we have at least 7 models.
We see that CLIP is particularly good for usability and full-shot risk, while MOCO is good in the few-shot regime. 
We also see that SimCLR a weak objective w.r.t. to few- and full-shot performance.
This shows that the newer objective brings some meaningful improvement compared to SimCLR.

\subsection{Optimization}
\label{appx:sec:res:optimization}

\begin{figure}[h]
    \centering
    \includegraphics[width=0.9\linewidth]{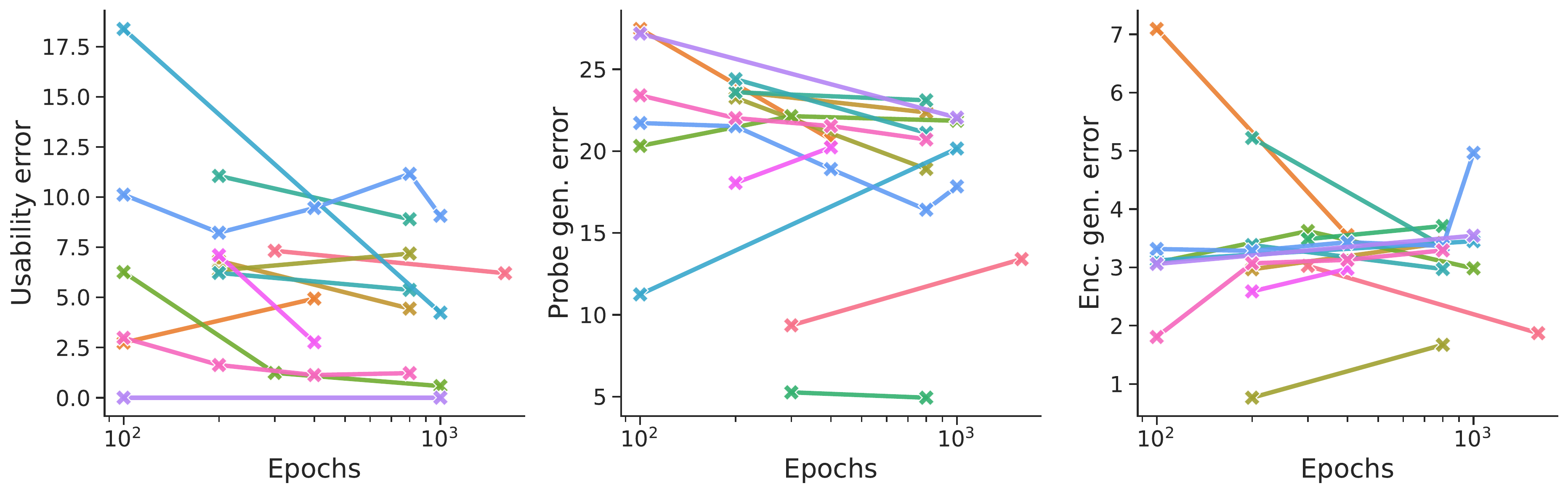}
    \caption{
        Effect of the projection head on usability, probe generalization, and encoder generalization error.
         All other hyperparameters are kept the same. 
    Each color shows a specific model.
}    \label{appcs:fig:optimization:controlled_epochs}
\end{figure}

\paragraph{Longer training improves usability and probe generalization}
\Cref{appx:fig:all_hparam_plots:epochs} suggests that increasing the number of epochs improves usability and probe generalization but might have a negative impact on encoder generalization.
A similar trend can also be somewhat seen from the controlled setting in \cref{appcs:fig:optimization:controlled_epochs} for usability (CA p-value: $2\sci{3}$, coefficient: $-1.37 \pm 0.55$) and to a lesser extent for probe generalization (CA coefficient: $-0.58  \pm 0.57$, p-value: $0.3$).
We see that for the encoder generalization, it is not very clear, for some models it improves, and for others, it makes it worse.

The improvements in usability and probe generalization can be partially understood from the fact that longer training with the proper SSL log loss with give rise to the collapse of equivalent representations \cite{dubois_improving_2022}, which should improve downstream sample efficiency and linear predictability.
The potential worsening or improvement of encoder generalization is likely due to the fact that at the beginning, training for longer allows you to better generalize but then the model starts overfitting given that you see multiple times the same examples.

\paragraph{Adam and AdamW improve probe generalization}
\cref{appx:fig:all_hparam_plots:optimizer} suggests that Adam and AdamW should be favored in both the full- and few-shot settings. Indeed, those optimizers seem to improve probe generalization.

\paragraph{Larger batch sizes can be beneficial for all components}
\cref{appx:fig:all_hparam_plots:batch_size} suggests that larger batch sizes can be beneficial but our global linear layer did  not recognize the impact as being significant.

% \paragraph{High learning rates might be detrimental }
% \cref{appx:fig:all_hparam_plots:learning_rate} suggests that high learning rates can be mildly detrimental but our global linear layer did nevertheless not recognize the impact as being significant.

% \paragraph{Weight decay does not have much impact}
% \cref{appx:fig:all_hparam_plots:weight_decay} suggests that weight decay (in standard ranges) is not a crucial component.
% Our global linear layer did not recognize any significant impact.

\subsection{Other}
\label{appx:sec:res:other}

\paragraph{The number of classes for clustering objectives can improve usability}
\cref{appx:fig:all_hparam_plots:n_classes} suggests that increasing the number of classes for clustering objectives (\eg teacher's output in SwAV, DINO, or DISSL) can improve usability at the detriment of probe generalization but our GLA did not recognize the impact as significant.
Both of those can be understood by \citepos{dubois_improving_2022} ISSL theory.
First, fewer equivalences classes mean that you need to see fewer downstream samples.
Second, if there are fewer teacher's classes than equivalence classes then the model might collapse examples that can differ in downstream labels, which will negatively impact usability.

\clearpage
\newpage

\section{All raw results}
\label{appx:sec:raw_results}

\subsection{Radar charts}
\label{appx:sec:results:all_radar}
\begin{figure}[H]
    \centering
\includegraphics[height=0.8\textheight]{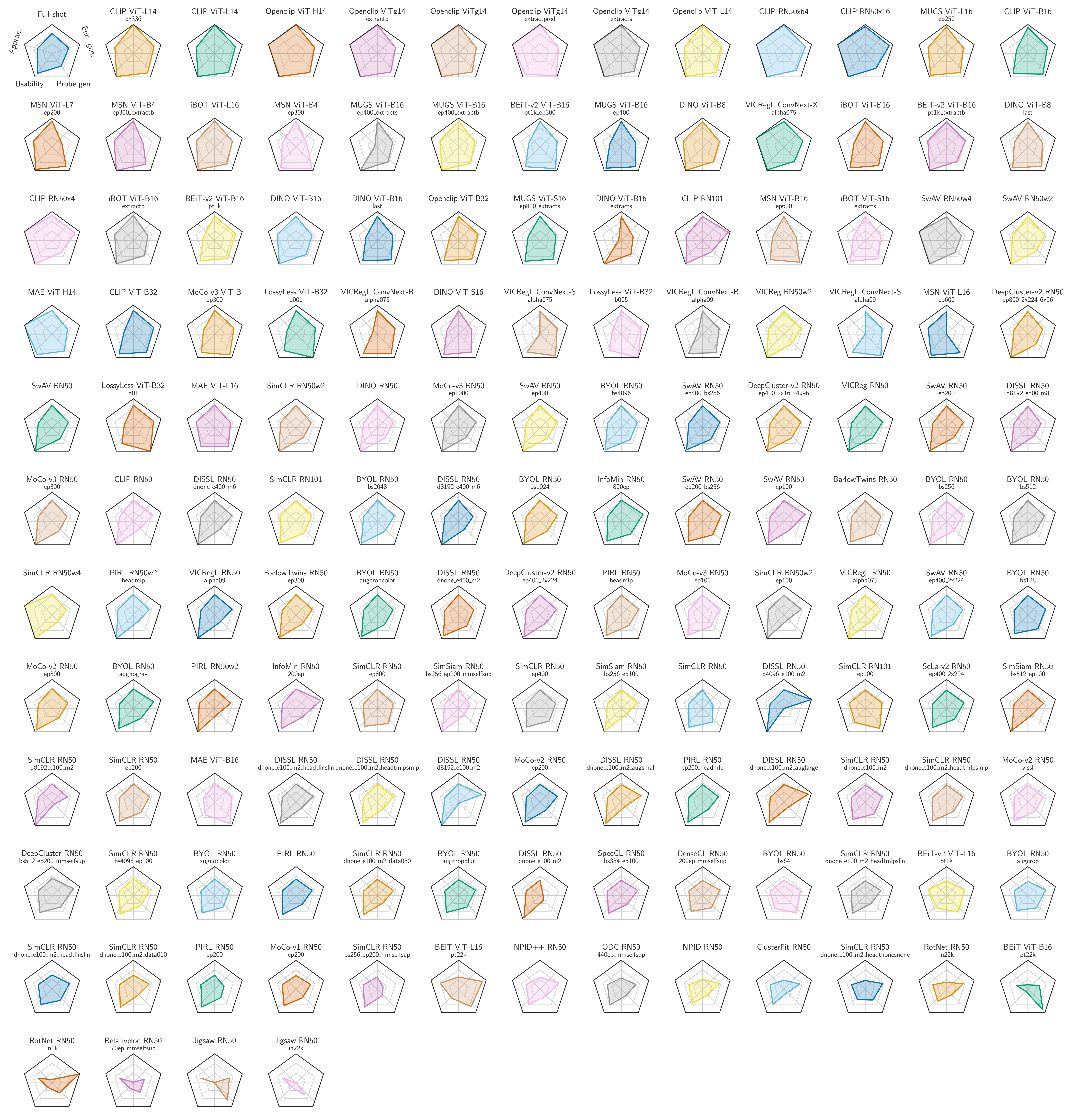} 
\caption{
\small
All risk components.
Starting from the top, %and moving clock-wise 
axes respectively show the standard linear probing ImageNet risk (``Agg. Risk''),  \encgen{} (``Enc. Gen.''),  \probegen{} (``Probe Gen.''), \usability{} (``Usability''), and \approxerr{} (``Approx.'').
Values are min-max scaled and substracted to 1 so that the worst model gets a 0 and the best gets a 1 (vertex).
The top left plot shows the average over models, all other plots show a specific model described by its title (SSL objective and architecture) and subtitle (additional hyperparameters corresponding to ``other'' in \cref{appx:tab:all_results}). 
 Colors are meaningless.
}
\label{appcs:fig:radar}
\end{figure}

\Cref{appcs:fig:radar} shows the relative risk component of nearly every evaluated model.
We do not show \Nnan{} models for which we did not find a supervised model with the same architecture, as we could not compute the approximation error for those models.

\subsection{Table}
\label{appx:sec:results:all_results}

The radar charts from \Cref{appcs:fig:radar} are useful to get a quick overview of each model but are not quantitative and do not allow comparison between risk components as each axes are normalized.
\Cref{appx:tab:all_results} provides all the raw metrics.

\begin{tiny}
\begin{longtable}[h]{l@{\hspace{0.75\tabcolsep}}lll@{}rrrrrrrrr}
\label{appx:tab:all_results} \\
\toprule
 &  &  &  & \multicolumn{4}{c}{\textbf{Risk Component}} & \multicolumn{5}{c}{\textbf{Aggregated Error}} \\
 \cmidrule(lr){5-8}  \cmidrule(lr){9-13} \ 
\textbf{ Objective } & \textbf{  Arch. } & \textbf{  Epochs } & \textbf{  Other } & \textbf{ Approx. } & \textbf{ Usability } & \textbf{ Probe gen. } & \textbf{ Enc. gen. } & \textbf{ 100\% } & \textbf{ 30 Shot } & \textbf{ 1\% } & \textbf{ 5 Shot } & \textbf{ 3 Shot } \\
\midrule
\endfirsthead
\toprule
 &  &  &  & \multicolumn{4}{c}{\textbf{Risk Component}} & \multicolumn{5}{c}{\textbf{Aggregated Error}} \\
 \cmidrule(lr){5-8}  \cmidrule(lr){9-13} \ 
\textbf{ Objective } & \textbf{  Arch. } & \textbf{  Epochs } & \textbf{  Other } & \textbf{ Approx. } & \textbf{ Usability } & \textbf{ Probe gen. } & \textbf{ Enc. gen. } & \textbf{ 100\% } & \textbf{ 30 Shot } & \textbf{ 1\% } & \textbf{ 5 Shot } & \textbf{ 3 Shot } \\
\midrule
\endhead
\midrule
\multicolumn{13}{r}{Continued on next page} \\
\midrule
\endfoot
\bottomrule
\endlastfoot
\multirow[t]{2}{*}{BEiT} & ViT-B16 & 800 & \miniscule{pt22k} & 1.00 & 41.06 & 10.53 & 4.61 & 57.19 & 94.35 & 89.67 & 94.35 & 95.92 \\
 & ViT-L16 & 800 & \miniscule{pt22k} & 0.55 & 27.96 & 13.96 & 1.09 & 43.55 & 93.60 & 87.08 & 93.60 & 95.72 \\
\midrule
\multirow[t]{4}{*}{BEiT-v2} & \multirow[t]{3}{*}{ViT-B16} & 300 & \miniscule{pt1k\_ep300} & 1.00 & 7.32 & 9.36 & 3.03 & 20.71 & 37.04 & 31.42 & 37.04 & 41.32 \\
 &  & \multirow[t]{2}{*}{1600} & \miniscule{pt1k\_extractb} & 0.54 & 0.84 & 17.31 & 2.54 & 21.23 & 41.93 & 34.93 & 41.93 & 47.53 \\
 &  &  & \miniscule{pt1k} & 1.00 & 6.21 & 13.41 & 1.87 & 22.48 & 41.53 & 35.42 & 41.53 & 46.43 \\
 & ViT-L16 & 1600 & \miniscule{pt1k} & 0.55 & 21.40 & 15.03 & 2.78 & 39.77 & 72.85 & 63.25 & 72.85 & 77.95 \\
\midrule
\multirow[t]{12}{*}{BYOL} & \multirow[t]{12}{*}{RN50} & \multirow[t]{12}{*}{1000} & \miniscule{augcropblur} & 0.85 & 12.90 & 20.26 & 3.13 & 37.14 & 71.59 & 63.67 & 71.59 & 75.81 \\
 &  &  & \miniscule{augcropcolor} & 0.85 & 3.74 & 21.98 & 3.46 & 30.02 & 60.70 & 52.50 & 60.70 & 66.12 \\
 &  &  & \miniscule{augcrop} & 0.85 & 17.00 & 19.54 & 2.86 & 40.25 & 75.64 & 67.86 & 75.64 & 79.66 \\
 &  &  & \miniscule{augnocolor} & 0.85 & 11.91 & 20.15 & 3.63 & 36.53 & 71.26 & 63.04 & 71.26 & 75.20 \\
 &  &  & \miniscule{augnogray} & 0.85 & 6.88 & 21.65 & 2.10 & 31.47 & 60.91 & 52.50 & 60.91 & 66.58 \\
 &  &  & \miniscule{bs1024} & 0.85 & 2.67 & 21.97 & 3.02 & 28.51 & 57.77 & 48.98 & 57.77 & 63.74 \\
 &  &  & \miniscule{bs128} & 0.85 & 9.22 & 17.70 & 2.86 & 30.63 & 59.80 & 51.73 & 59.80 & 65.09 \\
 &  &  & \miniscule{bs2048} & 0.85 & 2.72 & 21.82 & 3.07 & 28.45 & 57.82 & 49.05 & 57.82 & 63.64 \\
 &  &  & \miniscule{bs256} & 0.85 & 2.77 & 22.23 & 3.26 & 29.11 & 58.57 & 49.89 & 58.57 & 64.47 \\
 &  &  & \miniscule{bs4096} & 0.85 & 2.58 & 19.72 & 3.35 & 26.50 & 54.57 & 45.55 & 54.57 & 61.20 \\
 &  &  & \miniscule{bs512} & 0.85 & 3.34 & 21.31 & 3.19 & 28.68 & 57.53 & 49.22 & 57.53 & 63.31 \\
 &  &  & \miniscule{bs64} & 0.85 & 20.92 & 14.06 & 3.15 & 38.98 & 69.11 & 61.30 & 69.11 & 73.21 \\
\midrule
\multirow[t]{2}{*}{BarlowTwins} & \multirow[t]{2}{*}{RN50} & 300 & \miniscule{ep300} & 0.85 & 2.42 & 23.10 & 3.27 & 29.63 & 59.56 & 50.98 & 59.56 & 65.59 \\
 &  & 1000 & \miniscule{} & 0.85 & 5.61 & 18.93 & 3.43 & 28.82 & 57.33 & 49.09 & 57.33 & 63.30 \\
\midrule
\multirow[t]{13}{*}{CLIP} & RN101 & 32 & \miniscule{} & 0.71 & 1.60 & 20.33 & 0.57 & 23.21 & 51.35 & 41.62 & 51.35 & 58.25 \\
 & RN50 & 32 & \miniscule{} & 0.85 & 0.71 & 23.98 & 2.32 & 27.85 & 56.70 & 46.41 & 56.70 & 63.79 \\
 & RN50x16 & 32 & \miniscule{} & 0.00 & 0.62 & 16.62 & 1.06 & 18.30 & 41.21 & 32.64 & 41.21 & 48.25 \\
 & RN50x4 & 32 & \miniscule{} & 0.00 & 0.50 & 19.63 & 1.39 & 21.52 & 46.98 & 37.44 & 46.98 & 53.87 \\
 & RN50x64 & 32 & \miniscule{} & 0.00 & 0.49 & 14.49 & 1.73 & 16.72 & 35.74 & 28.16 & 35.74 & 42.28 \\
 & ViT-B16 & 32 & \miniscule{} & 1.00 & 6.30 & 10.51 & 2.28 & 20.08 & 40.65 & 32.79 & 40.65 & 46.88 \\
 & ViT-B32 & 32 & \miniscule{} & 1.13 & 7.53 & 13.13 & 2.11 & 23.90 & 47.55 & 39.03 & 47.55 & 53.62 \\
 & \multirow[t]{6}{*}{ViT-L14} & \multirow[t]{6}{*}{32} & \miniscule{px336\_extractb} & nan & nan & 12.37 & 2.09 & 14.95 & 32.81 & 25.36 & 32.81 & 39.31 \\
 &  &  & \miniscule{px336\_extractpredcls} & nan & nan & 11.52 & 2.14 & 14.95 & 30.61 & 24.51 & 30.61 & 36.44 \\
 &  &  & \miniscule{px336\_extractpred} & nan & nan & 9.09 & 1.52 & 15.10 & 30.52 & 24.14 & 30.52 & 35.94 \\
 &  &  & \miniscule{px336\_extracts} & nan & nan & 12.18 & 2.25 & 14.93 & 35.56 & 27.25 & 35.56 & 42.52 \\
 &  &  & \miniscule{px336} & 0.55 & 0.86 & 11.60 & 2.00 & 15.01 & 31.05 & 24.78 & 31.05 & 37.07 \\
 &  &  & \miniscule{} & 0.55 & 0.77 & 12.12 & 2.02 & 15.46 & 32.08 & 25.53 & 32.08 & 37.96 \\
\midrule
ClusterFit & RN50 & 105 & \miniscule{} & 0.85 & 16.36 & 28.48 & 3.38 & 49.07 & 84.53 & 77.58 & 84.53 & 88.32 \\
\midrule
\multirow[t]{10}{*}{DINO} & RN50 & 800 & \miniscule{} & 0.85 & 0.23 & 21.42 & 3.34 & 25.83 & 57.40 & 47.11 & 57.40 & 64.06 \\
 & \multirow[t]{3}{*}{ViT-B16} & \multirow[t]{3}{*}{400} & \miniscule{extracts} & 1.55 & 0.00 & 18.23 & 4.53 & 23.57 & 41.79 & 35.15 & 41.79 & 47.39 \\
 &  &  & \miniscule{last} & 1.00 & 6.69 & 11.91 & 3.51 & 23.10 & 37.44 & 32.55 & 37.44 & 41.68 \\
 &  &  & \miniscule{} & 0.54 & 1.07 & 17.78 & 3.38 & 22.76 & 40.50 & 34.05 & 40.50 & 46.20 \\
 & \multirow[t]{2}{*}{ViT-B8} & \multirow[t]{2}{*}{300} & \miniscule{last} & 0.86 & 4.90 & 11.83 & 3.83 & 21.42 & 34.23 & 29.74 & 34.23 & 38.21 \\
 &  &  & \miniscule{} & 0.51 & 0.49 & 16.75 & 3.13 & 20.88 & 36.78 & 30.66 & 36.78 & 41.74 \\
 & \multirow[t]{3}{*}{ViT-S16} & \multirow[t]{3}{*}{800} & \miniscule{extractb} & nan & nan & 10.44 & 4.10 & 25.11 & 44.17 & 37.22 & 44.17 & 50.39 \\
 &  &  & \miniscule{last} & nan & nan & 4.29 & 3.81 & 24.44 & 40.60 & 35.15 & 40.60 & 45.31 \\
 &  &  & \miniscule{} & 0.96 & 6.00 & 13.48 & 4.16 & 24.60 & 46.43 & 39.13 & 46.43 & 52.87 \\
 & ViT-S8 & 800 & \miniscule{last} & nan & nan & 4.45 & 3.82 & 21.79 & 34.26 & 29.57 & 34.26 & 38.05 \\
\midrule
\multirow[t]{11}{*}{DISSL} & \multirow[t]{11}{*}{RN50} & \multirow[t]{7}{*}{100} & \miniscule{d4096\_e100\_m2} & 0.85 & 0.00 & 32.50 & 0.00 & 32.85 & 66.94 & 57.74 & 66.94 & 72.82 \\
 &  &  & \miniscule{d8192\_e100\_m2} & 0.85 & 0.00 & 31.62 & 1.30 & 33.58 & 66.41 & 57.16 & 66.41 & 72.34 \\
 &  &  & \miniscule{dnone\_e100\_m2\_auglarge} & 0.85 & 6.01 & 26.74 & 0.99 & 34.59 & 70.10 & 60.75 & 70.10 & 75.77 \\
 &  &  & \miniscule{dnone\_e100\_m2\_augsmall} & 0.85 & 3.91 & 27.56 & 2.25 & 34.57 & 69.29 & 59.76 & 69.29 & 75.23 \\
 &  &  & \miniscule{dnone\_e100\_m2\_headtlinslin} & 0.85 & 4.45 & 25.52 & 3.02 & 33.84 & 68.67 & 58.96 & 68.67 & 74.55 \\
 &  &  & \miniscule{dnone\_e100\_m2\_headtmlpsmlp} & 0.85 & 5.20 & 24.74 & 3.18 & 33.96 & 70.44 & 60.86 & 70.44 & 75.88 \\
 &  &  & \miniscule{dnone\_e100\_m2} & 0.85 & 2.74 & 27.46 & 7.09 & 38.14 & 68.82 & 59.30 & 68.82 & 74.63 \\
 &  & \multirow[t]{4}{*}{400} & \miniscule{d8192\_e400\_m6} & 0.85 & 0.00 & 24.06 & 3.82 & 28.34 & 60.59 & 50.37 & 60.59 & 67.70 \\
 &  &  & \miniscule{d8192\_e800\_m8} & 0.85 & 0.00 & 23.42 & 4.12 & 28.00 & 61.12 & 50.86 & 61.12 & 68.26 \\
 &  &  & \miniscule{dnone\_e400\_m2} & 0.85 & 4.94 & 20.71 & 3.55 & 30.05 & 64.08 & 53.60 & 64.08 & 70.85 \\
 &  &  & \miniscule{dnone\_e400\_m6} & 0.85 & 0.45 & 24.17 & 2.91 & 28.38 & 64.15 & 53.24 & 64.15 & 71.58 \\
\midrule
DeepCluster & RN50 & 200 & \miniscule{bs512\_ep200\_mmselfsup} & 0.85 & 12.67 & 20.27 & 1.72 & 35.51 & 71.36 & 62.02 & 71.36 & 76.80 \\
\midrule
\multirow[t]{3}{*}{DeepCluster-v2} & \multirow[t]{3}{*}{RN50} & \multirow[t]{2}{*}{400} & \miniscule{ep400\_2x160\_4x96} & 0.85 & 1.61 & 21.44 & 3.09 & 26.99 & 56.57 & 47.37 & 56.57 & 62.77 \\
 &  &  & \miniscule{ep400\_2x224} & 0.85 & 2.87 & 23.68 & 3.15 & 30.55 & 61.95 & 53.48 & 61.95 & 67.89 \\
 &  & 800 & \miniscule{ep800\_2x224\_6x96} & 0.85 & 0.27 & 21.30 & 3.63 & 26.05 & 55.37 & 45.39 & 55.37 & 62.51 \\
\midrule
DenseCL & RN50 & 200 & \miniscule{200ep\_mmselfsup} & 0.85 & 15.28 & 19.44 & 3.00 & 38.57 & 70.52 & 63.38 & 70.52 & 74.92 \\
\midrule
\multirow[t]{2}{*}{InfoMin} & \multirow[t]{2}{*}{RN50} & 200 & \miniscule{200ep} & 0.85 & 6.35 & 23.26 & 0.76 & 31.22 & 64.73 & 56.64 & 64.73 & 69.78 \\
 &  & 800 & \miniscule{800ep} & 0.85 & 7.19 & 18.91 & 1.67 & 28.62 & 55.97 & 49.43 & 55.97 & 60.02 \\
\midrule
\multirow[t]{2}{*}{Jigsaw} & \multirow[t]{2}{*}{RN50} & \multirow[t]{2}{*}{105} & \miniscule{in22k} & 0.85 & 36.26 & 19.48 & 7.91 & 64.49 & 92.19 & 87.38 & 92.19 & 94.02 \\
 &  &  & \miniscule{} & 0.85 & 45.86 & 13.76 & 3.72 & 64.19 & 94.43 & 90.80 & 94.43 & 95.85 \\
\midrule
\multirow[t]{3}{*}{LossyLess} & \multirow[t]{3}{*}{ViT-B32} & \multirow[t]{3}{*}{32} & \miniscule{b001} & 1.13 & 13.65 & 7.43 & 2.38 & 24.59 & 46.72 & 38.44 & 46.72 & 52.79 \\
 &  &  & \miniscule{b005} & 1.13 & 13.99 & 7.53 & 2.21 & 24.86 & 47.04 & 38.92 & 47.04 & 53.18 \\
 &  &  & \miniscule{b01} & 1.13 & 14.93 & 7.55 & 2.18 & 25.80 & 47.83 & 39.47 & 47.83 & 53.90 \\
\midrule
\multirow[t]{3}{*}{MAE} & ViT-B16 & 1600 & \miniscule{} & 1.00 & 20.42 & 9.46 & 3.12 & 34.00 & 72.67 & 62.68 & 72.67 & 78.27 \\
 & ViT-H14 & 1600 & \miniscule{} & 0.00 & 6.03 & 14.51 & 3.47 & 24.01 & 64.29 & 49.80 & 64.29 & 73.08 \\
 & ViT-L16 & 1600 & \miniscule{} & 0.55 & 9.23 & 12.45 & 3.42 & 25.65 & 61.72 & 49.59 & 61.72 & 69.43 \\
\midrule
\multirow[t]{9}{*}{MSN} & ViT-B16 & 600 & \miniscule{ep600} & 1.00 & 8.78 & 9.37 & 4.42 & 23.57 & 33.60 & 30.23 & 33.60 & 36.49 \\
 & \multirow[t]{3}{*}{ViT-B4} & \multirow[t]{3}{*}{300} & \miniscule{ep300\_extractb} & 0.51 & 0.05 & 14.29 & 5.06 & 19.91 & 30.80 & 26.06 & 30.80 & 35.15 \\
 &  &  & \miniscule{ep300\_extracts} & nan & nan & 14.20 & 5.58 & 20.67 & 34.57 & 28.91 & 34.57 & 39.62 \\
 &  &  & \miniscule{ep300} & 0.86 & 5.07 & 9.15 & 4.83 & 19.91 & 27.69 & 24.86 & 27.69 & 30.70 \\
 & ViT-L16 & 300 & \miniscule{ep600} & 0.55 & 4.54 & 12.60 & 7.96 & 25.66 & 33.99 & 30.01 & 33.99 & 37.30 \\
 & \multirow[t]{3}{*}{ViT-L7} & \multirow[t]{3}{*}{200} & \miniscule{ep200\_extractb} & nan & nan & 14.46 & 4.93 & 19.99 & 29.08 & 25.60 & 29.08 & 32.17 \\
 &  &  & \miniscule{ep200\_extracts} & nan & nan & 14.34 & 6.59 & 21.79 & 28.29 & 25.66 & 28.29 & 30.89 \\
 &  &  & \miniscule{ep200} & 0.55 & 2.48 & 11.95 & 5.11 & 20.09 & 27.63 & 25.07 & 27.63 & 30.16 \\
 & ViT-S16 & 800 & \miniscule{ep800} & nan & nan & 5.07 & 3.29 & 23.89 & 36.35 & 32.51 & 36.35 & 39.64 \\
\midrule
\multirow[t]{10}{*}{MUGS} & \multirow[t]{3}{*}{ViT-B16} & \multirow[t]{3}{*}{400} & \miniscule{ep400\_extractb} & 0.54 & 1.54 & 15.02 & 3.27 & 20.37 & 30.83 & 27.32 & 30.83 & 33.82 \\
 &  &  & \miniscule{ep400\_extracts} & 1.55 & 0.00 & 16.59 & 3.52 & 20.70 & 35.13 & 29.79 & 35.13 & 39.91 \\
 &  &  & \miniscule{ep400} & 1.00 & 4.73 & 11.37 & 3.81 & 20.91 & 30.34 & 27.03 & 30.34 & 33.24 \\
 & \multirow[t]{3}{*}{ViT-L16} & \multirow[t]{3}{*}{400} & \miniscule{ep250\_extractb} & nan & nan & 13.92 & 3.29 & 19.12 & 29.28 & 25.72 & 29.28 & 31.24 \\
 &  &  & \miniscule{ep250\_extracts} & nan & nan & 10.17 & 3.65 & 19.69 & 30.89 & 27.01 & 30.89 & 33.87 \\
 &  &  & \miniscule{ep250} & 0.55 & 3.11 & 12.31 & 3.14 & 19.12 & 29.22 & 26.02 & 29.22 & 31.49 \\
 & \multirow[t]{4}{*}{ViT-S16} & 100 & \miniscule{ep100} & nan & nan & 5.11 & 3.07 & 25.83 & 43.86 & 38.22 & 43.86 & 48.68 \\
 &  & 300 & \miniscule{ep300} & nan & nan & 5.27 & 3.49 & 23.37 & 39.33 & 34.22 & 39.33 & 43.88 \\
 &  & \multirow[t]{2}{*}{800} & \miniscule{ep800\_extracts} & 0.96 & 5.59 & 12.76 & 3.37 & 22.69 & 44.12 & 37.13 & 44.12 & 50.36 \\
 &  &  & \miniscule{ep800} & nan & nan & 4.94 & 3.71 & 23.01 & 37.91 & 33.40 & 37.91 & 42.11 \\
\midrule
MoCo-v1 & RN50 & 200 & \miniscule{ep200} & 0.85 & 13.64 & 23.16 & 3.74 & 41.39 & 79.35 & 70.51 & 79.35 & 83.94 \\
\midrule
\multirow[t]{3}{*}{MoCo-v2} & \multirow[t]{3}{*}{RN50} & \multirow[t]{2}{*}{200} & \miniscule{ep200} & 0.85 & 6.83 & 23.64 & 2.97 & 34.28 & 68.86 & 61.07 & 68.86 & 74.17 \\
 &  &  & \miniscule{vissl} & 0.85 & 8.68 & 22.44 & 3.48 & 35.45 & 72.63 & 63.90 & 72.63 & 77.51 \\
 &  & 800 & \miniscule{ep800} & 0.85 & 4.44 & 22.36 & 3.42 & 31.07 & 60.46 & 53.30 & 60.46 & 64.75 \\
\midrule
\multirow[t]{5}{*}{MoCo-v3} & \multirow[t]{3}{*}{RN50} & 100 & \miniscule{ep100} & 0.85 & 6.26 & 20.32 & 3.08 & 30.51 & 63.99 & 54.58 & 63.99 & 69.19 \\
 &  & 300 & \miniscule{ep300} & 0.85 & 1.24 & 22.13 & 3.62 & 27.84 & 56.19 & 47.16 & 56.19 & 62.24 \\
 &  & 1000 & \miniscule{ep1000} & 0.85 & 0.58 & 21.85 & 2.98 & 26.26 & 53.00 & 44.46 & 53.00 & 59.58 \\
 & ViT-B & 300 & \miniscule{ep300} & 1.00 & 10.12 & 10.38 & 2.36 & 23.86 & 41.37 & 35.83 & 41.37 & 45.87 \\
 & ViT-S & 300 & \miniscule{ep300} & nan & nan & 5.42 & 3.85 & 27.94 & 46.05 & 40.41 & 46.05 & 50.40 \\
\midrule
NPID & RN50 & 200 & \miniscule{} & 0.85 & 18.44 & 26.01 & 3.02 & 48.32 & 86.15 & 78.70 & 86.15 & 89.85 \\
\midrule
NPID++ & RN50 & 800 & \miniscule{} & 0.85 & 16.39 & 24.75 & 2.46 & 44.45 & 83.04 & 74.72 & 83.04 & 87.23 \\
\midrule
ODC & RN50 & 440 & \miniscule{440ep\_mmselfsup} & 0.85 & 15.16 & 25.51 & 3.78 & 45.29 & 80.63 & 73.01 & 80.63 & 84.24 \\
\midrule
\multirow[t]{10}{*}{Openclip} & ViT-B32 & 32 & \miniscule{} & 1.13 & 6.51 & 12.93 & 2.34 & 22.91 & 43.83 & 35.83 & 43.83 & 50.02 \\
 & \multirow[t]{4}{*}{ViT-H14} & \multirow[t]{4}{*}{32} & \miniscule{extractb} & nan & nan & 12.27 & 2.94 & 15.73 & 30.52 & 24.48 & 30.52 & 36.09 \\
 &  &  & \miniscule{extractpred} & nan & nan & 9.36 & 2.45 & 15.73 & 29.24 & 23.54 & 29.24 & 34.55 \\
 &  &  & \miniscule{extracts} & nan & nan & 12.91 & 2.72 & 16.10 & 36.39 & 27.57 & 36.39 & 43.73 \\
 &  &  & \miniscule{} & 0.00 & 0.80 & 12.13 & 2.66 & 15.59 & 30.63 & 24.23 & 30.63 & 36.30 \\
 & ViT-L14 & 32 & \miniscule{} & 0.55 & 1.43 & 12.26 & 2.41 & 16.65 & 32.16 & 25.81 & 32.16 & 37.35 \\
 & \multirow[t]{4}{*}{ViTg14} & \multirow[t]{4}{*}{32} & \miniscule{extractb} & 0.00 & 0.51 & 12.66 & 2.73 & 15.91 & 32.67 & 25.51 & 32.67 & 38.55 \\
 &  &  & \miniscule{extractpred} & 0.00 & 5.33 & 8.35 & 2.66 & 16.34 & 29.84 & 24.12 & 29.84 & 35.14 \\
 &  &  & \miniscule{extracts} & 0.00 & 0.52 & 13.36 & 2.52 & 16.40 & 34.59 & 26.94 & 34.59 & 40.90 \\
 &  &  & \miniscule{} & 0.00 & 0.83 & 12.58 & 2.88 & 16.29 & 30.87 & 24.61 & 30.87 & 36.17 \\
\midrule
\multirow[t]{6}{*}{PIRL} & \multirow[t]{4}{*}{RN50} & \multirow[t]{2}{*}{200} & \miniscule{ep200\_headmlp} & 0.85 & 6.22 & 24.39 & 3.38 & 34.84 & 72.14 & 63.33 & 72.14 & 76.81 \\
 &  &  & \miniscule{ep200} & 0.85 & 11.06 & 23.60 & 5.22 & 40.72 & 78.59 & 69.32 & 78.59 & 83.22 \\
 &  & \multirow[t]{2}{*}{800} & \miniscule{headmlp} & 0.85 & 5.37 & 21.11 & 2.97 & 30.30 & 62.93 & 55.12 & 62.93 & 67.61 \\
 &  &  & \miniscule{} & 0.85 & 8.90 & 23.10 & 3.37 & 36.22 & 74.20 & 64.58 & 74.20 & 79.69 \\
 & \multirow[t]{2}{*}{RN50w2} & \multirow[t]{2}{*}{400} & \miniscule{headmlp} & 0.74 & 0.00 & 25.43 & 3.42 & 29.50 & 58.43 & 51.75 & 58.43 & 62.72 \\
 &  &  & \miniscule{} & 0.74 & 0.27 & 27.21 & 3.35 & 31.58 & 68.32 & 57.96 & 68.32 & 73.60 \\
\midrule
Relativeloc & RN50 & 70 & \miniscule{70ep\_mmselfsup} & 0.85 & 35.32 & 22.59 & 4.75 & 63.51 & 94.14 & 90.05 & 94.14 & 95.61 \\
\midrule
\multirow[t]{2}{*}{RotNet} & \multirow[t]{2}{*}{RN50} & \multirow[t]{2}{*}{105} & \miniscule{in1k} & 0.85 & 36.22 & 21.71 & 0.19 & 58.96 & 92.13 & 87.09 & 92.13 & 94.03 \\
 &  &  & \miniscule{in22k} & 0.85 & 22.63 & 25.88 & 3.07 & 52.42 & 88.48 & 82.62 & 88.48 & 91.17 \\
\midrule
SeLa-v2 & RN50 & 400 & \miniscule{ep400\_2x224} & 0.85 & 8.75 & 20.99 & 2.93 & 33.51 & 62.60 & 55.56 & 62.60 & 67.36 \\
\midrule
\multirow[t]{19}{*}{SimCLR} & \multirow[t]{2}{*}{RN101} & 100 & \miniscule{ep100} & 0.71 & 18.38 & 11.25 & 3.12 & 33.46 & 68.80 & 60.14 & 68.80 & 74.06 \\
 &  & 1000 & \miniscule{} & 0.71 & 4.24 & 20.16 & 3.45 & 28.56 & 60.89 & 51.17 & 60.89 & 67.13 \\
 & \multirow[t]{14}{*}{RN50} & \multirow[t]{7}{*}{100} & \miniscule{bs4096\_ep100} & 0.85 & 10.12 & 21.71 & 3.32 & 36.00 & 72.82 & 63.95 & 72.82 & 77.94 \\
 &  &  & \miniscule{d8192\_e100\_m2} & 0.85 & 0.00 & 29.97 & 3.49 & 33.92 & 71.93 & 62.36 & 71.93 & 77.40 \\
 &  &  & \miniscule{dnone\_e100\_m2\_headtlinslin} & 0.85 & 15.92 & 20.36 & 2.92 & 40.03 & 75.32 & 67.22 & 75.32 & 79.96 \\
 &  &  & \miniscule{dnone\_e100\_m2\_headtmlpslin} & 0.85 & 11.14 & 24.06 & 3.43 & 39.47 & 74.33 & 66.02 & 74.33 & 78.81 \\
 &  &  & \miniscule{dnone\_e100\_m2\_headtmlpsmlp} & 0.85 & 8.69 & 22.39 & 3.25 & 35.18 & 73.76 & 64.33 & 73.76 & 79.14 \\
 &  &  & \miniscule{dnone\_e100\_m2\_headtnonesnone} & 0.85 & 25.02 & 20.91 & 2.94 & 49.71 & 77.20 & 70.38 & 77.20 & 81.31 \\
 &  &  & \miniscule{dnone\_e100\_m2} & 0.85 & 11.59 & 19.60 & 3.12 & 35.16 & 74.16 & 64.37 & 74.16 & 79.44 \\
 &  & \multirow[t]{2}{*}{200} & \miniscule{bs256\_ep200\_mmselfsup} & 0.85 & 10.96 & 25.60 & 6.32 & 43.72 & 76.15 & 67.48 & 76.15 & 81.12 \\
 &  &  & \miniscule{ep200} & 0.85 & 8.23 & 21.52 & 3.28 & 33.87 & 70.62 & 61.39 & 70.62 & 76.19 \\
 &  & 300 & \miniscule{dnone\_e100\_m2\_data030} & 0.85 & 8.27 & 24.49 & 3.30 & 36.90 & 74.81 & 65.71 & 74.81 & 80.12 \\
 &  & 400 & \miniscule{ep400} & 0.85 & 9.46 & 18.90 & 3.43 & 32.64 & 69.00 & 59.24 & 69.00 & 74.35 \\
 &  & 800 & \miniscule{ep800} & 0.85 & 11.16 & 16.42 & 3.37 & 31.79 & 67.28 & 57.91 & 67.28 & 73.00 \\
 &  & \multirow[t]{2}{*}{1000} & \miniscule{dnone\_e100\_m2\_data010} & 0.85 & 10.72 & 26.01 & 3.40 & 40.97 & 76.67 & 68.19 & 76.67 & 81.12 \\
 &  &  & \miniscule{} & 0.85 & 9.07 & 17.84 & 4.97 & 32.72 & 66.67 & 57.26 & 66.67 & 72.77 \\
 & \multirow[t]{2}{*}{RN50w2} & 100 & \miniscule{ep100} & 0.74 & 0.00 & 27.17 & 3.06 & 30.71 & 66.81 & 56.99 & 66.81 & 72.53 \\
 &  & 1000 & \miniscule{} & 0.74 & 0.00 & 22.04 & 3.54 & 26.06 & 58.57 & 48.58 & 58.57 & 65.04 \\
 & RN50w4 & 1000 & \miniscule{} & 0.00 & 0.46 & 24.81 & 3.86 & 29.14 & 61.76 & 52.21 & 61.76 & 68.13 \\
\midrule
\multirow[t]{3}{*}{SimSiam} & \multirow[t]{3}{*}{RN50} & \multirow[t]{2}{*}{100} & \miniscule{bs256\_ep100} & 0.85 & 2.80 & 26.00 & 3.23 & 32.87 & 68.85 & 59.77 & 68.85 & 74.77 \\
 &  &  & \miniscule{bs512\_ep100} & 0.85 & 2.87 & 26.08 & 3.40 & 33.19 & 69.07 & 59.56 & 69.07 & 74.87 \\
 &  & 200 & \miniscule{bs256\_ep200\_mmselfsup} & 0.85 & 1.59 & 25.17 & 4.73 & 32.32 & 65.91 & 56.45 & 65.91 & 72.10 \\
\midrule
SpecCL & RN50 & 100 & \miniscule{bs384\_ep100} & 0.85 & 10.86 & 23.38 & 3.25 & 38.34 & 73.91 & 65.82 & 73.91 & 78.39 \\
\midrule
\multirow[t]{9}{*}{SwAV} & \multirow[t]{7}{*}{RN50} & 100 & \miniscule{ep100} & 0.85 & 2.98 & 23.40 & 1.81 & 29.04 & 62.09 & 52.83 & 62.09 & 68.61 \\
 &  & \multirow[t]{2}{*}{200} & \miniscule{ep200\_bs256} & 0.85 & 7.10 & 18.06 & 2.59 & 28.59 & 60.69 & 51.45 & 60.69 & 67.14 \\
 &  &  & \miniscule{ep200} & 0.85 & 1.63 & 22.02 & 3.07 & 27.56 & 60.27 & 50.36 & 60.27 & 66.78 \\
 &  & \multirow[t]{3}{*}{400} & \miniscule{ep400\_2x224} & 0.85 & 4.11 & 22.49 & 3.30 & 30.75 & 62.22 & 53.41 & 62.22 & 67.97 \\
 &  &  & \miniscule{ep400\_bs256} & 0.85 & 2.77 & 20.23 & 2.98 & 26.82 & 59.22 & 48.98 & 59.22 & 66.00 \\
 &  &  & \miniscule{ep400} & 0.85 & 1.13 & 21.53 & 3.13 & 26.63 & 58.80 & 48.99 & 58.80 & 65.83 \\
 &  & 800 & \miniscule{} & 0.85 & 1.23 & 20.71 & 3.29 & 26.07 & 57.91 & 47.63 & 57.91 & 64.89 \\
 & RN50w2 & 400 & \miniscule{} & 0.74 & 0.00 & 20.70 & 3.04 & 23.98 & 56.33 & 45.79 & 56.33 & 64.08 \\
 & RN50w4 & 400 & \miniscule{} & 0.00 & 0.24 & 19.85 & 3.67 & 23.76 & 54.70 & 43.80 & 54.70 & 63.11 \\
\midrule
\multirow[t]{2}{*}{VICReg} & RN50 & 1000 & \miniscule{} & 0.85 & 1.98 & 21.84 & 2.94 & 27.60 & 56.36 & 47.47 & 56.36 & 62.47 \\
 & RN50w2 & 1000 & \miniscule{} & 0.74 & 0.00 & 22.10 & 2.97 & 25.33 & 47.52 & 40.24 & 47.52 & 53.17 \\
\midrule
\multirow[t]{7}{*}{VICRegL} & \multirow[t]{2}{*}{ConvNext-B} & \multirow[t]{2}{*}{400} & \miniscule{alpha075} & 1.41 & 8.36 & 11.85 & 2.88 & 24.49 & 41.08 & 36.91 & 41.08 & 44.92 \\
 &  &  & \miniscule{alpha09} & 1.41 & 8.05 & 12.59 & 3.20 & 25.24 & 39.78 & 35.66 & 39.78 & 43.71 \\
 & \multirow[t]{2}{*}{ConvNext-S} & \multirow[t]{2}{*}{400} & \miniscule{alpha075} & 1.69 & 10.94 & 9.43 & 3.05 & 25.11 & 41.38 & 37.47 & 41.38 & 44.96 \\
 &  &  & \miniscule{alpha09} & 1.69 & 10.92 & 9.41 & 3.14 & 25.17 & 41.05 & 36.98 & 41.05 & 44.82 \\
 & ConvNext-XL & 150 & \miniscule{alpha075} & 0.00 & 1.42 & 17.26 & 2.44 & 21.12 & 44.11 & 38.05 & 44.11 & 48.44 \\
 & \multirow[t]{2}{*}{RN50} & \multirow[t]{2}{*}{300} & \miniscule{alpha075} & 0.85 & 1.46 & 24.98 & 3.41 & 30.69 & 60.99 & 52.38 & 60.99 & 66.90 \\
 &  &  & \miniscule{alpha09} & 0.85 & 1.14 & 24.80 & 2.93 & 29.71 & 60.25 & 51.26 & 60.25 & 65.99 \\
\midrule
\multirow[t]{7}{*}{iBOT} & \multirow[t]{2}{*}{ViT-B16} & \multirow[t]{2}{*}{400} & \miniscule{extractb} & 0.54 & 0.73 & 16.44 & 3.70 & 21.42 & 37.60 & 31.86 & 37.60 & 42.77 \\
 &  &  & \miniscule{} & 1.00 & 5.12 & 12.63 & 2.77 & 21.51 & 34.60 & 30.14 & 34.60 & 38.08 \\
 & \multirow[t]{3}{*}{ViT-L16} & \multirow[t]{3}{*}{250} & \miniscule{extractb} & nan & nan & 16.05 & 2.66 & 19.57 & 34.54 & 28.96 & 34.54 & 39.05 \\
 &  &  & \miniscule{extracts} & nan & nan & 15.08 & 3.37 & 20.46 & 31.75 & 27.49 & 31.75 & 35.19 \\
 &  &  & \miniscule{} & 0.55 & 1.89 & 14.57 & 2.90 & 19.91 & 31.72 & 27.38 & 31.72 & 35.51 \\
 & \multirow[t]{2}{*}{ViT-S16} & \multirow[t]{2}{*}{800} & \miniscule{extracts} & 0.96 & 5.58 & 13.24 & 3.51 & 23.29 & 45.35 & 37.91 & 45.35 & 51.57 \\
 &  &  & \miniscule{} & nan & nan & 4.51 & 2.74 & 23.13 & 39.42 & 34.03 & 39.42 & 43.71 \\
\end{longtable}

\end{tiny}

\clearpage
\newpage

\section{Secondary results}
\label{appx:sec:secondary_results}

\subsection{Validating the metrics}
\label{appx:sec:results:validating}

\begin{figure}[h!]
    \centering
    \includegraphics[width=0.5\linewidth]{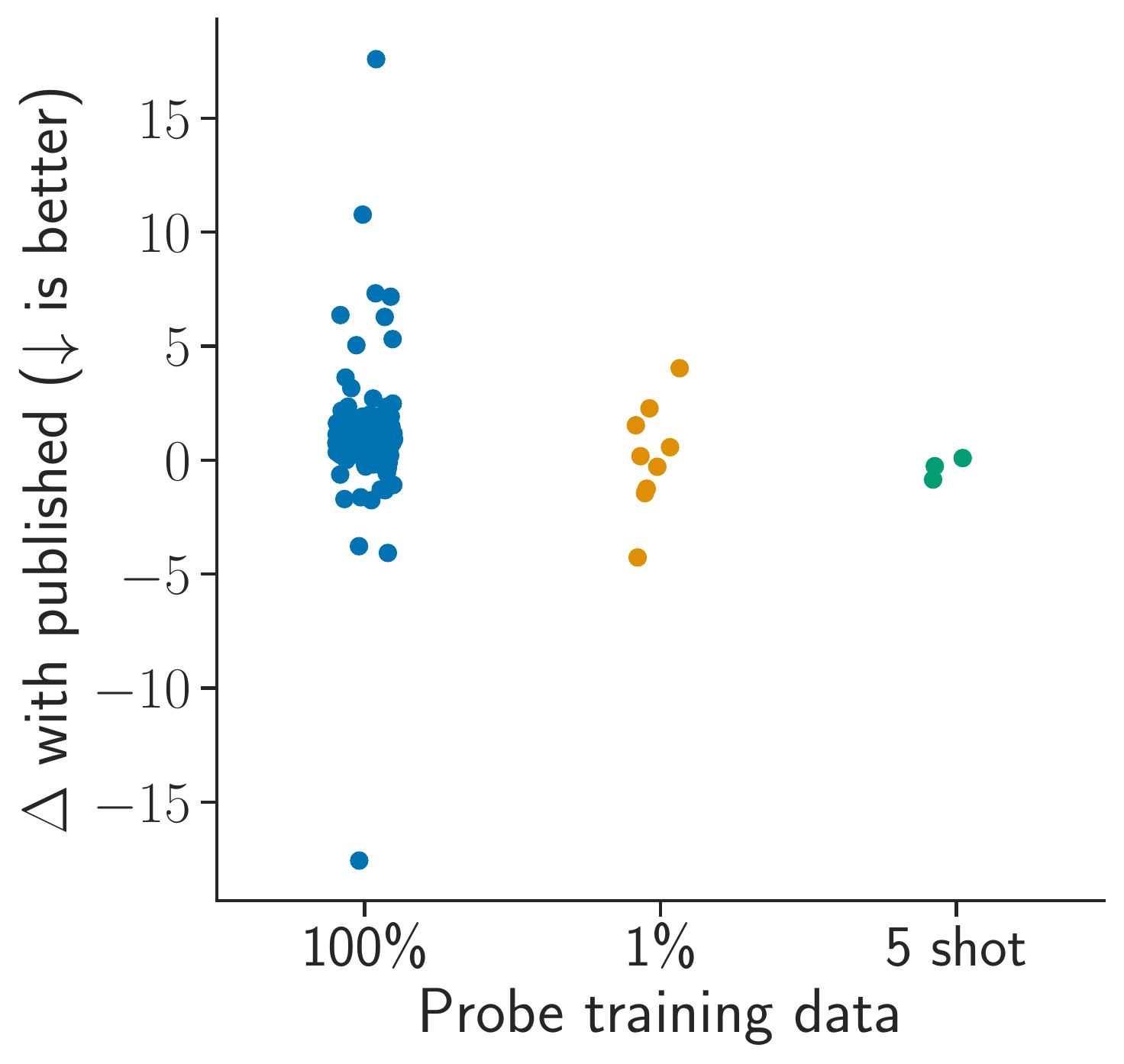}
    \label{appx:fig:diff_published}
\caption{
The difference between the standard metrics we found and the published values (when available).
Negative means that the models performed worst than previously stated.
}
\end{figure}

A secondary contribution of our paper is to evaluate \Npre{} pretrained models, in a controlled and fair fashion.
\Cref{appx:fig:diff_published} validates the metrics we computed by comparing it to previously published results (when available). 
We see that the values we found were generally close to the published results. 
But for the $100\%$ the values can sometimes be very different. 
As a reminder (see \cref{appx:sec:details:metrics}) the two main difference in this setting is that 
\begin{inlinelist}
    \item we do not apply data augmentations when training the probe as it is more realistic and computationally efficient; and
    \item we perform extensive hyperparameter tuning on a validation set.
\end{inlinelist}
The fact that we do not use data augmentations to train the probe, is likely why the values we found are   worst ($1.13 \pm 3.01$) than the published ones (p-value=$1\sci{4}$ with a paired t-test). 
Our extensive hyperparameter tuning can likely explain why for some models we improve the probing results. 
For $1\%$ and 5-shot, previous works also do not use data augmentations and should thus be more similar to our results. 
Indeed, we have that the respective  aggregated values are $0.14 \pm 2.39$ and $-0.34 \pm 0.48$ neither of which are statistically significant.

\begin{small}
\begin{table}[h]
\centering
\caption{Models for which the evaluated metrics are more than 3 accuracy points further than published results.}
\label{appx:tab:big_diff}
\begin{tabular}{llllrrr}
\toprule
\textbf{ Objective } & \textbf{  Arch. } & \textbf{  Epochs } & \textbf{  Other } & \textbf{ 100\% } & \textbf{ 1\% } & \textbf{ 5 Shot } \\
\midrule
BYOL & RN50 & 1000 & \small{augnocolor} & -4.07 &  &  \\
\midrule
DISSL & RN50 & 100 & \small{dnone\_e100\_m2} & 5.04 &  &  \\
\midrule
DeepCluster & RN50 & 200 & \small{bs512\_ep200\_mmselfsup} & -17.57 &  &  \\
\midrule
\multirow[t]{2}{*}{Jigsaw} & \multirow[t]{2}{*}{RN50} & \multirow[t]{2}{*}{105} & \small{in22k} & 17.58 &  &  \\
 &  &  & \small{} & 10.77 &  &  \\
\midrule
\multirow[t]{2}{*}{MSN} & ViT-B16 & 600 & \small{ep600} &  & -4.27 &  \\
 & ViT-L16 & 300 & \small{ep600} & 6.36 &  &  \\
\midrule
MUGS & ViT-S16 & 800 & \small{ep800\_extracts} & -1.71 & 4.03 &  \\
\midrule
PIRL & RN50 & 200 & \small{ep200} & 3.62 &  &  \\
\midrule
Relativeloc & RN50 & 70 & \small{70ep\_mmselfsup} & 3.16 &  &  \\
\midrule
\multirow[t]{2}{*}{RotNet} & \multirow[t]{2}{*}{RN50} & \multirow[t]{2}{*}{105} & \small{in1k} & 7.16 &  &  \\
 &  &  & \small{in22k} & 7.31 &  &  \\
\midrule
\multirow[t]{2}{*}{SimCLR} & RN101 & 100 & \small{ep100} & -3.78 &  &  \\
 & RN50 & 200 & \small{bs256\_ep200\_mmselfsup} & 6.28 &  &  \\
\midrule
SpecCL & RN50 & 100 & \small{bs384\_ep100} & 5.31 &  &  \\
\bottomrule
\end{tabular}
\end{table}

\end{small}

\Cref{appx:tab:big_diff} shows in more detail the values that are more than 3 accuracy points further than published results. 
We see that many of the highly positive values are older SSL models that predict the transformation (Jigsaw, RotNet, Relativeloc, RotNet) and are thus not invariant to data augmentations. 
We hypothesize that given that they are not invariant to the augmentations, training the probe with augmentations performs much better for them, which is why the differences are large.
For the case of MUGS, we note that they do not specify the evaluation pipeline of their models.
In particular, we do not know which block of the ViT they used as the representation. 
\Cref{appx:tab:big_diff} suggests that we likely chose a different feature as the performance of the model is worst on $1\%$ but better on $100\%$.

\subsection{Alternative decomposition}
\label{appx:sec:results:alternative}

In \cref{appcs:sec:risk_dec:alternatives} we have seen that our decomposition is not unique.  
There are two alternative decompositions, but only one of which would be useful for understanding the effect of representation learning.
This decomposition would essentially switch the order of the two generalization errors and thus keep the same interpretation as our decomposition. 
As previously discussed, the estimates for this decomposition would likely be worse than for our decompositions.
In the following, we compute those (worse) estimates of the generalization errors and compare them with the ones in the paper. 
The goal is to make sure that alternative decompositions and estimators do not change the main conclusions from our paper.

\begin{figure}[h!]
    \centering
    \includegraphics[width=0.9\linewidth]{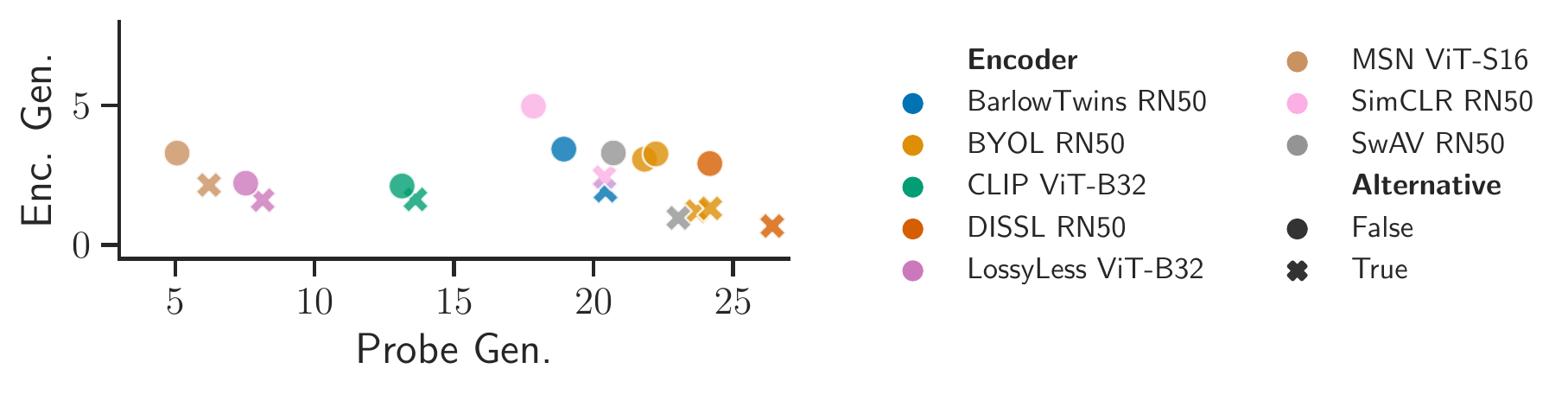}
    \caption{Our risk decomposition and the alternative \cref{appcs:eq:dec_switch_gen_error}, which switches the order of generalization errors, give similar results.
    Different colors show different encoder while the shape shows whether the generalization errors correspond to the alternative (cross) or our main decomposition (circles). 
    We only show the generalizations as the other components are exactly the same.
    As discussed in \cref{appcs:sec:risk_dec:alternatives} the alternative risk components are likely worst estimates.
    Axes are on the same scale.
    }
    \label{fig:results_alternative}
\end{figure}

\Cref{fig:results_alternative} shows that despite being different components and different estimators, the estimated probe and encoder generalization of the alternative and main decompositions are highly related.
This is reassuring as it suggests that using a different decomposition would not change our interpretation of the results (encoder generalization still seems small in absolute terms and the relative ordering of models seems similar).
Note that the plot is rectangular as the axes are on the same scale %(to understand whether the alternative decomposition would change the general trends and analysis of our paper) 
but encoder generalization is smaller than probe generalization.

\subsection{Trends over time}
\label{appx:sec:results:trends}

\begin{figure}[h!]
    \centering
    \begin{subfigure}[t]{0.35\linewidth}
         \centering
 \includegraphics[width=\linewidth]{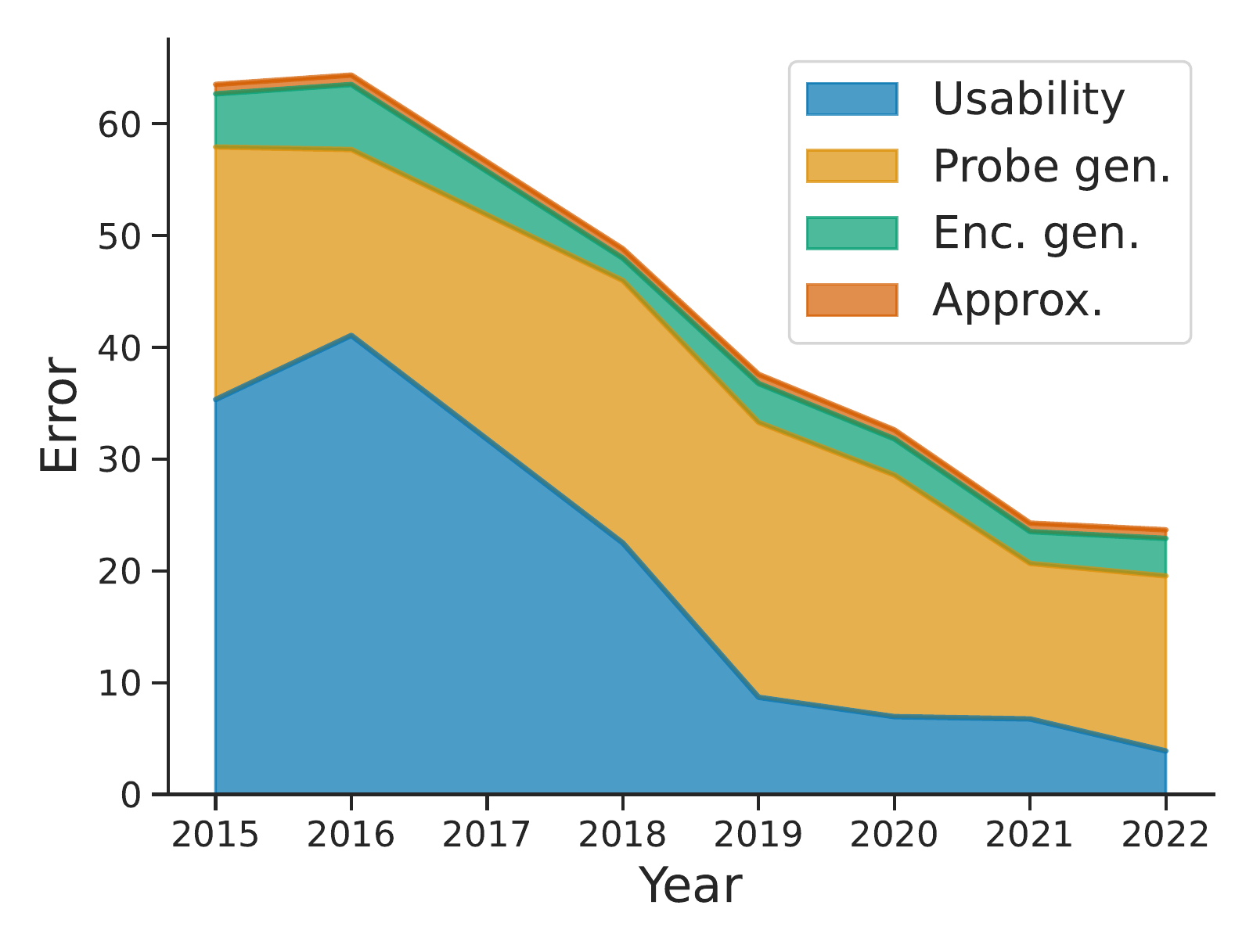} 
    \caption{Average per year}
    \label{fig:trends_avg}
    \end{subfigure}
    \hspace{1em}
    \begin{subfigure}[t]{0.54\linewidth}
\includegraphics[width=\linewidth]{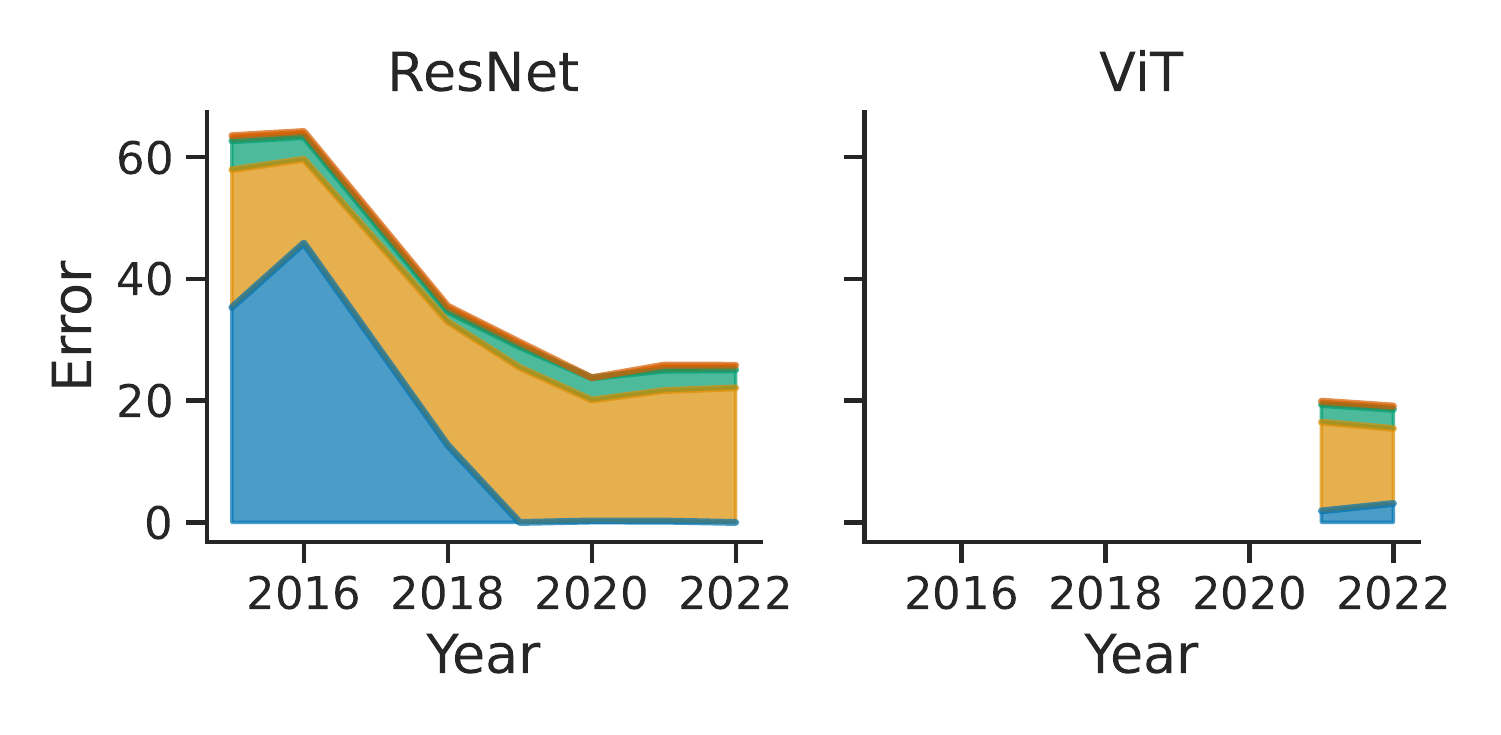}  
\caption{Best per year and neural family}
\label{fig:trends_arch}
    \end{subfigure}
    \caption{Evolution of risk components over time.
    Lower is better.
    (a) the risk components are averaged over all models published in a given year;
    (b) the risk components are the best over models published in a given year and for a specific family of encoder's architecture (ResNet and ViT).
    }
    \label{fig:trends_appx}
\end{figure}

In \cref{fig:trends_main}, we saw how risk components have been changing over time for the best ImageNet-pretrained models published in that year.
\Cref{fig:trends_avg} shows instead the average over all models published in that year (including those trained on the ImageNet-22K, LAION and CLIP dataset).
We see that the global trend is essentially the same: usability has been driving improvements but probe generalization is now what matters.

\begin{figure}[h!]
    \centering
    \includegraphics[width=\textwidth]{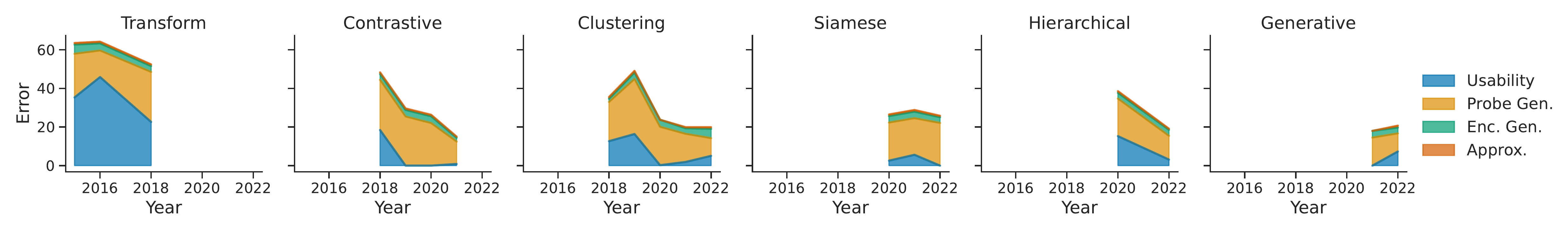}
    \caption{Evolution of average risk components over time. 
    For a specific SSL mode.
    }
    \label{fig:trends_mode}
\end{figure}

We can also consider more fine-grained trends by looking at how risk components have been changing for a type of self-supervised learning method (\cref{fig:trends_mode}) or encoder architecture (\cref{fig:trends_arch}).
\cref{fig:trends_mode} shows that much of the improvements in usability have been achieved under the contrastive learning framework. 
Using \cref{fig:trends_arch} we also see that the more recent improvements in probe generalization have mostly come from the clustering paradigm using ViTs.

\subsection{Scaling laws}
\label{appx:sec:results:scaling_law}

% To derive a scaling law, note that we know by definition (besides empirical) that:
% \begin{equation}
% \mathrm{perf}(n,m) = \eA(n,m) + \eR(n,m) +  \eE(m)  +  \eP(n,m)
% \end{equation}
% where $n$ is the number of samples and $m$ is the encoder. 
% Now by standard scaling laws we also have  :
% \begin{equation} \eP(n,m) = \frac{C(m)}{n^{\alpha(m)}}
% \end{equation}
% The issue is we do not want to fit $C(m)$ and $\alpha(m)$ for each model. 
% For $\alpha(m)$ we simply assume that it is the constant for each model $\alpha(m)=\alpha$. 
% For $C(m)$ we can use the fact that we know $\eP(N)$ where $N$ is ImageNet training size. We thus have:
% \begin{equation}  \eP(n,m)n^{\alpha}  =  C(m) = \eP(N,m)N^{\alpha}
% \end{equation}
% Putting all together we have:
% \begin{align}
% \mathrm{perf}(n,m) &= \eA(n,m) +  \eE(n,m)  + \eR(n,m) +  \eP(n,m) \\
% &= \eA(N,m) + \eR(N,m) +  \eE(N,m)  +   \eP(N,m) \left(\frac{N}{n}\right)^{\alpha}
% \end{align}

\begin{table}[h]

\caption{
Our scaling law predicts well performance. %in different settings.  
Numbers are $R^2$ on test data.%, \ie, variance normalized MSE substracted to one, which can thus be negative.
``Std'' is a standard scaling law fitted on all encoders.
``e=fam.'' fits separately ViT's and ResNet's, while ``e=arch.'' and `` e=obj.'' fit separate laws for each architecture and SSL objectives.
Columns show held-out test sets.
%``Random'' tests 3 settings for each encoder.
``IID'' test on 3/5 settings for each encoder.% (60\%).
``Cntr'', ``Enc.'', ``2022'',  respectively tests on all encoders that are contrastive, % (37\%), 
ViTs% (27\%)
, from last year. % (20\%).
Missing values mean scaling laws cannot predict this test set.
}
\label{tab:scaling_laws}
%\begin{small}
\begin{center}
\begin{tabular}{llrrrr}
\toprule
&&  \multicolumn{4}{c}{test set} 
 \\
  \cmidrule(lr){3-6}
 scaling law &  param. & \iid & 2022 & cntr. & ViT \\
\midrule
Std & 5 & 0.31 & -0.12 & 0.46 & -0.98 \\
\ \ e=family & 11 & 0.60 & 0.65 & 0.72 &  \\
\ \ e=arch. & 41 & 0.63 & 0.66 &  &  \\
\ \ e=obj. & 86 & 0.82 &  &  &  \\
%\ \ e=enc. & 422 & 0.97 &  &  &  \\
Ours & 2 & \textbf{0.94} & \textbf{0.91} & \textbf{0.96} & \textbf{0.84} \\
%Ours 1 & 1 & 0.90 & 0.86 & 0.92 & 0.72 \\
%test_perc & <NA> & 0.60 & 0.24 & 0.36 & 0.27 \\
\bottomrule
\end{tabular}
\end{center}
%\end{small}
\end{table}

We propose the following scaling law  based on our decomposition:
\begin{equation}\label{eq:our_scaling_law}
 \rUS(n) \approx \eA +\eE + (1-{\color{burntorange}W}) \eR + ({\color{burntorange}W}\eR + \eP )\pa{\frac{N}{n}}^{\color{burntorange}\alpha}
\end{equation}
where $\eA,\eE,\eR,\eP$ are respectively the risk components for the approximation, encoder generalzation, usability, and probe generalization.
$n$ is the number of samples used to train the probe, $N$ is the number of samples used to estimate the decomposition, and ${\color{burntorange}\alpha},{\color{burntorange} W}$ are fitted parameters % respectively
quantifying sample efficiency and $\eR$'s dependence on $n$.

\Cref{fig:evaluation_scalinglaw}
shows that \cref{eq:our_scaling_law} fits all results very well ($R^2=0.94$, ${\color{burntorange} \alpha}=0.15$, ${\color{burntorange} w}=0.51$).
\cref{tab:scaling_laws} shows that it predicts better the performance of held-out models compared to standard neural scaling laws \cite{kaplan_scaling_2020,rosenfeld_constructive_2020} of the form
$
\rUS(n,p,e) \approx {\color{burntorange}I_e} + \frac{\color{burntorange} C_e}{n^{\color{burntorange}\alpha_e}} + \frac{\color{burntorange} K}{p^{\color{burntorange}\beta}}
$
where $p$ is the number of probe's parameters, $e$ is a set of encoders for which we train the same scaling law (\eg those with the same architecture), and $\{{\color{burntorange}I_e}\}_e,\{{\color{burntorange} C_e}\}_e,\{{\color{burntorange} \alpha_e}\}_e,{\color{burntorange} K},{\color{burntorange} \beta}$ are fitted.

\subsection{Trade-offs}
\label{appx:sec:results:tradeoff}

As discussed in \cref{appcs:sec:risk_dec:tradeoffs} the standard approximation-estimation tradeoff implies three representation learning tradeoffs.
Let us look at all possible tradeoffs empirically.

\begin{figure}[h]
    \centering
    \begin{subfigure}[t]{0.48\linewidth}
         \centering
 \includegraphics[width=\linewidth]{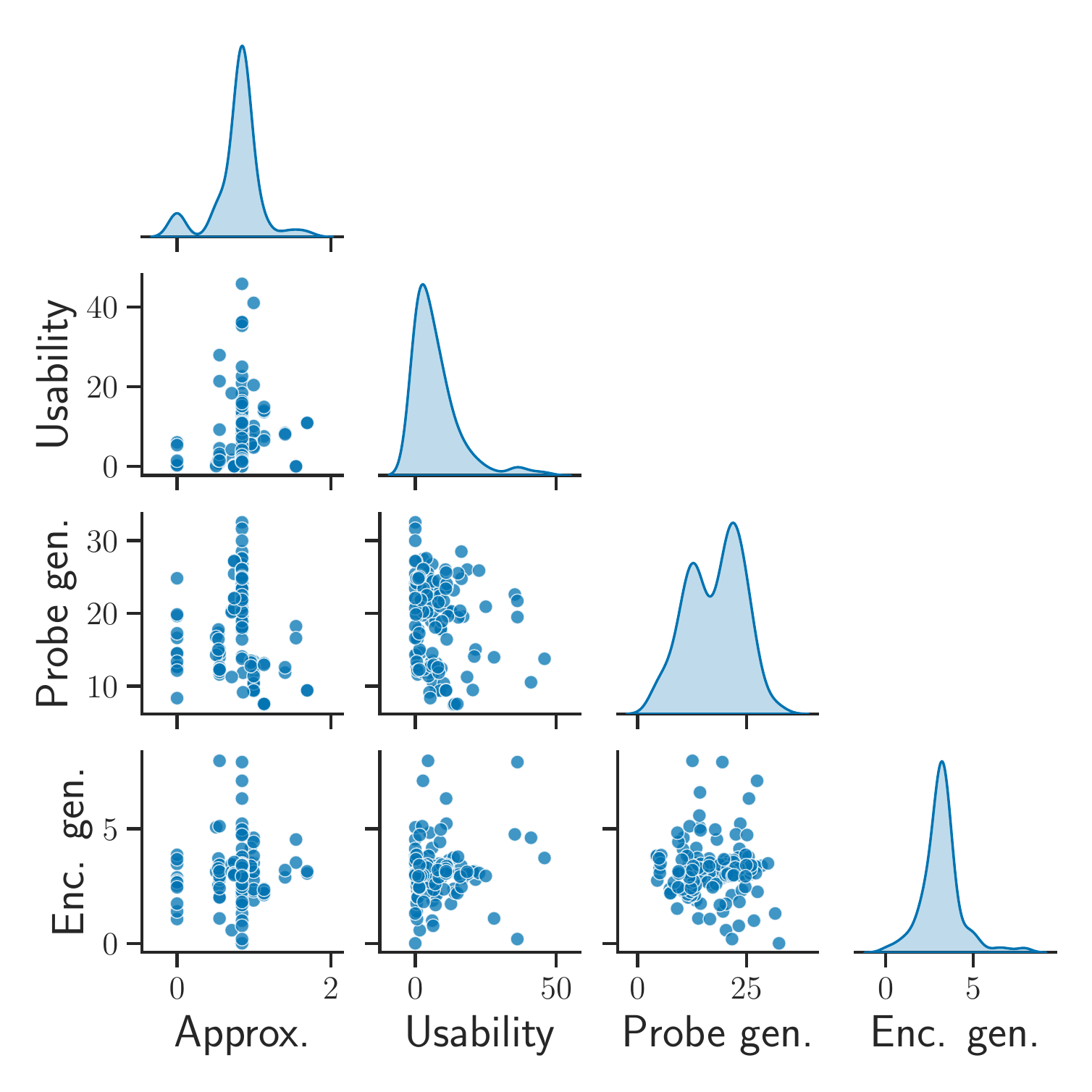}
    \caption{All models}
    \label{appx:fig:all_tradeoffs}
    \end{subfigure}
    \hfill{}
    \begin{subfigure}[t]{0.48\linewidth}
\includegraphics[width=\linewidth]{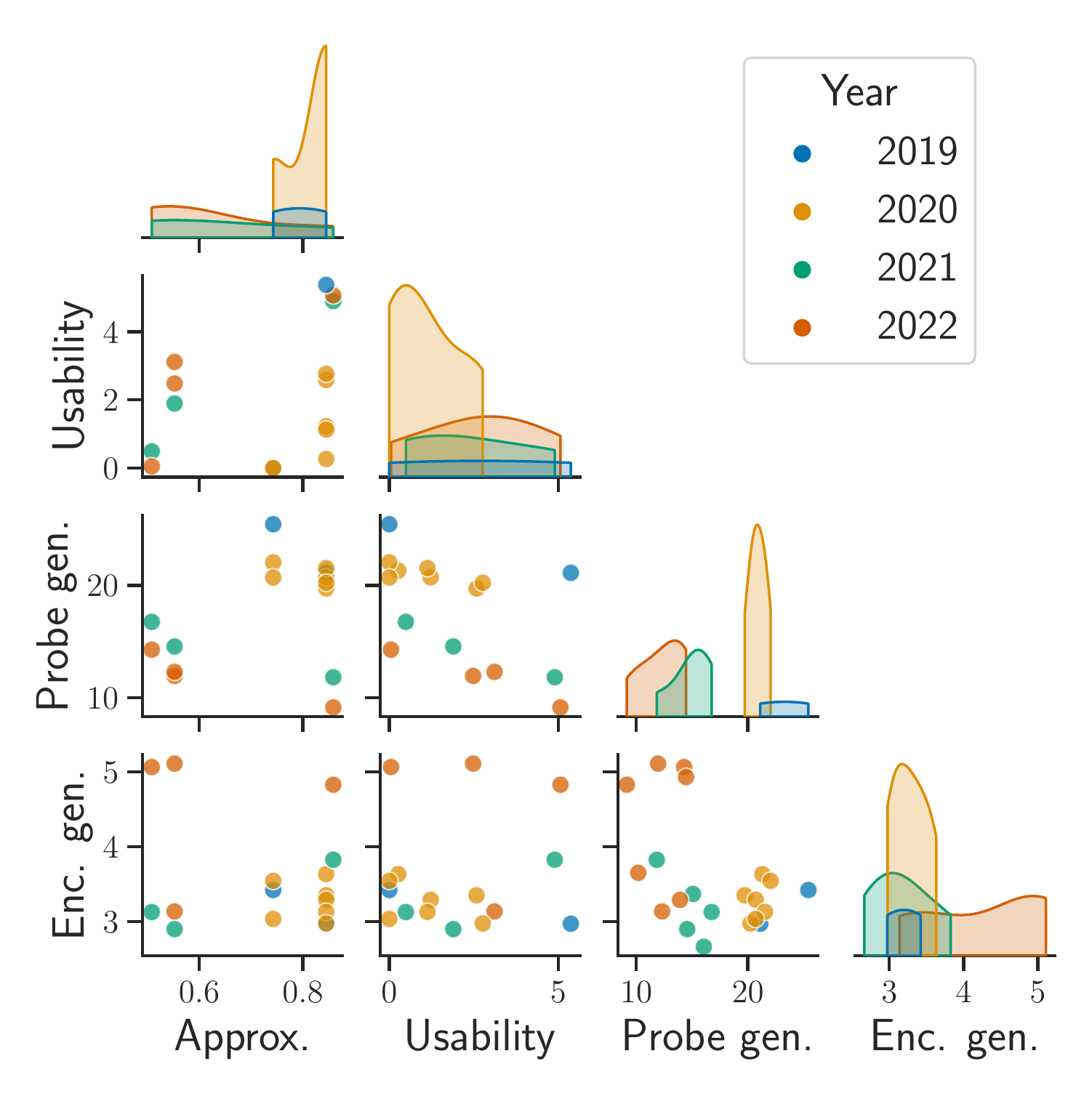}
\caption{Best 15\% per year}
\label{appx:fig:all_tradeoffs_best}
    \end{subfigure}
    \caption{
All potential tradeoffs between our risk components when considering
(a) all models pretrained on ImageNet.
(b) the best 10\% of model for recent years (since 2019).
    }
    \label{appx:fig:tradeoff}
\end{figure}

\Cref{appx:fig:tradeoff} shows all possible pairwise tradeoffs between the risk components.
We see that when considering all models in aggregation there seem to be no tradeoffs (\cref{appx:fig:all_tradeoffs}).
\Cref{appx:fig:all_tradeoffs_best} instead shows the best performing models for each (recent) year.
Although there are not many points, we can see the usability-probe gen tradeoff and a glimpse of the approximation-probe gen tradeoff.
But there does not seem to be any sign of any approximation-encoder generalization tradeoff.
The fact that there seems to be a tradeoff for the probe but not the encoder, might be related to the fact that over-parametrized models seem to not follow the standard approximation-estimation tradeoff \cite{belkin_reconciling_2019,yang_rethinking_2020,nakkiran_deep_2020,dar_farewell_2021,neal_modern_2018}.
This over-parametrization could potentially explain why the encoder generalization is smaller than the probe generalization.
That being said we see in \cref{appx:fig:all_tradeoffs_best} that the approximation error is really small and so tradeoffs that depend on it are likely not important for practical SSL.

We emphasize that the tradeoff curve given by the top performing models (\cref{fig:tradeoff_main} and \cref{appx:fig:all_tradeoffs_best}) does not correspond to modifying a single hyperparameter on the best performing model, but those are instead models trained with different SSL objective, architectures, epochs, and many other hyperparameters.
For example, the best-performing models for 2022 (in red in \cref{fig:tradeoff_main}) include \texttt{\detokenize{msn_vitb4_ep300}}, \texttt{\detokenize{msn_vitl7_ep200}}, \texttt{\detokenize{mugs_vitl16_ep250}}.
% I.e those are three different SSL objectives (BEiT, MUGS, MSN) using different ViT architectures and trained for different numbers of epochs.

\begin{figure}[h!]
    \centering
    \includegraphics[width=0.65\linewidth]{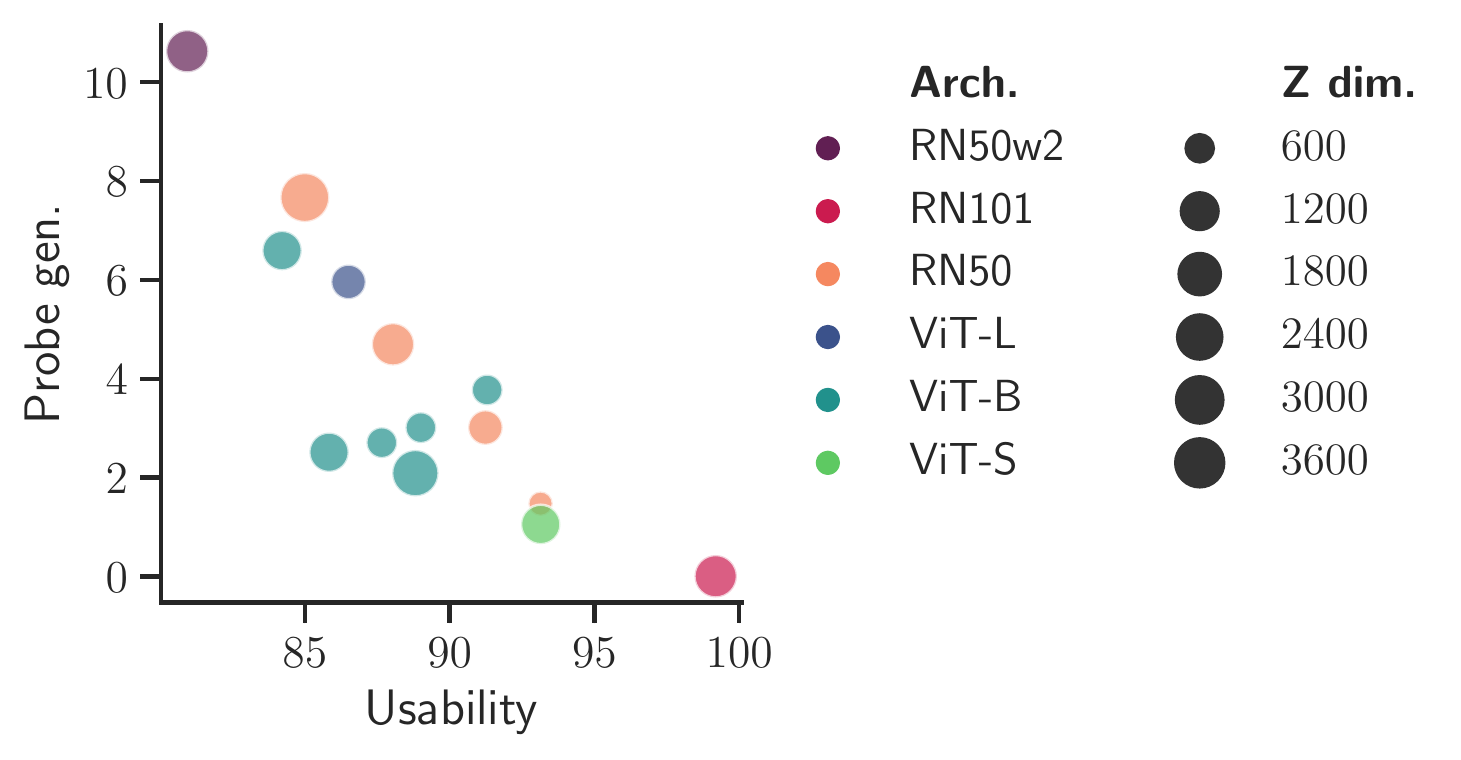}
    \caption{
    The tradeoff between probe generalization and usability also holds when the encoder is untrained.
    Each point shows the probe generalization (y-axis) and usability (x-axis) for representations that come from a different randomly initialized encoder.
    }
    \label{fig:results_initialized}
\end{figure}

We have seen the U-P tradeoff for encoders that are pretrained with SSL.
Our risk components and their tradeoffs are nevertheless not specific to SSL. 
A natural question is thus whether we see the same tradeoff for other representations.
\Cref{fig:results_initialized} provides evidence of such tradeoff more generally, by considering representations coming from untrained encoders.

\subsection{Uniformity, alignment, and effective dimensionality}
\label{appx:sec:results:statistics}

Many previous works have proposed different simple statistics to measure the quality of SSL representations. 
Three very common such statistics that are easy to compute are (we give code in PyTorch): 
\begin{description}
\item[Effective dimensionality]
\citet{dubois_improving_2022} recently proved that the effective dimensionality, \ie, the dimension of the space spanned by the representation's support, is a key property to ensure that downstream tasks with a few classes can be performed.
The requirement for large effective dimensionality was also indirectly suggested by theoretical arguments of \cite{saunshi_understanding_2022,haochen_provable_2021}.
For a fixed ambient dimensionality, the dimensionality collapse literature \cite{hua_on_2021,jing_understanding_2022}
also suggested that small effective dimensionality can be an issue.
To compute the effective dimensionality we simply compute the rank (under some small tolerance) of Pearson's correlation coefficient matrix of the represented training set as follows: \\
\pyth{ torch.linalg.matrix_rank(Z.T.corrcoef(), atol=1e-4, rtol=0.01)}.
\item[Uniformity] 
\citet{wang_understanding_2020} and follow-ups, \eg \cite{wang_understanding_2021}, show that contrastive learning forces representations to be approximately uniformly distributed on a hypersphere, and they hypothesize based on empirical results that this is a desired property. 
But more recent theories \cite{dubois_improving_2022,wang_chaos_2022} suggest the opposite.
We test the usefulness of uniformity using \citet{wang_understanding_2020} original estimator: \\
\pyth{  torch.pdist(F.normalize(Z, dim=-1), p=2).pow(2).mul(-2).exp().mean().log()}.
\item[Alignment] Countless works \cite{ericsson_why_2021,dubois_improving_2022,foster_improving_2021,mitrovic_representation_2021,dubois_lossy_2021,ruan_optimal_2022,foster_improving_2021,miao_learning_2022} have proven or hypothesized that good encoders should be invariant to data augmentations.
Although perfect invariance will not be achieved, it is natural to hypothesize that good encoders will map equivalent/augmented examples close together.
Such property is called alignment  \cite{wang_understanding_2020}, and can for example be quantified using the distance between augmented samples $z1,z2$:\\
\pyth{  (z1 - z2).norm(dim=-1).pow(2).mean()}.
\end{description}

\begin{figure}[h!]
    \centering
\includegraphics[width=0.7\linewidth]{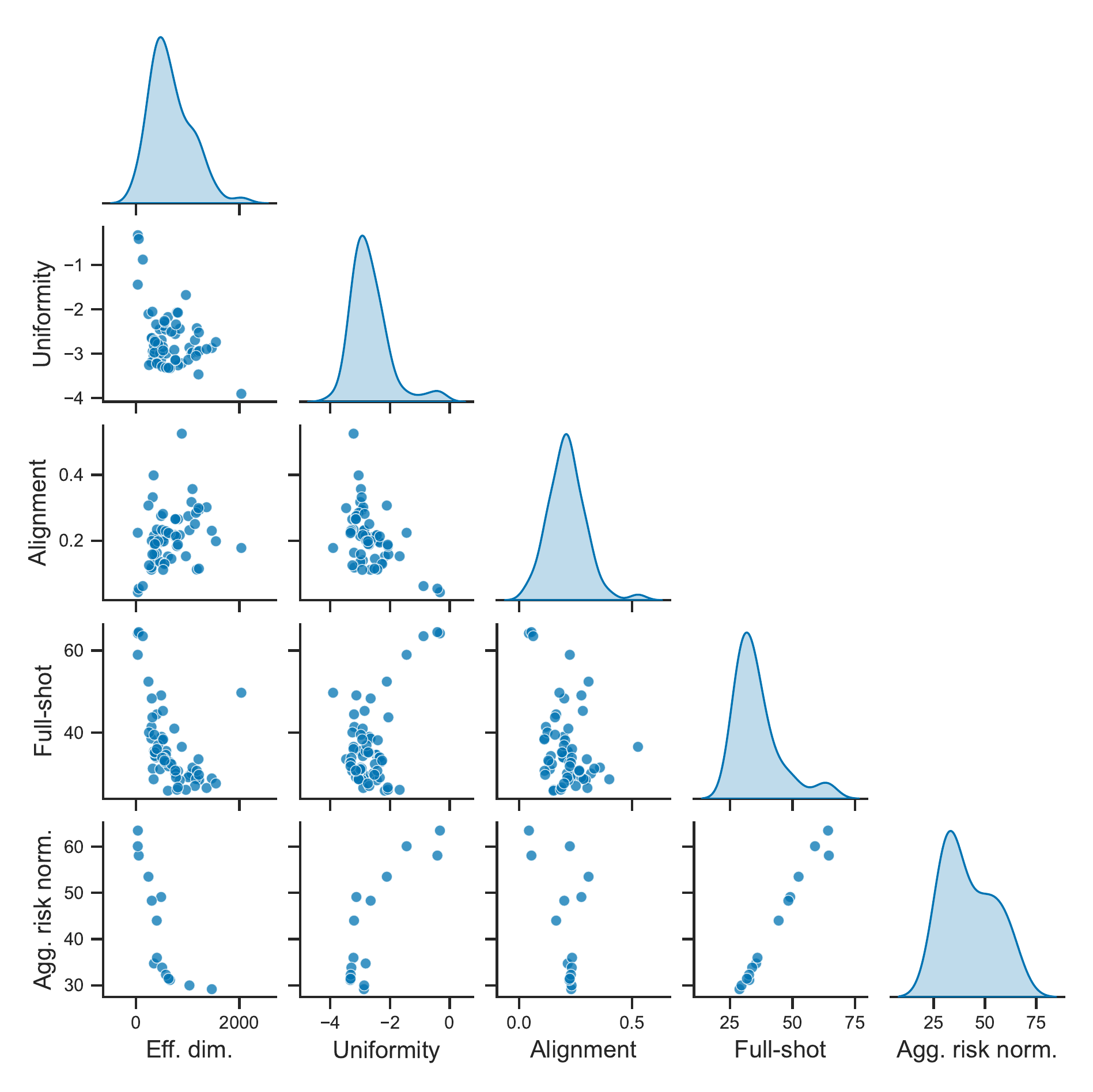}
    \caption{Relation between various statistics (``Uniformity'', ``Alignment'', ``Eff. Dim.'') and probing performance from normalized (``Agg. Risk Norm.'') and normalized representations (``Agg. Risk'') for the case of ResNet50s.
    For normalized representations, we only evaluated the models from \texttt{VISSL}.
    }
    \label{fig:statistics}
\end{figure}

In the following, we tested how well each of the previous statistics can predict the performance of a downstream model for the case of Resnet50s.
Note that both uniformity and alignment were proposed by \citet{wang_understanding_2020} in the case where the downstream representations are normalized before being probed.
This is not the standard probing regime and we found that normalizing representations decreases downstream performance by $0.44 \pm 0.28$. 
We nevertheless compared the statistics to the performance in both the normalized and the unnormalized regimes for a subset of the models (19 models from VISSL) to compare in a setting more similar to \cite{wang_understanding_2020}.
\cref{fig:statistics} shows qualitatively all the results.
For quantitative results, we evaluated the model $\log(\mathrm{agg\_risk}) = \delta + \alpha \log(\mathrm{eff\_dim}) + \beta \mathrm{uniformity} + \gamma \mathrm{alignment} $ to test how well each statistics (conditionally) correlates with the downstream performance.\footnote{
This was the best model for predicting the performance using a linear combination of effective dimensionality, uniformity, and alignment with potential log processing}
The fitted model is 
\begin{equation}\label{eq:quantitative_statistics}
    \mathrm{agg\_risk} = 93 - 9.5 \cdot \log(\mathrm{eff\_dim}) - 0.51 \cdot \mathrm{uniformity} + 4.4 \cdot \mathrm{alignment},
\end{equation}
it achieves an $R^2$ of $0.58$.

\begin{paragraph}{Effective dimensionality correlates with performance}
\cref{fig:statistics} shows that higher effective dimensionality seems to improve downstream performance.
Quantitatively, the effective dimensionality is statistically significant with a p-value of $4\sci{11}$ and the simple model \cref{eq:quantitative_statistics} suggests that increasing the effective dimensionality by a factor of 3 improves the accuracy by 10 percentage points.
\end{paragraph}

\begin{paragraph}{Uniformity is not predictive of performance}
Looking at \cref{fig:statistics} it seems that uniformity is correlated with performance.
But the quantitative results show that the p-value is $0.74$, so the (conditional) correlation is not statistically significant.
The difference between the quantitative and qualitative results comes from the fact that the estimated uniformity is actually highly correlated with effective dimensionality.
For example, if we removed the effective dimensionality from \cref{eq:quantitative_statistics} the coefficient of uniformity would increase to 6, and the p-value decrease to $0.001$ ($R^2$ is only 0.19).
This shows that although the estimator of uniformity does correlate with performance (as experimentally shown by \cite{wang_understanding_2020}) it is only because it correlates with effective dimensionality.
Beyond this, uniformity is not predictive of performance, which supports \citepos{dubois_improving_2022} theory.
\end{paragraph}

\begin{paragraph}{Alignment does not correlate with performance}
\cref{fig:statistics} shows that alignment does not seem to be correlated with (normalized or normalized) performance.
This is further supported quantitatively by the fact that its impact is not statistically significant (p-value of $0.59$),
This is surprising given that previous theories and experiments have shown that alignment does predict performance. 
We do not have a good explanation of why this is the case but note that alignment for examples of the same class (\eg alignment of 2 random dogs rather than the same dog with different augmentations) is highly correlated with performance (coefficient of $40.7$ and p-value of $6\sci{7}$).
\end{paragraph}

% \clearpage
% \newpage

% \bibliography{bib_tatsu}
% \bibliographystyle{icml2022}

\end{document}